%% file: main.tex
\documentclass[10pt,twocolumn,letterpaper]{article}
\usepackage[pagenumbers]{cvpr} 

\usepackage{graphicx}
\usepackage{amsmath}
\usepackage{amssymb}
\usepackage{booktabs}

\usepackage{comment}
\usepackage{multirow}
\usepackage{verbatim}
\usepackage[dvipsnames]{xcolor} %
\usepackage{shortbold}
\usepackage{float}
\makeatletter
\@namedef{ver@everyshi.sty}{}
\makeatother
\usepackage{tikz}
\usepackage{comment}
\usepackage{color}

\usepackage{cite}
\usepackage{booktabs}
\usepackage{graphicx}
\usepackage[export]{adjustbox}

\usepackage[accsupp]{axessibility}  %

\usepackage[pagebackref,breaklinks,colorlinks,allcolors=Green]{hyperref}

\def\cvprPaperID{5989} %
\def\confName{CVPR}
\def\confYear{2023}

\input{fig/teaser/v4/teaser.tex}

\input{preamble}
\begin{document}
\input{sec/0_metadata}
 \maketitle
 \input{sec/0_abstract}

\input{sec/1_introduction}

\input{sec/2_related}

 \input{sec/3_method}

\input{sec/4_experiment}

\input{sec/5_conclusions}
 \input{sec/X_acknowledgements}

 {
    \small
    \bibliographystyle{ieee_fullname}
    \bibliography{macros,main}
}
\newpage
\input{sec/X_supplementary.tex}
\end{document}

%% file: fig/teaser/v4/teaser.tex
\makeatletter
\g@addto@macro\@maketitle{
\vspace{-2.5em}
\begin{figure}[H]
   \setlength{\linewidth}{\textwidth}
\setlength{\hsize}{\textwidth}
\centering
\includegraphics[width=\linewidth]{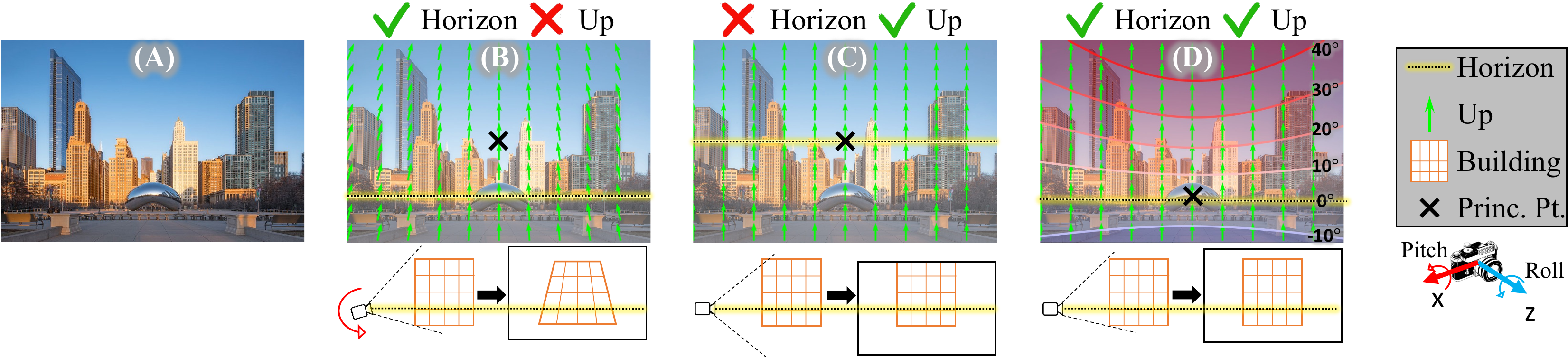} 
\caption{(A): a photo (credit David Clapp) with an off-centered principal point due to cropping.
(B), (C): assuming traditional pinhole model with principal point at the center, as used by~\cite{hold2018perceptual,lee2021ctrl, bogdan2018deepcalib},
there is no way to correctly represent both up directions (wrong in B) and horizon (wrong in C).
(D): Our proposed Perspective Fields model correctly models the Up-vectors ({\color{Green} arrows}) aligned with gravity, and Latitude values (contour line: $-\pi/2$ \includegraphics[width=0.4in,height=8pt]{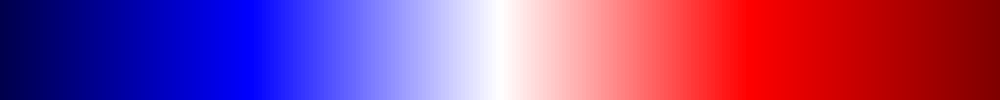} $\pi/2$) with $0^\circ$ on  the horizon.
We can further recover camera parameters Roll $-0.5^\circ$, Pitch $1.7^\circ$, FoV $64.6^\circ$ and principal point at $\times$ from the prediction.} 
     \label{fig:teaser}
\end{figure}
}

%% file: preamble.tex
\usepackage{overpic}
\usepackage{enumitem} %
\usepackage{overpic} %
\usepackage{color}

\definecolor{turquoise}{cmyk}{0.65,0,0.1,0.3}
\definecolor{purple}{rgb}{0.65,0,0.65}
\definecolor{dark_green}{rgb}{0, 0.5, 0}
\definecolor{red}{rgb}{0.8, 0.2, 0.2}
\definecolor{darkred}{rgb}{0.6, 0.1, 0.05}
\definecolor{blueish}{rgb}{0.0, 0.3, .6}
\definecolor{light_gray}{rgb}{0.7, 0.7, .7}
\definecolor{pink}{rgb}{1, 0, 1}
\definecolor{greyblue}{rgb}{0.25, 0.25, 1}
\definecolor{msorange}{rgb}{0.93,0.49,0.19}
\definecolor{msblue}{rgb}{0.33,0.56,0.76}

\newcommand{\bfpar}[1]{{\vspace{1mm} \par \noindent \bf{{#1}}}}

\newcommand{\Fig}[1]{Fig.~\ref{fig:#1}}

\newcommand{\Table}[1]{Table~\ref{tab:#1}}

\newcommand{\Eq}[1]{Eq.~\ref{eq:#1}}

\newcommand{\Sec}[1]{Sec.~\ref{sec:#1}}

\usepackage{blindtext}

\renewcommand{\paragraph}[1]{\vspace{1em}\noindent\textbf{#1}.}

\makeatletter
\DeclareRobustCommand\onedot{\futurelet\@let@token\@onedot}
\def\@onedot{\ifx\@let@token.\else.\null\fi\xspace}

\def\eg{\emph{e.g}\onedot}

\def\wrt{w.r.t\onedot} 
 
\def\etal{\emph{et al}\onedot}
\makeatother

\newcommand\blfootnote[1]{ %
  \begingroup
  \renewcommand\thefootnote{}\footnote{#1} %
  \addtocounter{footnote}{-1} %
  \endgroup
}

%% file: sec/0_metadata.tex
\title{Perspective Fields for Single Image Camera Calibration} 
\author{Linyi Jin$^{1*}$, Jianming Zhang$^2$, Yannick Hold-Geoffroy$^2$, Oliver Wang$^2$, \\Kevin Blackburn-Matzen$^2$, 
Matthew Sticha$^1$, David F. Fouhey$^1$ \\
University of Michigan$^1$, Adobe Research$^2$ \\
{\tt\small $^1$\{jinlinyi,msticha,fouhey\}@umich.edu} \\
{\tt\small $^2$\{jianmzha,holdgeof,owang,matzen\}@adobe.com}
}

%% file: sec/0_abstract.tex
\begin{abstract}
    Geometric camera calibration is often required for applications that understand the perspective of the image.
    We propose Perspective Fields as a representation that models the local perspective properties of an image.
    Perspective Fields contain per-pixel information about the camera view, parameterized as an Up-vector and a Latitude value.
    This representation has a number of advantages; 
    it makes minimal assumptions about the camera model and is invariant or equivariant to common image editing operations like cropping, warping, and rotation. 
    It is also more interpretable and aligned with human perception.
    We train a neural network to predict Perspective Fields and the predicted Perspective Fields can be converted to calibration parameters easily.
    We demonstrate the robustness of our approach under various scenarios compared with camera calibration-based methods and show example applications in image compositing. Project page: 
    {\small \url{https://jinlinyi.github.io/PerspectiveFields/}}
\blfootnote{* Work partially done during internship at Adobe.}
\end{abstract}

%% file: sec/1_introduction.tex
\section{Introduction}

\label{sec:intro}
Take a look at the left-most photo in the teaser~(\Fig{teaser}-A). 
Can you tell if the photo is captured from an everyday camera and if it has been geometrically edited?
The horizon location at the bottom of the image and the parallel vertical lines of the buildings do not follow a typical camera model: 
The horizon at the bottom of the image indicates the camera was tilted up (pitch $\neq$ 0), 
but this would instead produce converging vertical lines in the image due to perspective projection (\Fig{teaser}-B). 
Alternatively suppose the camera has 0 pitch, preserving the vertical lines of the buildings, 
the horizon line would instead be in the middle of the image (\Fig{teaser}-C). 
This contradiction is explained by the shift of the photo, yielding a non-center principal point, 
and breaking a usual assumption of many camera calibration systems. 

Many single-image camera calibration works make use of a simplified pinhole camera model~\cite{hartley2003multiple} that 
assumes a centered principal point~\cite{hold2018perceptual,lee2021ctrl,bogdan2018deepcalib} 
and is parameterized by extrinsic properties such as roll, pitch, and intrinsic properties such as field of view.
However, estimating the calibration of a camera is challenging for images in the wild since they are captured by various types of cameras and lenses. Moreover, like the example in Fig.~\ref{fig:teaser}, the images are often cropped~\cite{chen2017quantitative} or warped for aesthetic composition, which may shift the image center.

In this work, we propose \textit{Perspective Fields}, an over-parameterized per-pixel image-based camera representation. 
Perspective Fields consist of per-pixel Up-vectors and Latitude values that 
are useful on their own for alignment and can be converted to calibration parameters easily by solving a simple inverse problem. 
The Up-vector gives the world-coordinate up direction at each pixel, 
which equals the inverse gravity direction of the 3D scene projected onto the image.
The Latitude is the angle between the incoming light ray and the horizontal plane (see Fig.~\ref{fig:teaser}-D).
This enables our method to be robust to cropping and we show results on multiple camera projection models.

Perspective Fields have a strong correlation with local image features.
For example, the Up-vectors can be inferred by vertical edges in the image, and the Latitude is 0 at the horizon, positive above, and negative below.
Since Perspective Fields have this translation-equivariance property, they are especially well suited to prediction by convolutional neural networks.
We train a neural network to predict Perspective Fields from a single image by extracting crops from 360$^\circ$
panoramas where ground truth supervision can be easily obtained (see Fig.~\ref{fig:method}).
We also use a teacher-student distillation method to transfer Perspective Fields to object-cutouts, which lets us train models to predict Perspective Fields for object-centric images.

For applications that require traditional camera parameters (\eg roll, pitch, field of view and principal point), we propose ParamNet to efficiently derive camera parameters from Perspective Fields. Our method works on image crops and outperforms existing methods in single image camera parameter estimation. 
In addition, Perspective Fields can be used in image compositing to align the camera view between a foreground object and the background based on a local Perspective Field matching metric.
We show with a user study that this metric for view alignment more closely matches human perspective than existing camera models.

Our contributions are summarized as follows.
\begin{itemize}[noitemsep,topsep=0pt]
\item We propose Perspective Fields, a local and non-parametric representation of images with no assumption of camera projection models. 
\item We train a network to predict Perspective Fields that works on both scene-level and object-centric images, and we propose ParamNet to efficiently derive camera parameters from Perspective Fields. Our Perspective Fields achieve better performance on recovering camera parameters than existing approaches. On cropped images, we reduce the pitch error by 40\% over~\cite{lee2021ctrl}.

\item We propose a metric of Perspective Fields to estimate the low-level perspective consistency between two images. 
We show that this consistency measure is stronger in correlation with human perception of perspective mismatch than previous metrics such as Horizon line~\cite{workman2016horizon,hold2018perceptual}.
\end{itemize}

\input{fig/method}

%% file: fig/method.tex
\begin{figure*}[t]
\vspace{-2em}
\setlength{\belowcaptionskip}{-8pt}
\centering
\includegraphics[width=\linewidth]{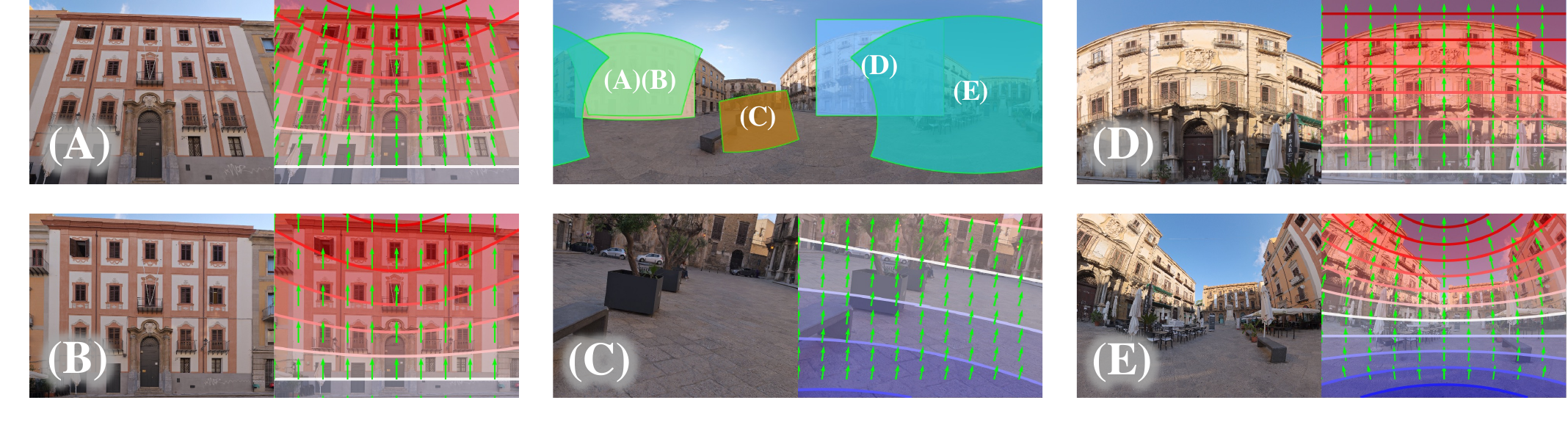}
\caption{
Example ground truth Perspective Fields for different camera parameters. 
Image (A) - (E) are generated from the 360$^\circ$ panorama (middle top). 
Image (A, B, C) is perspective projection (Up-vectors point to vertical vanishing point, Horizon is a straight line at Latitude $0^{\circ}$, and (B) has a shifted principal point to preserve parallel lines.
(D) is a rectangular crop from the equirectangular input (Up-vectors point vertically) and (E) has radial distortion~\cite{barreto2006unifying, mei2007single, bogdan2018deepcalib}.
For each view, we visualize the Up-vector field in green arrows and the Latitude field using a blue-red color map with contour lines.
Latitude colormap: $-\pi/2$ \includegraphics[width=0.4in,height=8pt]{fig/seismic.png} $\pi/2$.}
\label{fig:method}
\end{figure*}

%% file: sec/2_related.tex
\section{Related Work}
\label{sec:related}
\bfpar{Calibration for perspective images.} 
Most calibration methods aimed at consumer cameras assume a pinhole camera model~\cite{hartley2003multiple} to estimate both its intrinsics and extrinsics.
Traditional camera calibration processes require a reference object like chessboards or planar grids
~\cite{slama1980manual, atkinson1996close, gremban1988geometric, grossberg2001general, 
champleboux1992accurate, flexCamCal, sturm1999plane, heikkila1997four, chen2004camera, 4209702}, 
or multiple images~\cite{furukawa2009accurate,strecha2008benchmarking,hartley2003multiple}. 
Other methods strongly rely on the Manhattan world assumption to estimate camera parameters via vanishing points 
\cite{caprile1990using, beardsley1992camera, becker1995semiautomatic,  hartley2003multiple, rother2002new, melo2013unsupervised,lee2014upright}.
Recently, deep learning methods directly predict camera parameters from single images, including horizon line~\cite{workman2016horizon} and 
focal length~\cite{workman2015deepfocal}. 
Hold-Geoffroy \etal~\cite{hold2018perceptual} further extend a CNN to simultaneously predict camera roll, pitch, and FoV. 
UprightNet~\cite{xian2019uprightnet} predicts 3D surface geometry to optimize for camera rotation. 
A few works~\cite{zhai2016detecting,lee2021ctrl,lee2020neural} combine learned features with detected vanishing points to improve performance. 
However, these methods are limited to perspective images with a centered principal point and often do not work on images in the wild where the centered pinhole assumption does not hold due to cropping, warping, or other similar edits. 

\bfpar{Calibration for non-pinhole camera models.}
Besides the common pinhole camera model, prior works have proposed different non-linear models such as Brown-Conrady for 
small distortions~\cite{duane1971close}, 
the division model~\cite{fitzgibbon2001simultaneous} for fisheye cameras, and the unified spherical model~\cite{geyer2000unifying, barreto2006unifying, mei2007single}.
Assuming certain distortion models, learning-based methods can recover focal length and distortion parameter 
\cite{Antunes_2017_CVPR,bogdan2018deepcalib,li2019blind,lopez2019deep}. 
With a known 3D shape and its correspondences, \cite{camposeco2015non, pan2022camera} can recover lens distortions. 
Instead of relying on a specific lens model, we propose a generic representation that stores the up and latitude information for each pixel. 
This local representation encompasses multiple camera projection models. 
Our versatile Perspective Field can be used to recover the parameters of a specific model, if desired.

\bfpar{Perspective aware object placement.}
Many works aim to automate the image compositing process by directly learning to match lighting, scale, etc. \cite{zhu2020single, wang2020vplnet, lee2018context, zhang2020learning}. 
To plausibly composite an object in a background image, one can match their camera parameters. One way to achieve this is to match the horizon lines between two images \cite{lalonde2007photo, hold2018perceptual}. All these methods share the same limitations as the perspective image calibration methods due to their assumptions.

%% file: sec/3_method.tex
\section{Method}
We first define Perspective Fields and show some examples on various images. 
Then we show how we train a network to recover Perspective Fields from a single image.
Finally, we demonstrate some downstream applications that Perspective Fields enable, including camera parameter recovery, image compositing, and object cutout calibration. 

\subsection{Definition of Perspective Fields}
Each pixel $\xB\in\mathbb{R}^2$ on the image frame is originated from a light ray $\RB\in\mathbb{R}^3$ emitted from a 3D point in the world frame $\XB\in\mathbb{R}^3$.
When the ray travels through the camera, it is bent by the lens and projected onto the image frame. 
We assume an arbitrary projection function $\xB=\mathcal{P}(\XB)$ that maps a point in the world to the image plane. 
We denote the gravity direction in the world frame to be a unit vector $\gB$. 
For each pixel location $\xB$, a Perspective Field consists of a unit Up-vector $\uB_\xB$ and Latitude $\varphi_\xB$. 
The Up-vector $\uB_\xB$ is the projection of the up direction of $\XB$, or
\begin{equation}
    \uB_\xB = \lim_{c \to 0} \frac{\mathcal{P}(\XB-c\gB) - \mathcal{P}(\XB)}{||\mathcal{P}(\XB-c\gB) - \mathcal{P}(\XB)||_2}
    \label{eq:up}
\end{equation}
The limit is not required for perspective projection since it preserves straight lines. 
The Latitude $\varphi_\xB$ of this pixel is defined as the angle between the ray $\RB$ and the horizontal plane, or
\begin{equation}
    \varphi_\xB = \mathrm{arcsin}\left(\frac{\RB\cdot \gB}{||\RB||_2}\right).
    \label{eq:lat}
\end{equation}

This representation is applicable to arbitrary camera models. 
In Fig.~\ref{fig:method}, we illustrate the Perspective Field representation of images captured from commonly used cameras
extracted from a \href{https://polyhaven.com/a/piazza_bologni}{360$^\circ$ panorama}.
Although our representation is general, we mainly focus on perspective projection to compare with existing works and leave extensive applications to other camera models for future work.

\input{fig/network_arch}
\subsection{Estimating Perspective Fields}
Our goal is to train a neural network to estimate Perspective Fields from a single image. 
To do this, we introduce PerspectiveNet (Fig.~\ref{fig:networkarch} left), an end-to-end network that takes a single RGB image as input and outputs a per-pixel value for Up-vector and Latitude. 
Unlike previous camera calibration works where the network outputs a single vector of camera parameters~\cite{hold2018perceptual, lee2021ctrl}, 
the output of our system has the same dimension as the input, making it amenable to pixel-to-pixel architectures~\cite{ronneberger2015u, badrinarayanan2017segnet, xie2021segformer}. 
We train our PerspectiveNet on crops from 360$^\circ$ panoramas with cross entropy loss $\mathcal{L}_\mathrm{pers}$. (see \Sec{trainingdetails}.)

\bfpar{Camera parameters from Perspective Fields.}
\label{sec:method:camera-param}
When camera parameters from specific models are needed, 
we can recover the camera parameters from Perspective Fields. 
For instance, if we parameterize perspective projection by roll, pitch, and field of view following~\cite{hold2018perceptual, lee2021ctrl}, and optionally the principal point location, we can represent these with a vector $\thetaB$. 
As extracting these parameters requires combining potentially noisy Perspective Field estimates, we extract them by training a neural network named ParamNet that maps the Perspective Fields to the camera parameters, as shown in \Fig{networkarch}. 
This network is trained directly with a sum of $\ell_1$ losses $\mathcal{L}_\mathrm{param} = \Sigma||\theta_{i} - \hat\theta_{i}||_1$.

\bfpar{Perspective Fields as a metric for perspective mismatch.}
Our representation is easy to interpret: the Up-vectors align with structures that are upright, such as trees and vertical lines on buildings; 
the Latitude values align with viewpoint direction: if the top of an upright object is visible.
Therefore, we propose to use Perspective Fields agreement as a measurement for the image compositing quality between a foreground object and background scene.
We propose Perspective Field Discrepancy (PFD), which is defined as the sum of the difference between the Up-vectors and the Latitude values, or
\begin{equation}
    \mathcal{E}_{\mathrm{PFD}} = \lambda\mathrm{arccos}(\uB_1\cdot\uB_2) + (1- \lambda)||l_1 - l_2||_1,
    \label{eq:PFD}
\end{equation}
where $\uB_i$ is the Up-vector and $l_i$ is the Latitude value. 
The weight $\lambda=0.5$ is used in our experiments.  
Both the Up-vector and the Latitude are in an angular space, so we can take a weighted sum of their angular differences. 
We aggregate the metric by averaging the PFD over all the pixels, denoted as APFD.
The experiment in Sec.~\ref{sec:exp:userstudy} shows that the proposed metric strongly correlates with human perception. 

\bfpar{Object cutout calibration.}
Image composition often involves compositing a segmented object with a scene. 
As foreground objects contain little to no background information, camera calibration methods, including our scene level Perspective Field prediction network, fail on such images due to the domain gap between the panorama training data and the real object images (see Table~\ref{tab:dense-obj}).

We can easily train Perspective Fields on objects by taking COCO~\cite{lin2014microsoft} and doing distillation training using our scene level model as a teacher. 
Since the Perspective Fields are stored per-pixel, 
we can crop out an object in the image and its corresponding pseudo ground truth Perspective Field to form a new training pair.

\input{tab/dense}
\subsection{Implementation details}
\label{sec:trainingdetails}
To learn Perspective Fields from single images, we use the architecture of SegFormer~\cite{xie2021segformer} with Mix Transformer-B3 encoder which was originally used for semantic segmentation tasks. 
The transformer based encoder is effective to enforce global consistency in the Perspective Fields. We use two decoder heads to output a per-pixel probability over  discretized Latitude and Up-vector bins. 
We use cross-entropy loss $\mathcal{L}_{pers}=\ell_\mathrm{CE}$, which we empirically found better than regression.
The ParamNet uses ConvNeXt-tiny~\cite{liu2022convnet} to predict a vector of camera parameters trained with $\ell_1$ loss.

%% file: fig/network_arch.tex
\begin{figure*}[t]
    \vspace{-2em}
    \centering
    \includegraphics[width=.95\linewidth]{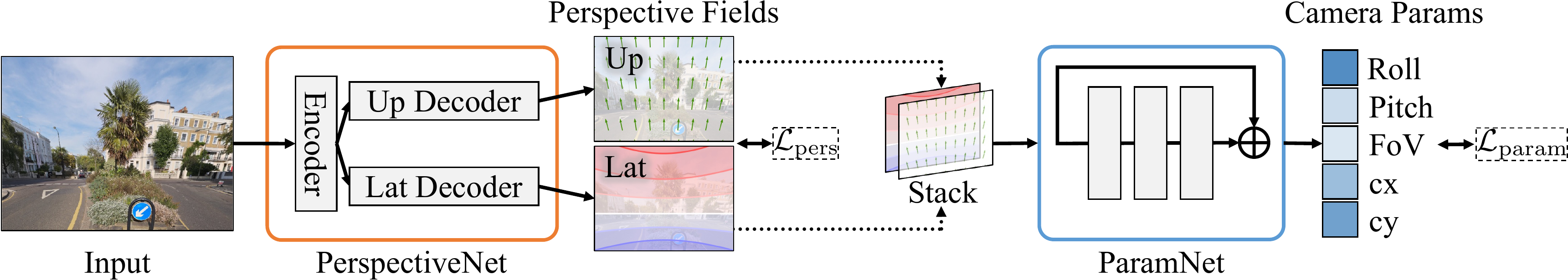}
    \caption{Left: We use a pixel-to-pixel network (\textcolor{msorange}{PerspectiveNet}) to predict Perspective Fields from a single image. 
    Right: When classical camera parameters are needed, we use a ConvNet (\textcolor{msblue}{ParamNet}) to extract this information directly from the Perspective Fields.}
\label{fig:networkarch}
\end{figure*}

%% file: tab/dense.tex
\begin{table*}[t]
\vspace{-2em}
    \centering
        \caption{Quantitative evaluation for scene-level Perspective Field prediction. Perturb:  None on {\it centered} principal point images; Crop on {\it uncentered} principal point images.
        We re-implement Percep.~\cite{hold2018perceptual} using the same backbone and 
        training data as ours. 
        None of the methods have been trained on Stanford2D3D~\cite{armeni2017joint} or TartanAir~\cite{tartanair2020iros}. 
        Results on warped test data and qualitative results are in the supp.
    } %
    \label{tab:dense}
    \resizebox{\textwidth}{!}{ %
    \begin{tabular}{@{}l|c|cccccc|cccccc@{}}
    \toprule
     Dataset & &  \multicolumn{6}{|c}{Stanford2D3D~\cite{armeni2017joint}} & \multicolumn{6}{|c}{TartanAir~\cite{tartanair2020iros}} \\
     \midrule
     & & 
     \multicolumn{3}{|c}{Up ($^o$)} & \multicolumn{3}{c}{Latitude  ($^o$)} & \multicolumn{3}{|c}{Up ($^o$)} & \multicolumn{3}{c}{Latitude  ($^o$)} \\
    Method & Perturb &
    Mean $\downarrow$ & Median $\downarrow$ & $\%<5^o$ $\uparrow$ & 
    Mean $\downarrow$ & Median $\downarrow$ & $\%<5^o$ $\uparrow$ & 
    Mean $\downarrow$ & Median $\downarrow$ & $\%<5^o$ $\uparrow$ & 
    Mean $\downarrow$ & Median $\downarrow$ & $\%<5^o$ $\uparrow$ \\
    \midrule
    Upright~\cite{lee2014upright} & None &
    3.63 & 3.28 & 64.97 & 7.03 & 7.03 & 41.12 &
    3.53 & 3.19 & 65.36 & 5.63 & 5.59 & 49.71 \\
    Percep.~\cite{hold2018perceptual} & None &
    3.58 & 3.32 & 64.19 & 6.27 & 6.07 & 42.36 &
    7.30 & 6.86 & 47.04 & 11.35 & 11.22 & 27.69 \\
    CTRL-C~\cite{lee2021ctrl} & None &
    7.39 & 6.87 & 42.49 & 10.21 & 9.96 & 28.67 &
    8.64 & 7.67 & 41.21 & 9.76 & 10.22 & 28.21 \\
    Ours & None &
    \textbf{2.18} & \textbf{1.88} & \textbf{82.83} & \textbf{3.40} & \textbf{3.06} & \textbf{68.27} &
    \textbf{3.47} & \textbf{2.86} & \textbf{67.45} & \textbf{4.01} & \textbf{3.60} & \textbf{61.73} \\
    \midrule
    Upright~\cite{lee2014upright} & Crop &
    4.49 & 4.19 & 55.58 & 11.43 & 10.93 & 27.37 &
    5.89 & 5.38 & 51.87 & 10.28 & 9.85 & 28.92 \\
    Percep.~\cite{hold2018perceptual} & Crop &
    5.78 & 5.55 & 45.52 & 9.76 & 9.65 & 29.13 &
    5.54 & 5.18 & 51.72 & 9.22 & 8.66 & 30.10 \\
    CTRL-C~\cite{lee2021ctrl} & Crop &
    8.52 & 8.18 & 38.63 & 12.13 & 11.63 & 24.22 &
    7.32 & 6.78 & 43.93 & 9.64 & 9.66 & 27.37 \\
    Ours & Crop &
    \textbf{2.21} & \textbf{1.87} & \textbf{78.80} & \textbf{5.57} & \textbf{5.15} & \textbf{50.36} &
    \textbf{2.81} & \textbf{2.35} & \textbf{71.89} & \textbf{5.73} & \textbf{5.28} & \textbf{50.16} \\

    \bottomrule
    \end{tabular}
    } %

    \end{table*}

%% file: sec/4_experiment.tex
\section{Experiments}
\bfpar{Overview.}
In the following experiments, we study three questions.
First, (Sec.~\ref{sec:exp:dense}), can methods that recover a global set of camera parameters (\eg pitch) produce accurate Perspective Fields.
We verify that {\it directly} producing Perspective Fields produces more accurate camera calibrations, especially on cropped images.
We then ask in Sec.~\ref{sec:exp:pin-hole} the reverse statement: whether our Perspective Field method can be used to recover global camera parameters well. 
We find that our method matches and often outperforms previous methods on images with a centered principal point and substantially outperforms these methods on cropped images. 
Next, we ask whether errors in Perspective Fields match human judgments so that the evaluation in Perspective Field error is meaningful. 
We conduct a user study in Sec.~\ref{sec:exp:userstudy} to evaluate our proposed metric with human perception 
and show that humans are more sensitive to the Perspective Fields discrepancy
than other existing measurements on image perspective. 
We finally show image editing applications from Perspective Fields in Sec.~\ref{sec:exp:application}.

\subsection{Predicting Perspective Fields} 
\label{sec:exp:dense}

We first evaluate our PerspectiveNet on both natural scenes and object-centric images. 

\bfpar{Training data and training details.}
We train our network on a diverse dataset of panorama scenes which includes 30,534 indoor, 51,157 natural and 110,879 street views 
from 360Cities.\footnote{\scriptsize{\url{https://www.360cities.net/}}}
Although we can generate arbitrary types of image projections from the panoramas, we choose to train on perspective images for a fair comparison with previous methods.
To do this, we uniformly sample crops from the panoramas with camera roll in $[-45^{\circ}, 45^{\circ}]$,  pitch in $[-90^{\circ}, 90^{\circ}]$ and FoV in $[30^{\circ}, 120^{\circ}]$.
Our training and validation set consist of 190,830/1,740 panorama images respectively. 
We augment training data with random color jittering, blurring, horizontal flipping, rotation and cropping.
We later show results on other camera models such as fisheye images.

\bfpar{{\it Ours-distill}}: We distill our network on COCO~\cite{lin2014microsoft} images by using pseudo ground truth predicted by our scene level network. 
We crop out the foreground object and the pseudo ground truth to generate the training pairs, and randomly (70\% of the time) remove the background of the image using segmentation masks as data augmentation to generalize to object cutouts.

\input{fig/densewall_objectron}

\bfpar{Test data.}
We test generalization of different methods on publicly available datasets including Stanford2D3D~\cite{armeni2017joint} and TartanAir~\cite{tartanair2020iros} where ground truth camera parameters are available. 
None of the methods compared were trained on the test set.
Stanford2D3D is an indoor panorama dataset where arbitrary camera views can be extracted. 
TartanAir is a photo-realistic dataset captured by drones with extreme viewpoint and diverse scenes (indoor, outdoor, natural, and man-made structures) rendered with different lighting and weather conditions. 
Assuming perspective projection, we uniformly sample 2,415 views from Stanford2D3D with camera roll in $[-45^{\circ}, 45^{\circ}]$, 
pitch in $[-50^{\circ}, 50^{\circ}]$ and FoV in $[30^{\circ}, 120^{\circ}]$. 
For TartanAir, we randomly sample 2,000 images from its test sequences with roll ranging in  $[-20^{\circ}, 20^{\circ}]$, 
pitch in $[-45^{\circ}, 30^{\circ}]$, and fixed FoV (74$^{\circ}$). 
To test the robustness of methods, we add image crop perturbation to the test image, details in supp.

For object-centric test images, we randomly sample 600 views from 6 classes of the Objectron~\cite{objectron2021} test set, and compute foreground cutouts based on the object bounding box with a margin of 20\% box size. 
In some tests, the object is isolated by removing the background using the segmentation mask predicted by PointRend~\cite{kirillov2019pointrend}, which we refer to as ({\it Isolated}). 
We use the camera pose annotation to get the ground truth Perspective Fields with camera roll ranging in $[-45^{\circ}, 45^{\circ}]$, 
pitch in $[-82^{\circ}, -4^{\circ}]$ and FoV in $[46^{\circ}, 53^{\circ}]$.

\bfpar{Baselines.}
The closest task to Perspective Fields prediction is to recover a global set of camera parameters, and then convert them to Perspective Fields using \Eq{up} and \Eq{lat}. 
We compare our method with the following baselines: 
Upright~\cite{lee2014upright}, Perceptual measure~\cite{hold2018perceptual} and CTRL-C~\cite{lee2021ctrl}, among which Upright is the only non-learning based method. 
They all predict camera roll, pitch, and FoV from a single RGB image and assume that the principal point is at the image center. 
From the predicted camera parameters, we calculate their Perspective Fields for evaluation. 
We re-implement~\cite{hold2018perceptual} using the same backbone and train it on our data. 
For Upright and CTRL-C, we use the official model and code for evaluation. 
None of these methods have seen any training data from the test datasets.

\bfpar{Metrics.} 
We calculate the angular error of Up-vector and Latitude fields and report three metrics: the mean error {\it (Mean)}, median error {\it (Med)}, 
and fraction of pixels with error less than a threshold (in our case $5^\circ$). 
For methods that output camera parameters, we convert the predicted parameters to Perspective Fields.

\bfpar{Results on scene images.}
We show the results on Stanford2D3D and TartanAir in \Table{dense}. 
Predicting the Perspective Fields is more effective than recovering camera parameters from previous methods.
On centered principal point images (Perturb: {\it None}), our method outperforms the second best by a large margin.
On shifted principal point images (Perturb: {\it Crop}), (simulating images found in the wild that have undergone cropping), our method has less degradation than previous baselines. 
Our performance on Up-vector prediction is robust to cropping, with comparable numbers (4\% drop in \%$<5^\circ$ of Up on Stanford2D3D). 
Other methods, have large performance drop in both Up-vector and Latitude prediction in this setting. 
Nevertheless, our method outperforms the competing methods on Latitude. 
Visual results can be found for qualitative evaluation in our supplementary material.

\input{tab/object_centric}
\input{fig/wild/wild_good}

\bfpar{Results on object-centric images.}
The results on the Objectron dataset~\cite{objectron2021} are shown in \Table{dense-obj}. 
Our model trained on COCO ({\it Ours-distill}) using the proposed distillation method significantly improves over its teacher model, especially for isolated object images. 
Both our teacher model trained on panorama scene images and \cite{hold2018perceptual} have a big performance drop on isolated object images due to a major data domain gap. 
In contrast, Upright and {\it Ours-distill} only have a mild performance drop. 
It is quite surprising that CTRL-C's accuracy on Latitude field improves on the isolated object images. 
We suspect that the structure in the background might contradict CTRL-C's assumption of a centered principal point, 
as the the cropping will often shift the principal point. 
Our student model ({\it Ours-distill}) achieves overall better accuracy than the baselines. 
Some visual results are shown in Fig.~\ref{fig:densewallobjectroncomparison}.

We also test each method on some challenging in-the-wild web images in Fig.~\ref{fig:wild_merged}. 
These web images may have been cropped or warped for aesthetic composition. 
Camera calibration methods with rigid scene and camera assumptions cannot robustly handle these images. 
Our method tends to provide better estimations. More results can be found in our supplementary material. 
\input{fig/non_perspective/non_perspective}

\input{tab/camera_gsv_crop}
\input{tab/camera_gsv}
\bfpar{Generalization to non-perspective projections.}
In this section, we ask whether we can recover the Perspective Field for images with non perspective properties without explicitly training on them.
We take advantage of the local representation and use a sliding window inference technique for images that are out of our training distribution.
We inference on small crops and aggregate the prediction for each pixel from overlapping windows. Using this technique, 
we show in~\Fig{nonperspective}-{\it Sliding Win.} that, without fine-tuning, the recovered Up-vectors are already tangential to the upward curves 
and the horizon line is curved, which is close to the ground truth.
In {\it Fine-tune}, we show results after fine-tuning on distorted images, \eg~\Fig{method}-(E), which has comparable predictions in Up-vectors
and slightly better predictions in Latitude. 
In~\Fig{nonperspective} row 2, we show results on a challenging multiperspective image from the Inception movie, using the same sliding window technique.
The network is able to pick up the negative Latitude on top of the building and the Up-vector distortion at the top right corner.
In row 3 and 4, we show more results from the PerspectiveNet on art works with non physically plausible cameras.

\subsection{Camera parameter estimation}
\label{sec:exp:pin-hole}
We have shown in Sec.~\ref{sec:exp:dense} that Perspective Fields from predicted camera parameters are less effective than directly predicting them; 
can camera parameters be effectively recovered from Perspective Fields?
In this section, we use the ParamNet in~Sec.~\ref{sec:method:camera-param}~\Fig{networkarch} to recover camera parameters from Perspective Field predictions
and compare with methods that directly predict them~\cite{lee2014upright, hold2018perceptual,lee2021ctrl}.

\bfpar{Setup.}
We test on perspective images and recover roll, pitch, and FoV for uncropped images as well as principal point for cropped images.
All methods are trained and evaluated on Google Street View (GSV)~\cite{anguelov2010google} for fair comparison, following~\cite{lee2021ctrl}. 
Besides the test images used in~\cite{lee2021ctrl}, we generate a more general set of uncentered principal-point images by cropping. 
See Supp for detailed dataset settings.

\bfpar{Metrics.} 
Since FoV is undefined for cropped images, 
we define it (FoV$^*$) as follows: Denote camera pinhole as $O$ and the middle points of the top and bottom edges of the image as $M_1, M_2$. 
FoV$^*$ is the angle between $OM_1$ and $OM_2$. 
The principal point location (cx and cy) is relative to the image size. 
We use $\ell_1$ error between the prediction and ground truth as our metric for all camera parameters (Roll, Pitch, FoV, cx, cy).
We also measure the Up and Latitude errors of the Perspective Fields recovered from the predicted camera parameters using Eq.~\ref{eq:up} and~\ref{eq:lat}
to show the impact of errors from camera parameters on Up and Latitude. 

\bfpar{Results.}
We start with the principal-centered test set that previous methods use, which is a highly constrained setting. 
We report the angle differences of roll, pitch, and FoV in Table \ref{tab:test_gsv}. 
Our method gets comparable camera calibration performance compared to previous methods on principal-point centered images, 
with lower errors in pitch and FoV and comparable median error in roll. 
Our Perspective Field representation is a dense local representation, which leads to a robust way to estimate the global camera parameters. 
The Up-vector field and the Latitude field provide interpretable cues for the estimation of camera roll and pitch respectively. 

We then test on a more general cropped dataset, where images have an uncentered principal point.
As shown in Table~\ref{tab:test_gsv_crop}, our method outperforms all the other baselines by a large margin on all metrics. 
Compared to CTRL-C, we reduce the error on roll (19\%), pitch (40\%), and FoV (31\%), which reflects a large increase in Latitude accuracy by over 13\% and Up-vector accuracy by 2\%. None of the baselines above handles principal point by design, therefore, 
we perform two ablations: 
1) {\it No $\mathcal{L}_{pers}$}: we train without $\mathcal{L}_\mathrm{pers}$ but with $\mathcal{L}_\mathrm{param}$. This is a network that predicts parameters end-to-end. Compared to Perceptural~\cite{hold2018perceptual}, it has the same backbone followed by the ParamNet while additionally predicting the principal point. It differs from our method by lacking the $\mathcal{L}_{pers}$. This is to test whether using Perspective Fields as the intermediate representation is helpful. Results show that it works better than existing methods but is still worse than Ours.
2) {\it No Center Shift}: our ParamNet in Table~\ref{tab:test_gsv} assumes a central principal point and only predicts roll, pitch and FoV. 
We improve the relative principal point shift accuracy by over 36\%. 
By recovering the principal point, we improve Up-vector accuracy ($\%{<}5^\circ$) by ${>}4$\% and Latitude accuracy by ${>}6$\%.

Although our method does not directly learn to predict the camera parameters, our results show that they can be recovered from the Perspective Fields alone without RGB data and still outperform SOTA methods.

\input{tab/metric_corr}

\subsection{User study for perspective matching metrics}
\label{sec:exp:userstudy}
To validate our proposed APFD metric, we conduct a user study to analyze its correlation with human perception of perspective consistency for image compositing. 

Given a background image with known camera settings, we render a 3D object with 10 randomly perturbed cameras and composite it to the background. Example images are in the supp material. 
These 10 images are ranked by participants using a two alternative forced-choice (2AFC) test. 
In this test, two composites are displayed side by side, and a user picks the one that looks better in term of perspective consistency. 
We compute the Pearson's correlation coefficient of APFD \wrt the human ranking scores. 
We receive 18 votes for each image pair and we repeat the experiment on 8 scenes (background-object pairs). 

Table~\ref{tab:user_corr} shows the median correlation scores on the 8 scenes for different metrics. 
See the supplementary material for full statistics and more details of the test.
Our proposed APFD metric has the highest correlation with human perception. 
The Up-vector field error captures the local perspective distortion well, thus providing good performance among individual metrics. 
The APFD metric which combines both Up-vector and Latitude gives a slightly higher correlation score.
Single camera parameter metrics have widely varying correlations. 
Among them, deviation in FoV is a poor indicator of human perception, also shown in~\cite{hold2018perceptual}. 
The change in pitch is a dominant factor in perspective mismatch. 
Summing the parameter difference ({\it Camera-All}) does not improve correlation scores, which shows the difficulty of using camera parameters to measure perceived perspective consistency.
The horizon line used in~\cite{hold2018perceptual} performs comparably with our Latitude field metric, since they measure similar quantities, 
however horizon lines are not always visible in images.

\subsection{Image editing applications}
We conclude with applications of  Perspective Fields.
\input{fig/application-retrieval/retrieval-demo}
\label{sec:exp:application}
\bfpar{Perspective-aware image recommendation.}
Perspective Fields can guide the retrieval of images from a database of 2D images. 
Our method works on images with extreme viewpoints, while horizon-based perspective matching methods like \cite{hold2018perceptual} fail on this type of image because the horizon is far outside the image. 
We demonstrate this in \Fig{retrieval-demo}, where we estimate  Perspective Fields on 10 images of hot air balloons with diverse view angles. 
We use our Perspective Field metric to retrieve the best balloon sprite based on the bounding box given by the user (yellow and green boxes). 
The system calculates the APFD error (\Eq{PFD}) between the background and foreground fields, and adjusts the best candidate with a similarity transformation to better align the Perspective Fields with the background. 
The left columns rank the balloon sprites by error from low to high. 

\bfpar{AR effect.}  
Our Perspective Fields can be used in AR effect applications related to gravity \eg simulating snows, hanging a chandelier to the ceiling, etc. 
In~\Fig{rain}, we demonstrate rain effect using the Up-vector prediction. Our compositing {\it w/ Up} which considers the Up-vector prediction
looks more natural than vertical raindrops in the image frame {\it w/o Up}.

\input{fig/pf_rain/rain.tex}

\input{fig/composite_idealcity/idealcity}
\bfpar{3D object insertion.}
Our Perspective Fields can be used to achieve better compositing.
In \Fig{idealcity}, we render 3D models in a renaissance painting. The painting does not follow the centered pinhole assumptions 
of past methods~\cite{hold2018perceptual,lee2021ctrl}. 
On the left, we use camera parameters from~\cite{hold2018perceptual} to render and insert the 3D objects.
The lamp looks off since the top seems to be tilted. 
This is because it predicts pitch $=7^\circ$ that matches the horizon line, causing distortion in the up direction in that region.
On the right, we use our Perspective Fields to estimate the camera view and correctly maintain the perspective consistency for the objects.

%% file: fig/densewall_objectron.tex
\begin{figure}[!t]
    \centering
    \scriptsize
    \setlength{\tabcolsep}{1pt}
    \begin{tabular}{ccccccc}

\frame{\includegraphics[width=0.16\linewidth]{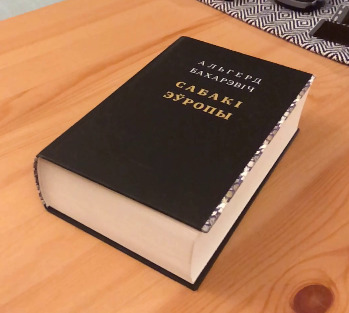}} &
\frame{\includegraphics[width=0.16\linewidth]{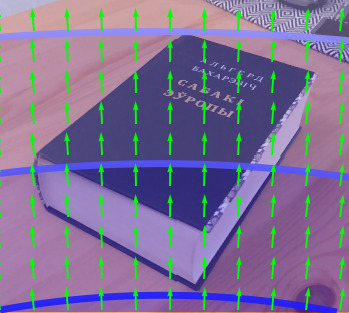}} &
\frame{\includegraphics[width=0.16\linewidth]{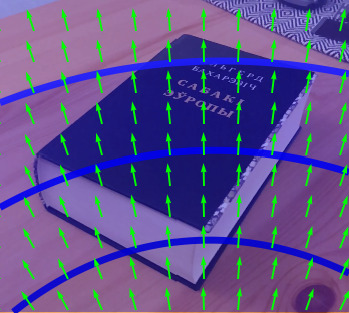}} &
\frame{\includegraphics[width=0.16\linewidth]{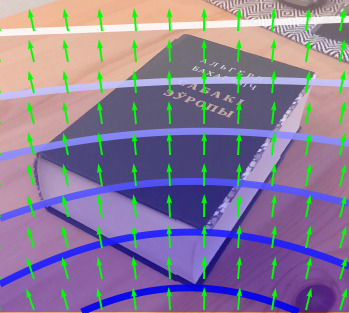}} &
\frame{\includegraphics[width=0.16\linewidth]{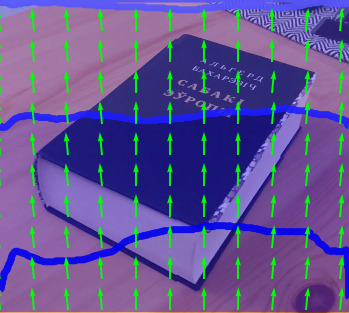}} &
\frame{\includegraphics[width=0.16\linewidth]{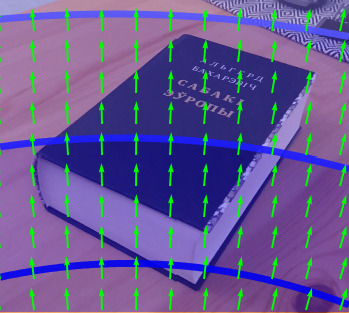}} \\
\frame{\includegraphics[width=0.16\linewidth]{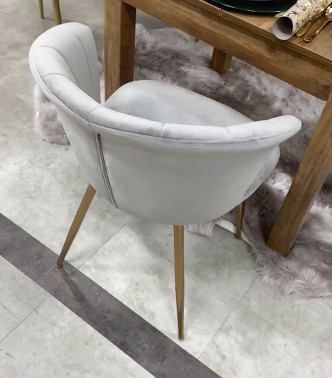}} &
\frame{\includegraphics[width=0.16\linewidth]{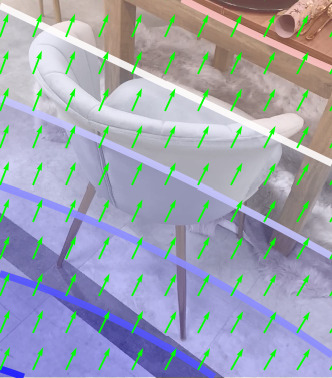}} &
\frame{\includegraphics[width=0.16\linewidth]{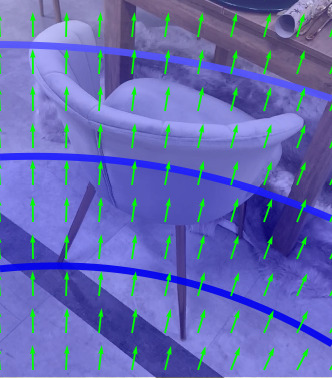}} &
\frame{\includegraphics[width=0.16\linewidth]{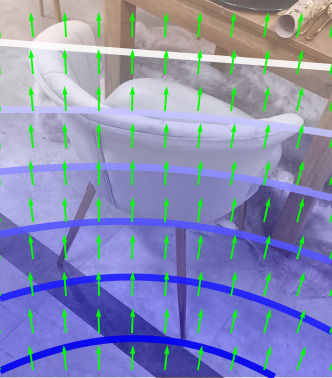}} &
\frame{\includegraphics[width=0.16\linewidth]{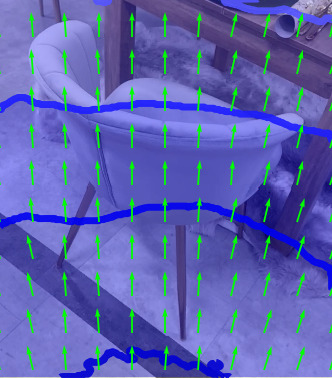}} &
\frame{\includegraphics[width=0.16\linewidth]{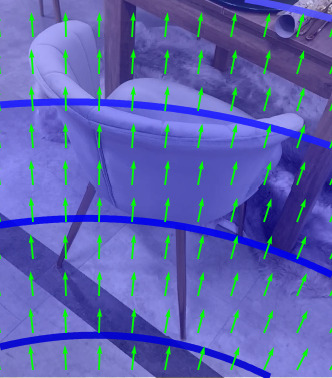}} \\
\frame{\includegraphics[width=0.16\linewidth]{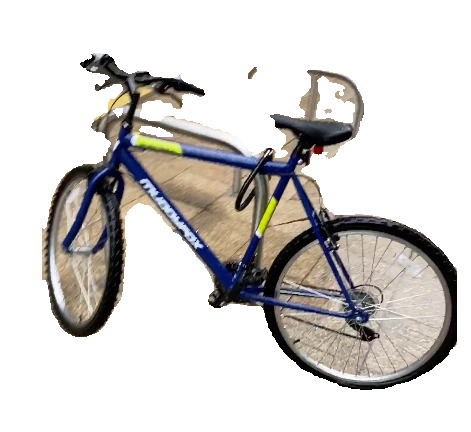}} &
\frame{\includegraphics[width=0.16\linewidth]{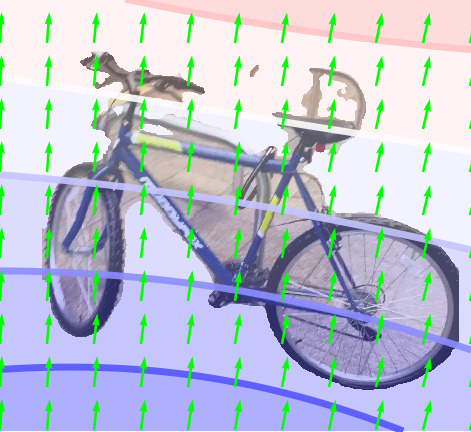}} &
\frame{\includegraphics[width=0.16\linewidth]{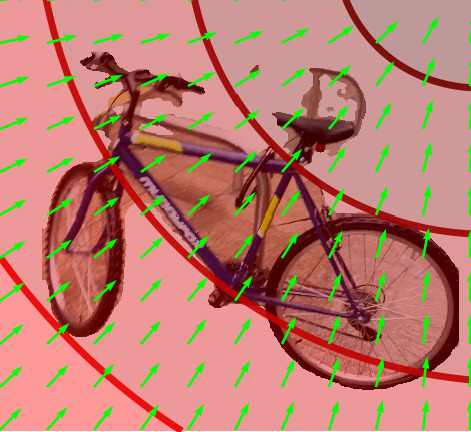}} &
\frame{\includegraphics[width=0.16\linewidth]{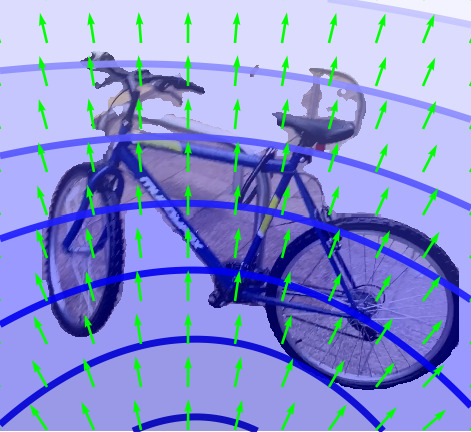}} &
\frame{\includegraphics[width=0.16\linewidth]{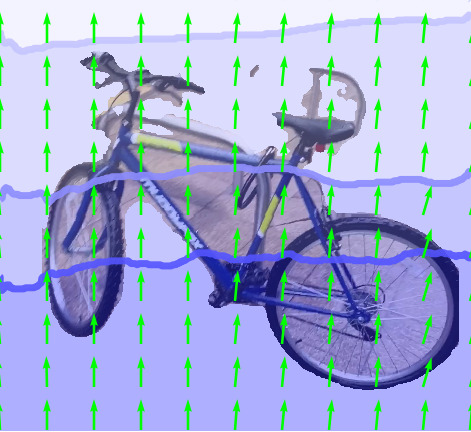}} &
\frame{\includegraphics[width=0.16\linewidth]{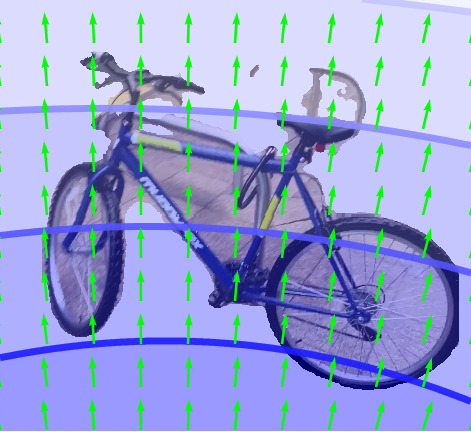}} \\
\frame{\includegraphics[width=0.16\linewidth]{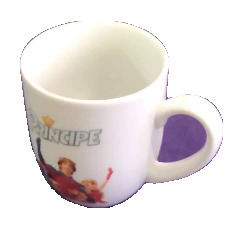}} &
\frame{\includegraphics[width=0.16\linewidth]{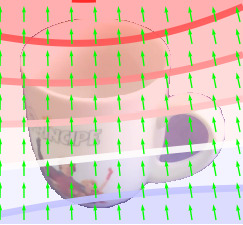}} &
\frame{\includegraphics[width=0.16\linewidth]{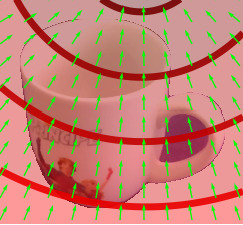}} &
\frame{\includegraphics[width=0.16\linewidth]{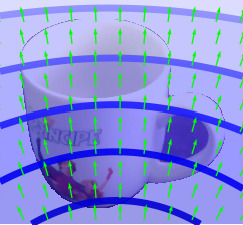}} &
\frame{\includegraphics[width=0.16\linewidth]{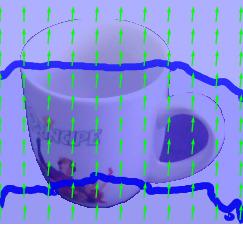}} &
\frame{\includegraphics[width=0.16\linewidth]{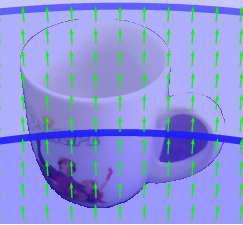}} \\
Input & Upright~\cite{lee2014upright} & Percep.~\cite{hold2018perceptual} & CTRL-C~\cite{lee2021ctrl} & Ours-Distill & GT \\    
    \end{tabular}
    \caption{Qualitative results on Objectron~\cite{objectron2021}. The top two rows show the results on the original image crops. The bottom row shows the results on isolated object images. Upright~\cite{lee2014upright} and Perceptual~\cite{hold2018perceptual} often fail dramatically on these images. 
    Up-vectors are shown in the green vectors. Latitude are visualized by colormap: $-\pi/2$ \includegraphics[width=0.4in,height=8pt]{fig/seismic.png} $\pi/2$.}
    \label{fig:densewallobjectroncomparison}
    \vspace{-2em}
\end{figure}

%% file: tab/object_centric.tex
\begin{table}[t!]

\vspace{-2em}
    \centering
    \caption{Quantitative evaluation for object-centric prediction. None of the compared methods have been trained on Objectron.
    } %
    
    \resizebox{\linewidth}{!}{ %
    \begin{tabular}{@{}l|c|cccccc@{}}
    \toprule
     Dataset & & \multicolumn{6}{c}{Objectron~\cite{objectron2021}}  \\
     \midrule
     & & \multicolumn{3}{c}{Up ($^o$)} & \multicolumn{3}{c}{Latitude  ($^o$)} \\
    Method & Perturb & 
    Mean $\downarrow$ & Median $\downarrow$ & $\%<5^o$ $\uparrow$ & 
    Mean $\downarrow$ & Median $\downarrow$ & $\%<5^o$ $\uparrow$ \\
    \midrule
    Upright~\cite{lee2014upright} & crop &
    7.57 & 7.03 & 44.31 & 22.59 & 22.20 & 18.49 \\
    Percep.~\cite{hold2018perceptual} & crop &
    7.85 & 7.21 & 39.39 & 11.60 & 11.69 & 22.97 \\
    CTRL-C~\cite{lee2021ctrl} & crop &
    7.50 & 7.09 & 40.02 & 20.93 & 21.00 & 11.26 \\
    \midrule
    Ours & crop &
    4.96 & 4.42 & 53.90 & 8.49 & 8.01 & 32.31 \\
    Ours-distill & crop &
    \textbf{4.19} & \textbf{3.76} & \textbf{57.71} & \textbf{7.71} & \textbf{7.57} & \textbf{33.54} \\
    \midrule
    \midrule
    Upright~\cite{lee2014upright} & isolated &
    8.14 & 7.71 & 41.45 & 28.49 & 28.38 & 12.05 \\
    Percep.~\cite{hold2018perceptual} & isolated &
    38.70 & 31.63 & 11.27 & 99.35 & 100.32 & 1.05 \\
    CTRL-C~\cite{lee2021ctrl} & isolated &
    7.49 & 7.13 & 39.38 & 9.87 & 9.85 & \textbf{27.32} \\
    \midrule
    Ours & isolated &
    17.24 & 11.30 & 33.96 & 84.61 & 84.69 & 0.49 \\
    Ours-distill & isolated &
    \textbf{4.45} & \textbf{4.12} & \textbf{54.88} & \textbf{9.65} & \textbf{9.56} & 25.82 \\
    \bottomrule
    \end{tabular}
    } %
    
    \label{tab:dense-obj}
    \end{table}

%% file: fig/wild/wild_good.tex
\begin{figure}[!t]
    \centering
    \scriptsize
    \setlength{\tabcolsep}{1pt}
    \begin{tabular}{ccccc}
    
    \frame{\includegraphics[width=0.19\linewidth]{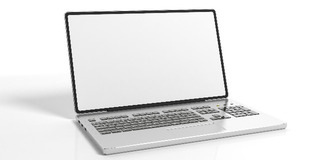}} &
    \frame{\includegraphics[width=0.19\linewidth]{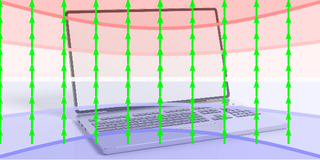}} &
    \frame{\includegraphics[width=0.19\linewidth]{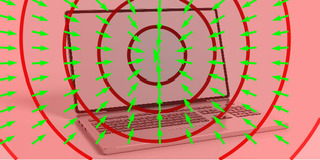}} &
    \frame{\includegraphics[width=0.19\linewidth]{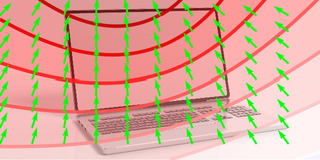}} &
    \frame{\includegraphics[width=0.19\linewidth]{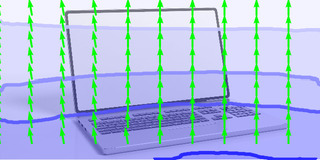}} \\

    \frame{\includegraphics[width=0.19\linewidth]{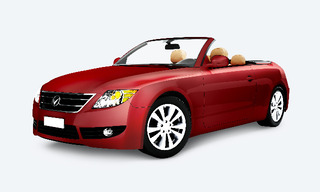}} &
    \frame{\includegraphics[width=0.19\linewidth]{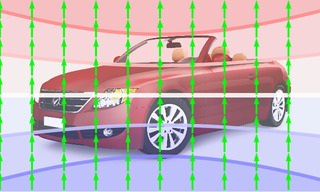}} &
    \frame{\includegraphics[width=0.19\linewidth]{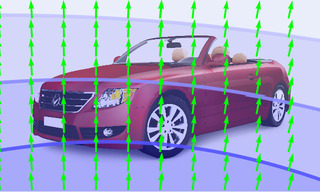}} &
    \frame{\includegraphics[width=0.19\linewidth]{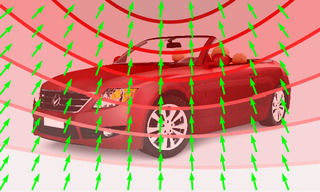}} &
    \frame{\includegraphics[width=0.19\linewidth]{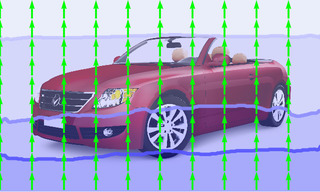}} \\

    \frame{\includegraphics[width=0.19\linewidth]{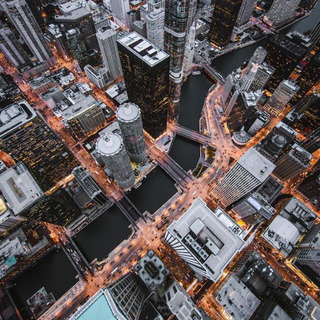}} &
    \frame{\includegraphics[width=0.19\linewidth]{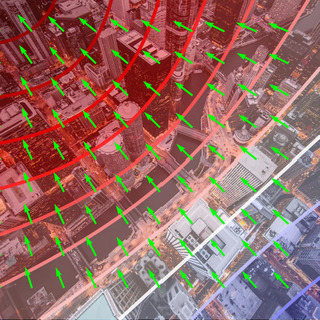}} &
    \frame{\includegraphics[width=0.19\linewidth]{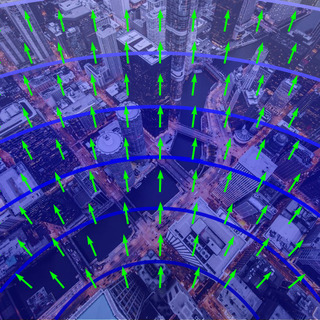}} &
    \frame{\includegraphics[width=0.19\linewidth]{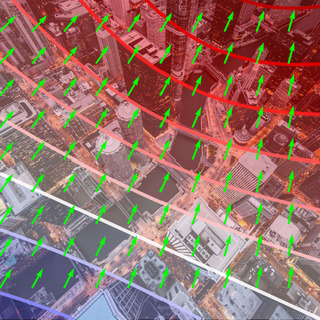}} &
    \frame{\includegraphics[width=0.19\linewidth]{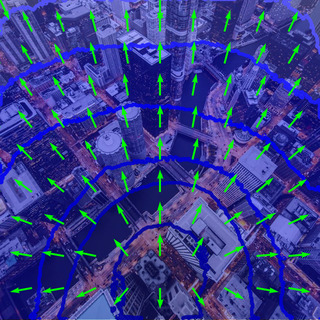}} \\
    \frame{\includegraphics[width=0.19\linewidth]{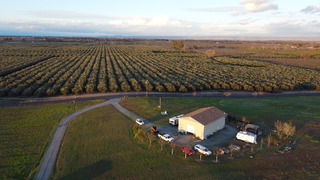}} &
    \frame{\includegraphics[width=0.19\linewidth]{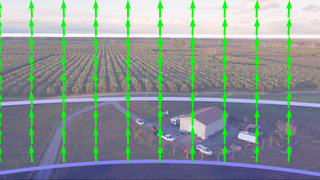}} &
    \frame{\includegraphics[width=0.19\linewidth]{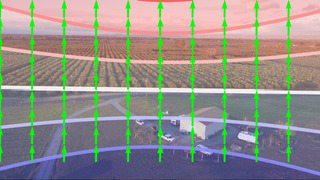}} &
    \frame{\includegraphics[width=0.19\linewidth]{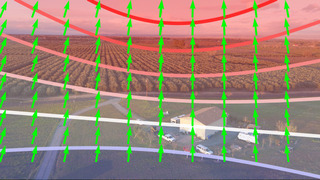}} &
    \frame{\includegraphics[width=0.19\linewidth]{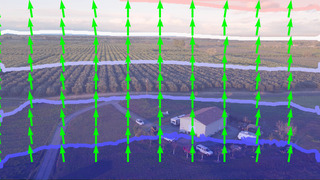}} \\
    Input & Upright~\cite{lee2014upright} & Percep.~\cite{hold2018perceptual} & CTRL-C~\cite{lee2021ctrl} & Ours \\
    \end{tabular}
    \caption{Qualitative results on web images. Our approach produces better results compared to~\cite{lee2014upright}, \cite{hold2018perceptual}, 
    and~\cite{lee2021ctrl}. There is no ground truth available, see Supp on how to infer the GT horizon line for the laptop example. 
    }
    \label{fig:wild_merged}
\end{figure}

%% file: fig/non_perspective/non_perspective.tex
\begin{figure}[!t]

\vspace{-2em}
    \setlength{\tabcolsep}{1pt}
    \centering
\includegraphics[width=\linewidth]{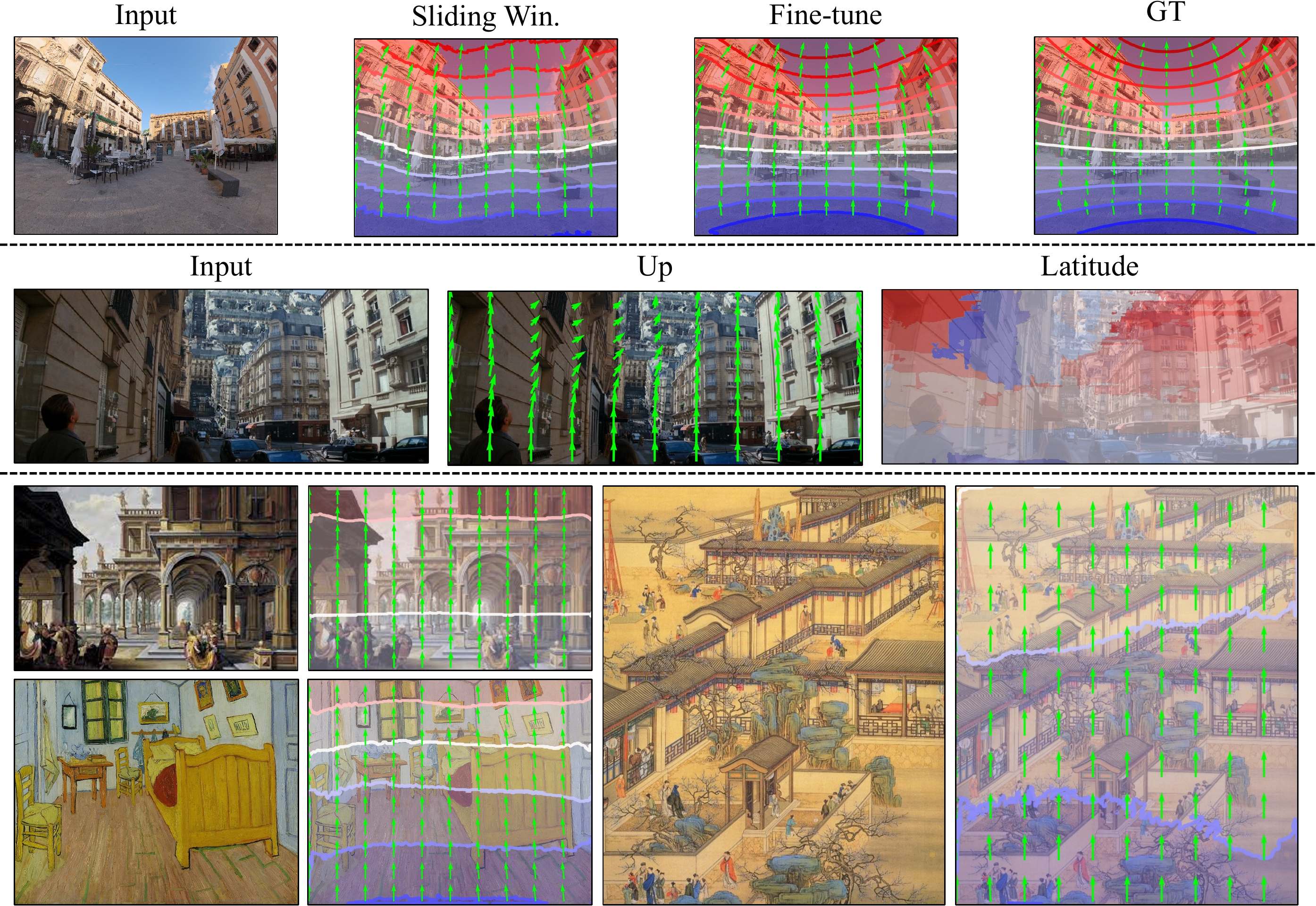}
    \caption{
        Generalization to non-pinhole images. (1st row) Fisheye images (top) are unseen during training. We show results by computing inference on small crops with a sliding window, or fine-tuning the network on fisheye images.
        (2nd row) A screen shot from the movie Inception shows our method identifies the correct distortion at the top right corner and negative Latitude (in Blue) on top of the building. (4th row) More results on artworks with various camera models.
    }
    \label{fig:nonperspective}
    \vspace{-1em}
\end{figure}

%% file: tab/camera_gsv_crop.tex
\begin{table*}[t!]
    \vspace{-2em}
    \renewcommand{\arraystretch}{0.9}
    \caption{GSV {\it uncentered} principal-point results. 
    Our method recovers the principal point and outperforms the baselines on all metrics.}
    \vspace{-1em}
    \label{tab:test_gsv_crop}
    \resizebox{\linewidth}{!}{ %

    \begin{tabular}{@{}l|cc|cc|cc|cc|cc|cc|cc@{}}
    \toprule
    Method & \multicolumn{2}{c|}{Roll ($^\circ$) $\downarrow$} & \multicolumn{2}{c|}{Pitch ($^\circ$) $\downarrow$} & \multicolumn{2}{c|}{FoV$^*$ ($^\circ$) $\downarrow$} &\multicolumn{2}{c|}{cx $\downarrow$} &\multicolumn{2}{c|}{cy $\downarrow$} &\multicolumn{2}{c|}{Up($^\circ$) } &\multicolumn{2}{c}{Latitude($^\circ$)} \\
      & Mean & Med. & Mean & Med. & Mean & Med. & Mean & Med. & Mean & Med.  & Med. $\downarrow$& $\%<5^o$$\uparrow$ & Med. $\downarrow$& $\%<5^o$ $\uparrow$\\ 
    \midrule
    Upright~\cite{lee2014upright}           & 2.73 & 1.55 & 7.23 & 4.98 & 10.50 & 7.67 & - & - & - & - &  1.83 & 74.57 & 6.32 & 41.08 \\
    Perceptual~\cite{hold2018perceptual}    &  2.39 & 1.45 & 5.24 & 4.05 & 8.47 & 7.22 & - & - & - & - & 1.70 & 93.15 & 3.58 & 70.35 \\
    CTRL-C~\cite{lee2021ctrl}               &  1.92 & 1.21 & 4.51 & 3.64 & 5.57 & 4.66 & - & - & - & - & 1.29 & 96.84 & 2.98 & 75.70 \\
    Ours                         &  \textbf{1.37} & \textbf{0.97} & \textbf{2.60} & \textbf{2.14} & \textbf{3.75} & \textbf{3.19} & \textbf{0.09} & \textbf{0.07} & \textbf{0.08} & \textbf{0.06} & \textbf{1.05} & \textbf{98.95} & \textbf{2.17}  & \textbf{89.47} \\
    \midrule
    No $\mathcal{L}_{pers}$                     & 1.68 & 1.28 & 2.88 & 2.33 & 3.95 & 3.25 & \textbf{0.09} & \textbf{0.07} & \textbf{0.08} & 0.07 & 1.44 & 95.94 & 2.37 & 84.20\\
    No Center Shift                         & 1.98 & 1.19 & 4.23 & 3.58 & 6.18 & 4.82 & 0.13 & 0.11 & 0.12 & 0.11 & 1.12 & 94.88 & 2.69 & 82.77 \\
    \bottomrule
\end{tabular}
} %
\end{table*}

%% file: tab/camera_gsv.tex
\begin{table}[t!]
    \renewcommand{\arraystretch}{0.9}
    \caption{Results on GSV {\it centered} principal-point images.
    Our method has comparable performance to the baselines. }
    \vspace{-1em}
    \label{tab:test_gsv}
    \resizebox{\linewidth}{!}{ %
    \begin{tabular}{@{}l|cc|cc|cc@{}}
    \toprule
    Method & \multicolumn{2}{c|}{Roll ($^\circ$) $\downarrow$} & \multicolumn{2}{c|}{Pitch ($^\circ$) $\downarrow$} & \multicolumn{2}{c}{FoV ($^\circ$) $\downarrow$}  \\
      & Mean & Med. & Mean & Med. & Mean & Med. \\ 
    \midrule
    Upright~\cite{lee2014upright}           &  6.19 & \textbf{0.43} &  2.90 &  1.80 & 9.47 &  4.42  \\
    Perceptual~\cite{hold2018perceptual}    &   0.94 & 0.67 & 2.24 & 1.81 & 4.37 & 3.58 \\
    CTRL-C~\cite{lee2021ctrl}               & \textbf{0.66} & 0.53 & 1.58 & 1.31 & 3.59 & 2.72  \\
    Ours                                    & \textbf{0.66} & 0.52 & \textbf{1.36} & \textbf{1.18} & \textbf{3.07} & \textbf{2.33} \\
    \bottomrule
\end{tabular}
} %
\end{table}

%% file: tab/metric_corr.tex
\begin{table}[t!]
    \caption{Pearson's correlation for different metrics \emph{\wrt} human perception.  
    Our APFD metric has strongest correlation with human perception. 
    More statistics and visual results are in the supp. }%
    \label{tab:user_corr}
    \resizebox{\linewidth}{!}{ %
    \begin{tabular}{cccccccccc}
        \toprule
 & Camera-All & Roll & Pitch & FoV & Prin. Point & Horizon & Lati & Up & APFD \\
 \midrule
Median & 0.59 & 0.21 & 0.73 & -0.08 & 0.49 & 0.71 & 0.65 & 0.80 & \textbf{0.87} \\
\bottomrule
\end{tabular}
} %
\end{table}

%% file: fig/application-retrieval/retrieval-demo.tex
\begin{figure}[t]
\vspace{-2em}
\centering
\includegraphics[width=\linewidth]{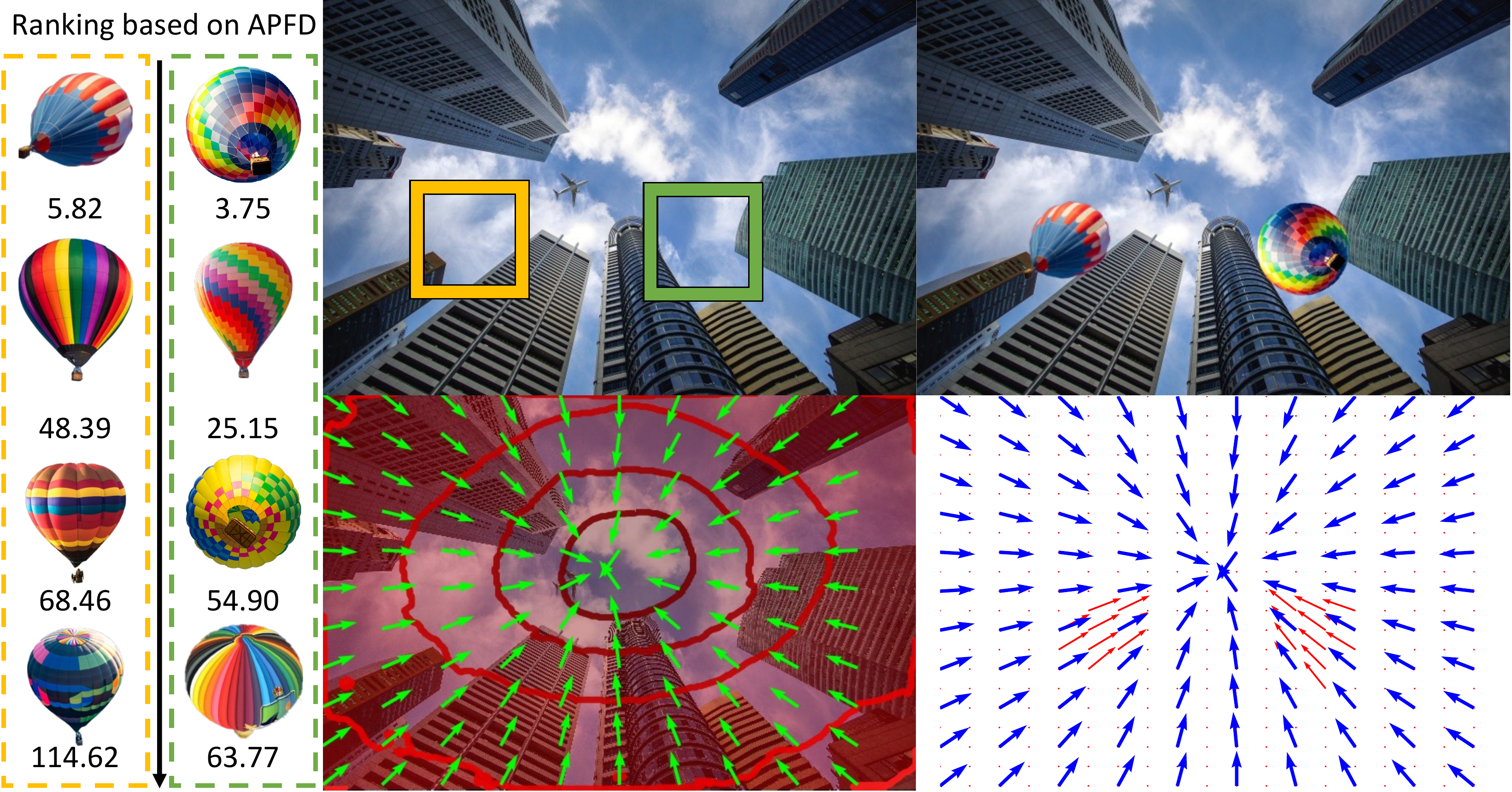}
\caption{
Given selected locations on a skyscraper image, our system computes the local Perspective Field for the background and 
retrieves foreground objects from a set of air balloons that best match the predicted fields.  
\textbf{Left:} ranking of air balloon images for two insertion locations, APFD error shown at the bottom. 
\textbf{Middle:} background image with two boxes as the insertion locations (top) and predicted Perspective Fields (bottom). 
\textbf{Right}: image composition with 2D rotation adjustment of foreground sprites (top) and visualization of Up-vector fields after compositing (bottom).}
\label{fig:retrieval-demo}
\end{figure}

%% file: fig/pf_rain/rain.tex
\begin{figure}[!t]

\vspace{-2em}
    \centering
    \scriptsize
    \setlength{\tabcolsep}{1pt}
    \begin{tabular}{cccc}

\frame{\includegraphics[width=0.24\linewidth]{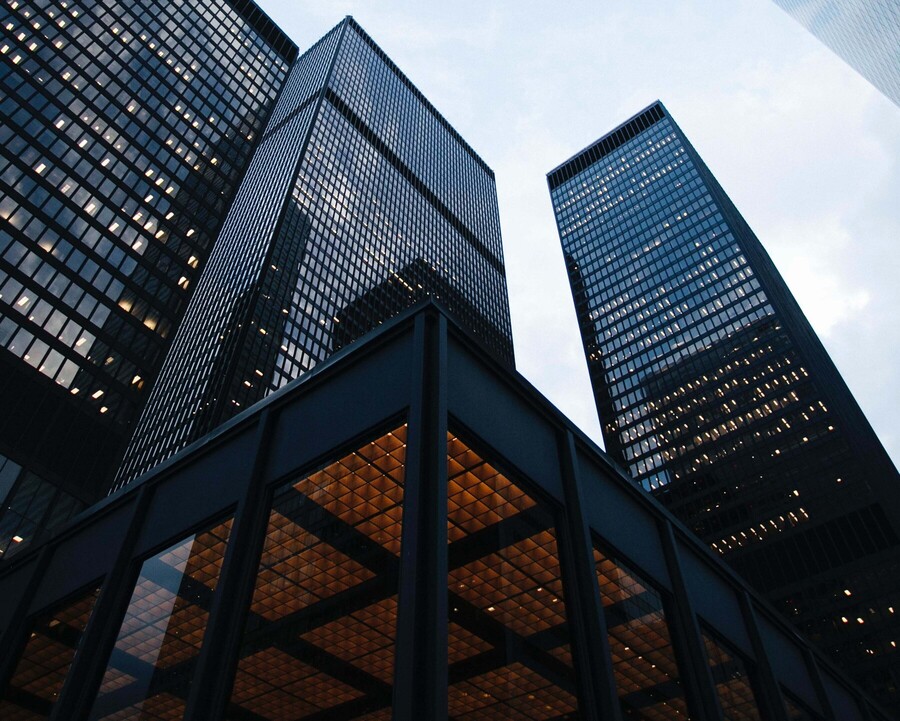}} &
\frame{\includegraphics[width=0.24\linewidth]{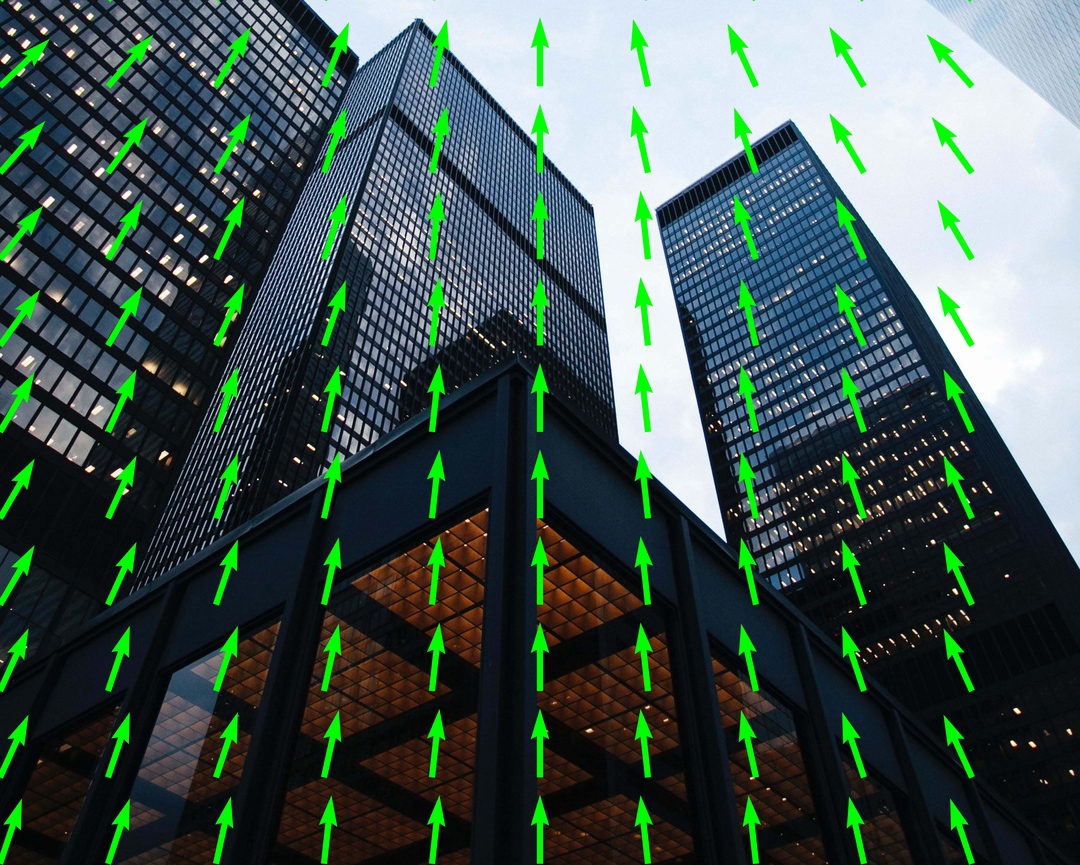}} &
\frame{\includegraphics[width=0.24\linewidth]{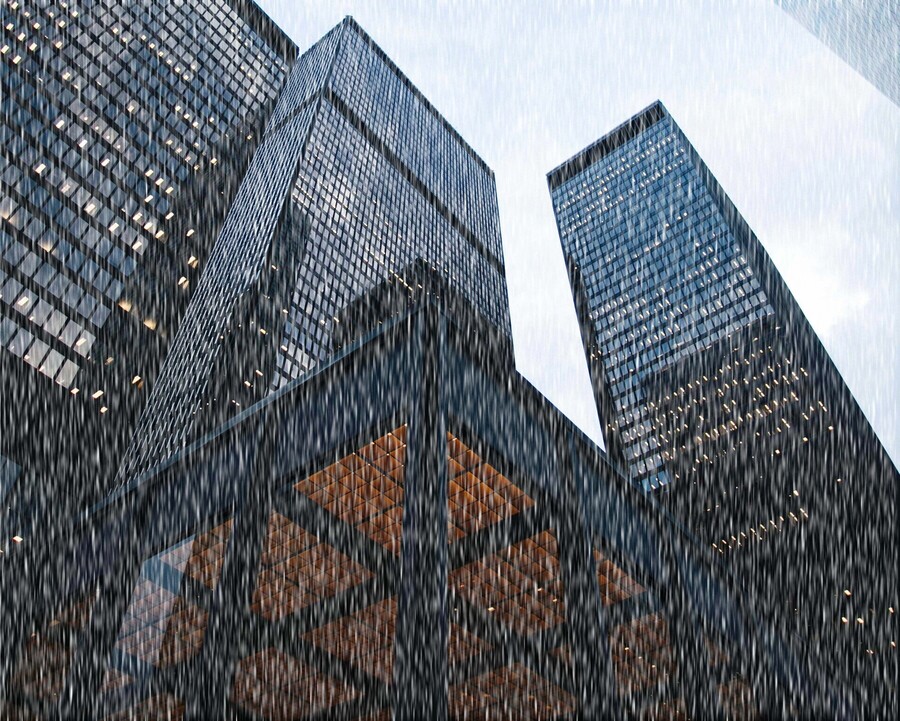}} &
\frame{\includegraphics[width=0.24\linewidth]{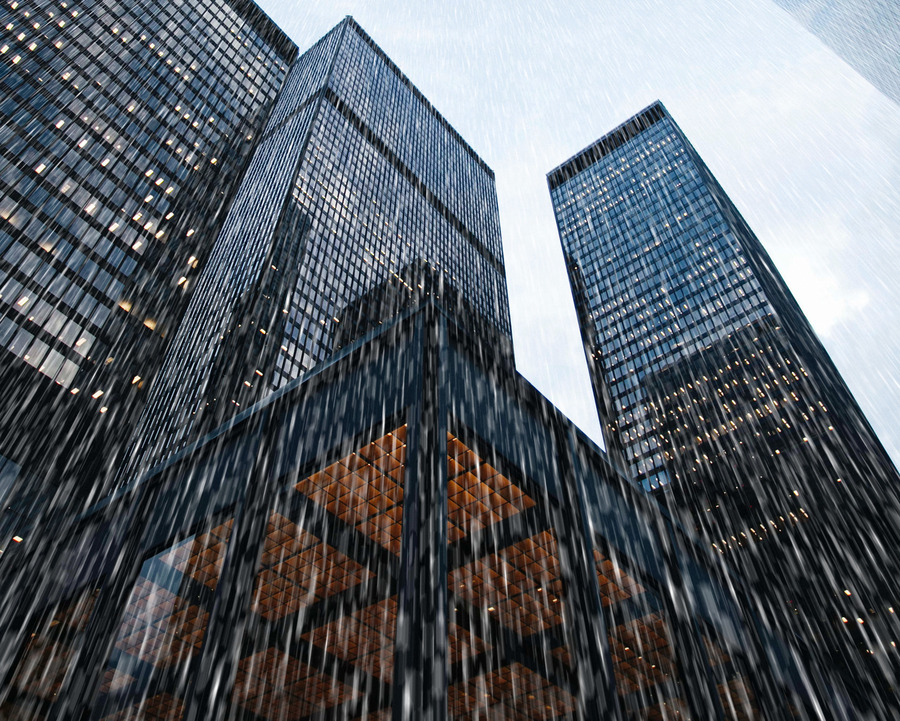}}  \\ 
    Input & Up Prediction & w/o Up & w/ Up \\
    \end{tabular}
    \vspace{-2mm}
    \caption{ Rain effect based on Perspective Fields. See supp video for a dynamic composite. 
    Our compositing {\it w/ Up} which considers the Up-vector prediction
    looks more natural than vertical raindrops in the image frame {\it w/o Up}.}
    \label{fig:rain}
    \vspace{-1em}
\end{figure}

%% file: fig/composite_idealcity/idealcity.tex
\begin{figure}[t]
    \centering
    \includegraphics[width=\linewidth]{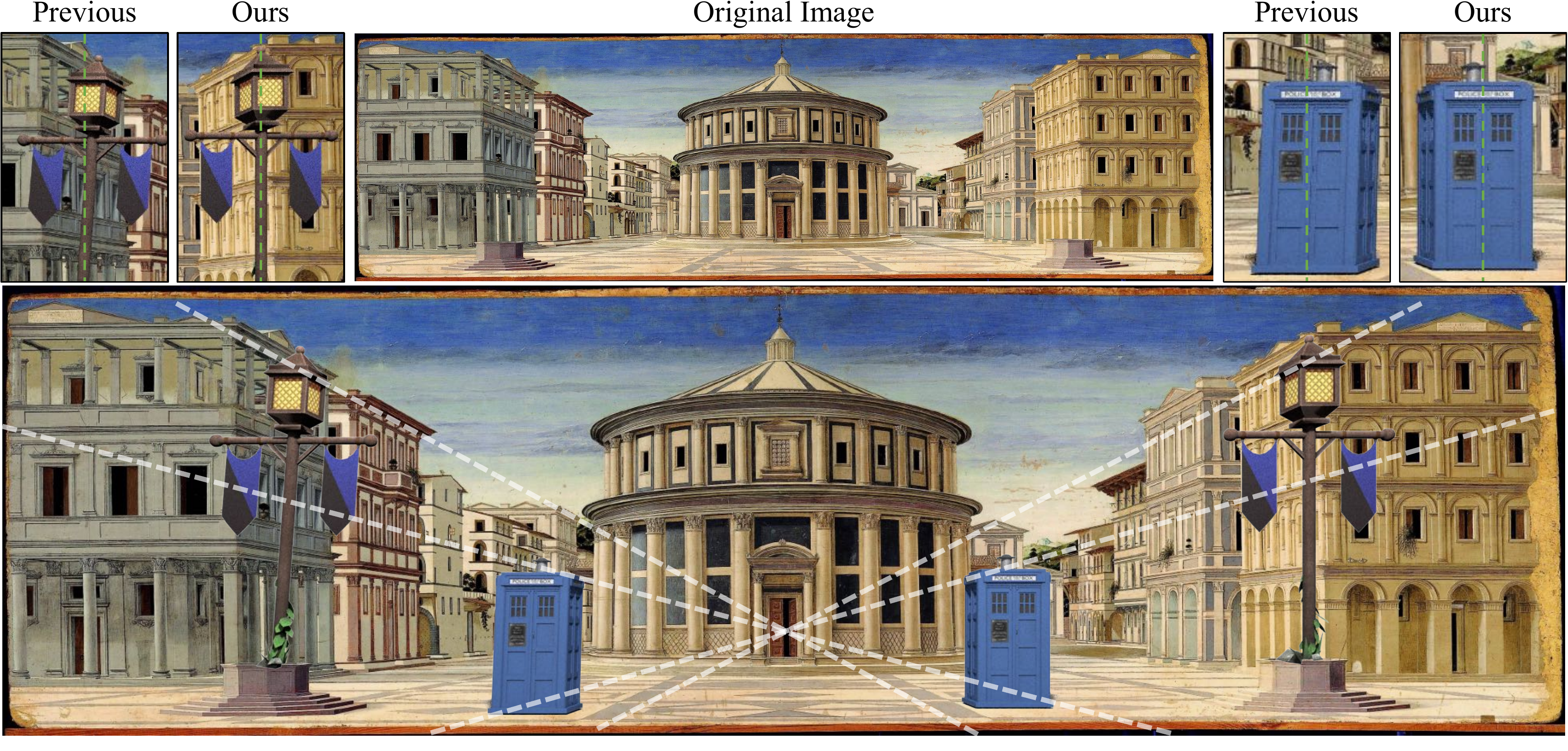}
    \caption{
        On the left half: we estimate camera by \cite{hold2018perceptual} and insert a 3D lamp and a Doctor Who Police Box. 
        One may notice that it looks a bit off, and especially that the tip of the lamp is tilted. 
        This is because~\cite{hold2018perceptual} predicts pitch $=7^\circ$ which matches the horizon line, but causes distortion in the up direction in that region.
        In contrast, on the right half, using our Perspective Fields to estimate the camera view, 
        we can correctly maintain the perspective consistency for the temple on the right.
        The white dashed lines intersect at the horizon location, the green dashed lines are up direction.}
    \label{fig:idealcity}
    \vspace{-0.2in}
\end{figure}

%% file: sec/5_conclusions.tex
\section{Conclusion}

We propose Perspective Fields, an over-complete representation capturing the local perspective properties of an image. 
We introduce a neural network model to predict Perspective Fields for various image types, and ParamNet which recovers camera parameters directly from Perspective Fields. 
Perspective Fields can serve as a metric to quantify perspective matching quality in image compositing. 
As a local representation, it is robust to different camera models and lens types, and several image editing operations, a complete study of which we leave to future work. 

%% file: sec/X_acknowledgements.tex
\bfpar{Acknowledgements.} 
This work was partially funded by the DARPA Machine Common Sense Program.
We thank Geoffrey Oxholm for the help with
Upright, and Aaron Hertzmann, Scott Cohen, 
Ang Cao, Dandan Shan, Mohamed El Banani, Sarah Jabbour, Shengyi Qian for discussions.

%% file: sec/X_supplementary.tex
\setcounter{page}{1}

\appendix

\section{Video}
Please check out the \href{https://www.youtube.com/watch?v=sN5B_ZvMva8}{video} for a demo.

\section{Additional experiments}

\subsection{Perspective Fields on warped images}

\input{tab/dense_supp}
\input{tab/opt_crop}
Table \ref{tab:dense-supp} shows the additional test results on warped images, extending Table 1 of the main paper. On warped images, which is another common operation of image post-processing, our method continues to outperform other baselines and keeps the error on Up and Latitude low. Previous methods which assume a global set of parameters poorly describe the perspective of the image and have a large performance drop.

\subsection {Ablations: training on centered principal point images.}
\input{tab/dense_centered_rebuttal}
\input{tab/dense_centered_obj_rebuttal}
Our method is trained on non-centered principal points images. In Table~\ref{tab:dense_centered_rebuttal}, we re-train Ours without \texttt{RandomResizedCrop} during data augmentation ({\it Ours-centered}) so that all the methods are trained on centered principal point images. We show results on the test set and compare to Ours and the most competitive baseline Percep.~\cite{hold2018perceptual}.
\noindent When tested on centered principal point images ({\it Perturb=None}), {\it Ours-centered} is  better than {\it Ours} in Table 1. When tested on image crops ({\it Perturb=Crop}), {\it Ours-centered} is slightly worse than {\it Ours}, but better than all other baselines. 
We obtain similar results on the TartanAir [45] dataset (not shown due to limited space).
Even when trained on centered principal points, the dense per-pixel nature of the representation makes {\it Ours-centered} to be robust to image crops.

In Table 2, distilling the {\it Ours-centered} version on crops for COCO also improves over the baselines and is comparable to {\it Ours-distill}, see Table~\ref{tab:dense-obj-rebuttal}. For example, when {\it Perturb=crop}, it has 3.93 vs 3.76 median error for Up and 6.66 vs 7.57 median error for Latitude; when {\it Perturb=isolated}, it has 4.57 vs 4.12 median error for Up and 10.08 vs 9.56 median error for Latitude ({\it Ours-centered-distill} vs {\it Ours-distill}). 

\subsection{Camera parameter estimation using optimization}
In Sec. 4.2 we have shown that camera parameters can be accurately recovered from Perspective Fields using the ParamNet. In this section, we will show that optimization can also be used to recover camera parameters and, in some cases, to improve upon predictions from ParamNet.
\bfpar{Setup.} 
The optimization problem is five dimensional as the five optimizable parameters are roll, pitch, relative focal length and the principal point (cx, cy). The relative focal length is defined as the focal length divided by image height. Relative focal length is then converted to FoV for evaluation. Adam was chosen as the optimizer with a learning rate of $10^{-4}$. The optimization runs for $\mathrm{1000}$ iterations and stops if $\mathrm{loss < 10^{-7}}$ or if $\mathrm{loss - previous\_loss < 10^{-9}}$. To perform the optimization, the Up-vector and Latitude fields are generated from the optimizable camera parameters. The loss is then calculated between these predicted fields and the ground truth Up-vector and Latitude fields. The objective function we minimize is the APFD metric
\begin{equation}
    \mathrm{Loss} = \lambda\mathrm{arccos}(\uB_1\cdot\uB_2) + (1- \lambda)||l_1 - l_2||_1,
    \label{eq:Loss}
\end{equation}
where $\uB_i$ is the Up-vector and $l_i$ is the Latitude value. The weight $\lambda=0.5$ is used in our experiments.
\bfpar{Parameter Initialization.}
We experiment with two different methods of initializing the camera parameters for the optimization. {\it Opt:} Let $\uB_x$ be defined as the center of the Up-vector field and $l_{x}$ be defined as the center of the Latitude field. The camera roll is initialized to 
    $-\mathrm{arctan}({\uB_{x_{0}}}/{-\uB_{x_{1}}})$.
The pitch is initialized to $l_{x}$. Let $l_1$ be the value of the Latitude map at the top center of the image and $l_2$ be the value of the Latitude map at the bottom center of the image. The FoV is initialized to $l_1 - l_2$. The second method of initializing the camera parameters {\it (ParamNet + Opt)} initializes them to the output of ParamNet. The FoV is converted to relative focal length for the optimization.
\bfpar{Results.}
Results for both of these initialization methods on cropped images are shown in table \ref{tab:opt_gsv_crop}. 
{\it ParamNet} is our method described in the main paper that regresses the camera parameters from predicted Perspective Fields. We show that our camera parameters can be further improved by using optimization to adjust the predicted camera parameters to minimize the APFD error with the predicted Perspective Fields. The method that combines ParamNet and optimization has 3.8\% higher accuracy in the Latitude value.

\section{Evaluation Details}
\subsection{Dataset}
\input{fig/training_set}
\bfpar{Scene level training set.}
Our training dataset contains 360$^\circ$ panoramas in equirectangular format which covers $180^\circ$ vertically and $360^\circ$ horizontally. The dataset contains diverse scenes including 30,534 indoor, 51,157 natural and 110,879 street views. We sample crops from the panoramas with camera roll in $[-45^{\circ}, 45^{\circ}]$, pitch in $[-90^{\circ}, 90^{\circ}]$ and FoV in $[30^{\circ}, 120^{\circ}]$. Our training and validation set consist of 190830/1740 panorama images respectively. We crop one perspective image per panorama and filter out ones without too much context (if all pixels values are white or black). \Fig{training-dist} shows the camera parameter distribution of our training dataset.

\bfpar{Object centric training set.}
We choose images from COCO training set and inference our perspective field predictor to generate pseudo ground truth. We select categories in ``bicycle", ``book", ``bottle", ``chair", ``laptop'' and large objects whose area are greater than $96^2=9216$ pixels. We also disgard examples with low entropy value ($<$ 3.5) from our network classification results. As a result, we generate a training set with 8192 images.

\subsection{Training details}
We use a transformer-based backbone from SegFormer~\cite{xie2021segformer} to extract features from the input RGB image. Specifically, we use the Mix Transformer encoders (MiT-B3) designed in SegFormer to extract hierarchical features. It extracts course and fine features from the hierarchical Transformer encoder using embedding dimensions of 64, 128, 320, 512.
We find that the transformer based encoder is effective for our task since it can enforce global consistency in the perspective fields well.

The features are then fed into the All-MLP decoder in SegFormer. The decoder produces a distribution over a set of up directions or latitude values with the same resolution as the input image.
The up-vector head predicts $k_{\mathrm{up}}=72$ classes representing evenly spaced unit vectors in 2D space. The latitude head predicts $k_{\mathrm{lati}}=180$ classes representing a discrete set of latitude value for each pixel evenly spaced from $-\pi/2$ to $\pi/2$.

The input resolution is 320$\times$320. We apply random flipping, rotation, color jittering and blurring to the training data. Since our perspective fields are translation invariant and defined on images with different geometric operations such as cropping, we also have random cropping and resizing on both the input image and ground truth perspective fields as part of the data augmentation.
We use the SGD optimizer with momentum of 0.9. The learning rate is 0.01. The batch size is 32.

\subsection{Test set details}
\bfpar{Stanford2d3d / TartanAir test set generation.}
Assuming perspective projection, we uniformly sample 2,415 views from Stanford2D3D with camera roll in $[-45^{\circ}, 45^{\circ}]$, 
pitch in $[-50^{\circ}, 50^{\circ}]$ and FoV in $[30^{\circ}, 120^{\circ}]$. 
For TartanAir, we randomly sample 2,000 images from its test sequences with roll ranging in  $[-20^{\circ}, 20^{\circ}]$, 
pitch in $[-45^{\circ}, 30^{\circ}]$, and fixed FoV (74$^{\circ}$). 
To test the robustness of methods, we add image crop perturbation to the test image. 
We randomly crop a quarter of the original image of aspect ratio 1, which is implemented by RandomResizedCrop function from the Albumentation~\cite{info11020125} package. The ground truth Perspective Fields can simply be cropped in the same way to match the RGB image. 
For warp perturbation, we perform a random four point perspective transform of the original image, the operation is also implemented in Albumentation~\cite{info11020125}, with hyperparameters set as scale=(0.1, 0.2), fit\_output=False.
The ground truth Latitude map is warped the same way as the RGB image. The corresponding Up-vectors are calculated by the Homography.

\input{fig/crop_uniform_distribution/gsv_crop_distribution}

\bfpar{GSV uncentered principal-point test set generation.} 
We randomly sample crops from the GSV views. 
\Fig{supp:gsv_crop_dist} shows the camera parameter distribution for the GSV {\it uncentered} principal-point dataset.

\subsection{Infer ground truth for in the wild images.}
\input{fig/laptop_gt_horizon/laptop.tex}

The qualitative examples in Figure 5 do not have a ground truth since they are from the internet. 
To help infer the ground truth, in \Fig{supp:laptop} we show the location of the GT horizon location. 
Assuming the laptop is placed on a horizontal surface, we find the vanishing points of the two pairs of parallel lines ({\color{Cyan} cyan dashed lines}) at the base. 
The horizon line can be found by connecting the vanishing points ({\color{orange} orange dashed lines}), which is outside of the image. 
Our method has more accurate Latitude prediction compared to other baselines as shown in Figure 5 of the paper.

\subsection{User study for perspective matching metrics.}

\input{fig/userstudyboxplot/userstudy_boxplot}

\input{fig/user_study/user_rankings_supp}

\Fig{supp:user_corr_boxplot} shows the statistics of the correlation scores for each metric. The box plot shows the minimum, maximum, median, 1st quartile, 3rd quartile and outliers of each metric following the standard box plot convention\footnote{\url{https://en.wikipedia.org/wiki/Box_plot}}. For the camera parameter metrics, such as deviation in roll, pith, FoV and the principal point ({\it Prin.~Point}), the correlation score distributions vary wildly. Among them, deviation in FoV is a poor indication of human perception, which is consistent with \cite{hold2018perceptual}. The change in pitch is a dominant factor in perspective mismatch in our setting. 
Summing the parameter difference ({\it Camera All}) does not improve  correlation scores, which shows the difficulty of using camera parameters to measure perceived perspective consistency.
\Fig{user_ranking_supp} shows the user rankings and APFD scores on different test images. 

\section{Additional Qualitative Results}

\bfpar{Additional Qualitative Results on Test Set.}
We show qualitative results on Stanford2D3D and TartanAir test sets in \Fig{supp:densewallcomparison:stanford2d3d} and \Fig{supp:densewallcomparison:tartanair}.

\input{fig/supp_dense_wall_img_stanford2d3d}

\input{fig/supp_dense_wall_img_tartanair}

\bfpar{Additional Qualitative Results on Web Images.}
We show additional qualitative results on web images in \Fig{wild_supp} and \Fig{wild_supp_obj}.

\bfpar{Qualitative Results on Fisheye Images.}
We show qualitative results of predicting Perspective Fields for fisheye images in \Fig{supp:fisheye}. 
{\it Sliding Win.}:  We take advantage of the local representation and use a sliding window inference technique for images that are out of our training distribution.
We inference on small crops and aggregate the prediction
for each pixel from overlapping windows.
The results in \Fig{supp:fisheye} use a window of size $(0.5  \mathrm{img\_height}) \times (0.5  \mathrm{img\_width})$.
This window slides along a $12 \times 18$ grid uniformly on the image and at each point predicts the Up-vectors and Latitude Map within the window. The final output for these values at each pixel is the mean of that pixels values in each window that it was apart of.
{\it Fine-tune}, we show results after fine-tuning the PerspectiveNet on distorted images.

\bfpar{Additional Qualitative Results on Google Street View}
In \Fig{supp:gsv_pers} we show additional qualitative results from PerspectiveNet as well as Persepctive Fields generated from the ParamNet predictions on GSV {\it uncentered principal-point} test set.

\input{fig/wild/wild_supp}

\input{fig/wild/wild_supp_obj}

\input{fig/supp_fisheye_wall}

\input{fig/supp_pers}

\section{User Study Data Collection Interface}
\Fig{instruction} shows the instruction users see and \Fig{interface} is the interface users use when collecting human perceptual preferences.
\input{fig/user_study/instruction}

\clearpage
\input{fig/user_study/interface}

\clearpage

%% file: tab/dense_supp.tex
\begin{table*}[t]
    \centering
        \caption{Quantitative evaluation for scene-level Perspective Field prediction on warped images, extending Table 1. 
        We re-implement Percep.~\cite{hold2018perceptual} using the same backbone and 
        training data as ours. 
        None of the methods have been trained on Stanford2D3D~\cite{armeni2017joint} or TartanAir~\cite{tartanair2020iros}. 
    } %
    \label{tab:dense-supp}
    \resizebox{\textwidth}{!}{ %
    \begin{tabular}{@{}l|c|cccccc|cccccc@{}}
    \toprule
     Dataset & &  \multicolumn{6}{|c}{Stanford2D3D~\cite{armeni2017joint}} & \multicolumn{6}{|c}{TartanAir~\cite{tartanair2020iros}} \\
     & & 
     \multicolumn{3}{|c}{Up ($^o$)} & \multicolumn{3}{c}{Latitude  ($^o$)} & \multicolumn{3}{|c}{Up ($^o$)} & \multicolumn{3}{c}{Latitude  ($^o$)} \\
    Method & Perturb &
    Mean $\downarrow$ & Median $\downarrow$ & $\%<5^o$ $\uparrow$ & 
    Mean $\downarrow$ & Median $\downarrow$ & $\%<5^o$ $\uparrow$ & 
    Mean $\downarrow$ & Median $\downarrow$ & $\%<5^o$ $\uparrow$ & 
    Mean $\downarrow$ & Median $\downarrow$ & $\%<5^o$ $\uparrow$ \\
     \midrule
Upright~\cite{lee2014upright} & Warp &
11.16 & 10.47 & 38.46 & 20.50 & 20.38 & 13.65 &
13.77 & 13.11 & 34.82 & 18.20 & 18.44 & 15.89 \\
Percep.~\cite{hold2018perceptual} & Warp &
10.01 & 9.25 & 34.29 & 14.23 & 13.77 & 20.93 &
9.55 & 8.76 & 33.84 & 9.85 & 9.59 & 27.14 \\
CTRL-C~\cite{lee2021ctrl} & Warp &
15.92 & 14.79 & 19.86 & 13.09 & 12.38 & 22.96 &
14.61 & 13.34 & 20.72 & 10.86 & 10.66 & 24.44 \\
Ours & Warp &
\textbf{3.39} & \textbf{2.72} & \textbf{66.82} & \textbf{5.95} & \textbf{5.48} & \textbf{46.79} &
\textbf{4.11} & \textbf{3.45} & \textbf{61.08} & \textbf{5.47} & \textbf{5.12} & \textbf{48.62} \\
    \bottomrule
    \end{tabular}
    } %

    \end{table*}

%% file: tab/opt_crop.tex
\begin{table*}[t!]
    \renewcommand{\arraystretch}{0.9}
    \caption{GSV {\it uncentered} principal-point optimization results. ParamNet is our method described in the main paper that regresses the camera parameters from predicted Perspective Fields. We show that camera parameters can be further improved by using optimization to adjust the predicted camera parameters to better match the Perspective Fields.}
    \vspace{-1em}
    \label{tab:opt_gsv_crop}
    \resizebox{\linewidth}{!}{ %

    \begin{tabular}{@{}l|cc|cc|cc|cc|cc|cc|cc@{}}
    \toprule
    Method & \multicolumn{2}{c|}{Roll ($^\circ$) $\downarrow$} & \multicolumn{2}{c|}{Pitch ($^\circ$) $\downarrow$} & \multicolumn{2}{c|}{FoV$^*$ ($^\circ$) $\downarrow$} &\multicolumn{2}{c|}{cx $\downarrow$} &\multicolumn{2}{c|}{cy $\downarrow$} &\multicolumn{2}{c|}{Up($^\circ$) } &\multicolumn{2}{c}{Latitude($^\circ$)} \\
      & Mean & Med. & Mean & Med. & Mean & Med. & Mean & Med. & Mean & Med.  & Mean. $\downarrow$& $\%<5^o$$\uparrow$ & Mean. $\downarrow$& $\%<5^o$ $\uparrow$\\ 
    \midrule
    
    ParamNet                         &  \textbf{1.37} & 0.97 & \textbf{2.60} & \textbf{2.14} & 3.75 & 3.19 & \textbf{0.09} & \textbf{0.07} & \textbf{0.08} & \textbf{0.06} & 1.05 & 98.95 & 2.17  & 89.47 \\
    Opt & 1.90 & 1.15 & 3.68 & 2.90 & 3.80 & \textbf{3.16} & 0.12 & 0.10 & 0.09 & 0.07 & 1.00 & \textbf{99.40} & 1.93 & 93.15 \\
    ParamNet + Opt & 1.41 & \textbf{0.95} & \textbf{2.60} & \textbf{2.14} & \textbf{3.72} & 3.17 & 0.10 & 0.08 & \textbf{0.08} & \textbf{0.06} & \textbf{0.80} & \textbf{99.40} & \textbf{1.91} & \textbf{93.30} \\
    \bottomrule
\end{tabular}
} %
\end{table*}

%% file: tab/dense_centered_rebuttal.tex
\begin{table*}[t]
    \centering
        \caption{We re-train Ours without \texttt{RandomResizedCrop} during data augmentation ({\it Ours-centered}) so that all the methods are trained on centered principal point images, extending Table 1. We compare to Ours and the most competitive baseline Percep.~\cite{hold2018perceptual}.} %
    \label{tab:dense_centered_rebuttal}
    \resizebox{\textwidth}{!}{ %
    \begin{tabular}{@{}l|c|cccccc|cccccc@{}}
    \toprule
     Dataset & &  \multicolumn{6}{|c}{Stanford2D3D~\cite{armeni2017joint}} & \multicolumn{6}{|c}{TartanAir~\cite{tartanair2020iros}} \\
     \midrule
     & & 
     \multicolumn{3}{|c}{Up ($^o$)} & \multicolumn{3}{c}{Latitude  ($^o$)} & \multicolumn{3}{|c}{Up ($^o$)} & \multicolumn{3}{c}{Latitude  ($^o$)} \\
    Method & Perturb &
    Mean $\downarrow$ & Median $\downarrow$ & $\%<5^o$ $\uparrow$ & 
    Mean $\downarrow$ & Median $\downarrow$ & $\%<5^o$ $\uparrow$ & 
    Mean $\downarrow$ & Median $\downarrow$ & $\%<5^o$ $\uparrow$ & 
    Mean $\downarrow$ & Median $\downarrow$ & $\%<5^o$ $\uparrow$ \\
    \midrule
    Percep.~\cite{hold2018perceptual} & None &
    3.58 & 3.32 & 64.19 & 6.27 & 6.07 & 42.36 &
    7.30 & 6.86 & 47.04 & 11.35 & 11.22 & 27.69 \\
    Ours & None &
    2.18 & 1.88 & 82.83 & 3.40 & 3.06 & 68.27 &
    3.47 & 2.86 & 67.45 & 4.01 &3.60 & 61.73 \\
    Ours-centered & None & 
    \textbf{1.83} & \textbf{1.66} & \textbf{89.09} & \textbf{2.06} & \textbf{1.88} & \textbf{82.02} &
\textbf{2.11} & \textbf{1.86} & \textbf{83.06} & \textbf{2.23} & \textbf{2.04} & \textbf{81.03} \\
    \midrule
    Percep.~\cite{hold2018perceptual} & Crop &
    5.78 & 5.55 & 45.52 & 9.76 & 9.65 & 29.13 &
    5.54 & 5.18 & 51.72 & 9.22 & 8.66 & 30.10 \\
    Ours & Crop &
    \textbf{2.21} & \textbf{1.87} & \textbf{78.80} & \textbf{5.57} & \textbf{5.15} & \textbf{50.36} &
    \textbf{2.81} & \textbf{2.35} & \textbf{71.89} & 5.73 & 5.28 & 50.16 \\
     Ours-centered & Crop & 
     3.07 & 2.89 & 65.91 & 5.93 & 5.56 & 45.65 &
3.64 & 3.33 & 64.13 & \textbf{5.69} & \textbf{5.26} & \textbf{49.52} \\

    \bottomrule
    \end{tabular}
    } %

    \end{table*}

%% file: tab/dense_centered_obj_rebuttal.tex
\begin{table}[t!]
    \centering
    \caption{Ablation study for training on centered principal point images only, extending Table 2.
    } %
    
    \resizebox{\linewidth}{!}{ %
    \begin{tabular}{@{}l|c|cccccc@{}}
    \toprule
     Dataset & & \multicolumn{6}{c}{Objectron~\cite{objectron2021}}  \\
     \midrule
     & & \multicolumn{3}{c}{Up ($^o$)} & \multicolumn{3}{c}{Latitude  ($^o$)} \\
    Method & Perturb & 
    Mean $\downarrow$ & Median $\downarrow$ & $\%<5^o$ $\uparrow$ & 
    Mean $\downarrow$ & Median $\downarrow$ & $\%<5^o$ $\uparrow$ \\
    \midrule
    CTRL-C~\cite{lee2021ctrl} & crop &
    7.50 & 7.09 & 40.02 & 20.93 & 21.00 & 11.26 \\
    Ours-distill & crop &
    \textbf{4.19} & \textbf{3.76} & \textbf{57.71} & 7.71 & 7.57 & 33.54 \\
    Ours-distill-centered & crop &
\textbf{4.19} & 3.93 & 57.31 & \textbf{7.02} & \textbf{6.66} & \textbf{36.76} \\
    \midrule
    \midrule
    CTRL-C~\cite{lee2021ctrl} & isolated &
    7.49 & 7.13 & 39.38 & 9.87 & 9.85 & \textbf{27.32} \\
    Ours-distill & isolated &
    \textbf{4.45} & \textbf{4.12} & \textbf{54.88} & \textbf{9.65} & \textbf{9.56} & 25.82 \\
    Ours-distill-centered & isolated &
4.85 & 4.57 & 52.21 & 10.39 & 10.08 & 27.24 \\
    \bottomrule
    \end{tabular}
    } %
    
    \label{tab:dense-obj-rebuttal}
    \end{table}

%% file: fig/training_set.tex
\begin{figure*}[h]
\setlength{\belowcaptionskip}{-8pt}
\centering
\includegraphics[width=\linewidth]{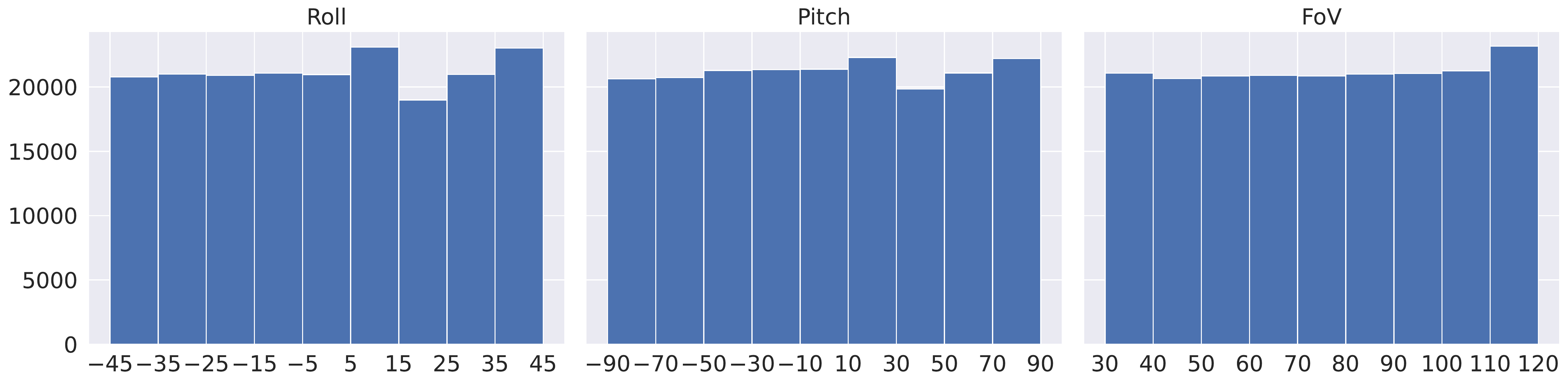}
\caption{
Training set camera distribution.}
\label{fig:training-dist}
\end{figure*}

%% file: fig/crop_uniform_distribution/gsv_crop_distribution.tex
\begin{figure*}[!t]
    \centering
    
\includegraphics[width=\linewidth]{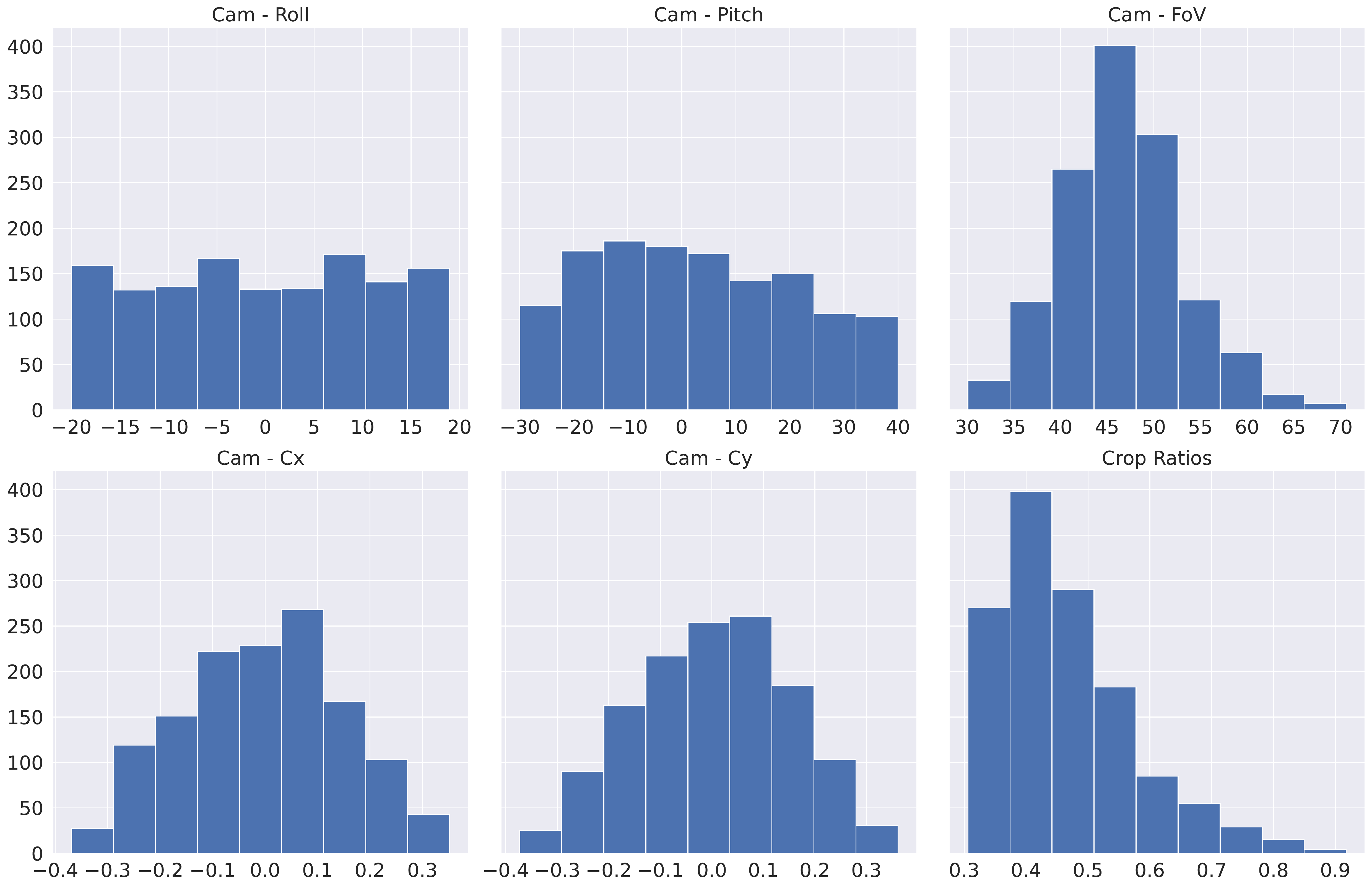} 
    \caption{Camera parameter distribution for GSV  {\it uncentered} principal-point dataset.}
    \label{fig:supp:gsv_crop_dist}
    \vspace{-1em}
\end{figure*}

%% file: fig/laptop_gt_horizon/laptop.tex
\begin{figure*}[!t]
    \centering
    \begin{tabular}{cc}
\includegraphics[width=0.8\textwidth]{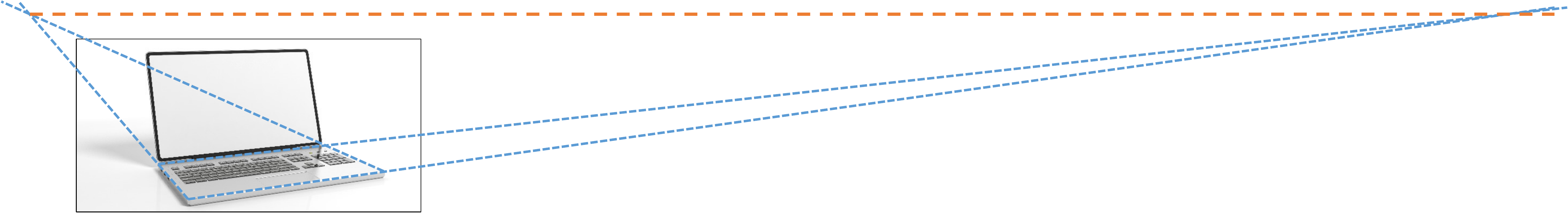} &

\frame{\includegraphics[width=0.2\linewidth]{fig/wild/merged/AdobeStock_133910589/perspective_pred.jpg}} \\ 
GT Horizon line & Ours \\
\end{tabular}
    \caption{The GT horizon location ({\color{orange} Orange dashed line}) of the laptop example from the web image. 
    Our method has more accurate Latitude prediction compared to other baselines as shown in Figure 5 of the paper.}
    \label{fig:supp:laptop}
\end{figure*}

%% file: fig/userstudyboxplot/userstudy_boxplot.tex
\begin{figure}[t]
    \centering
    \scriptsize
    \resizebox{\linewidth}{!}{ %
    \begin{tabular}{c}
        \includegraphics[width=.37\linewidth]{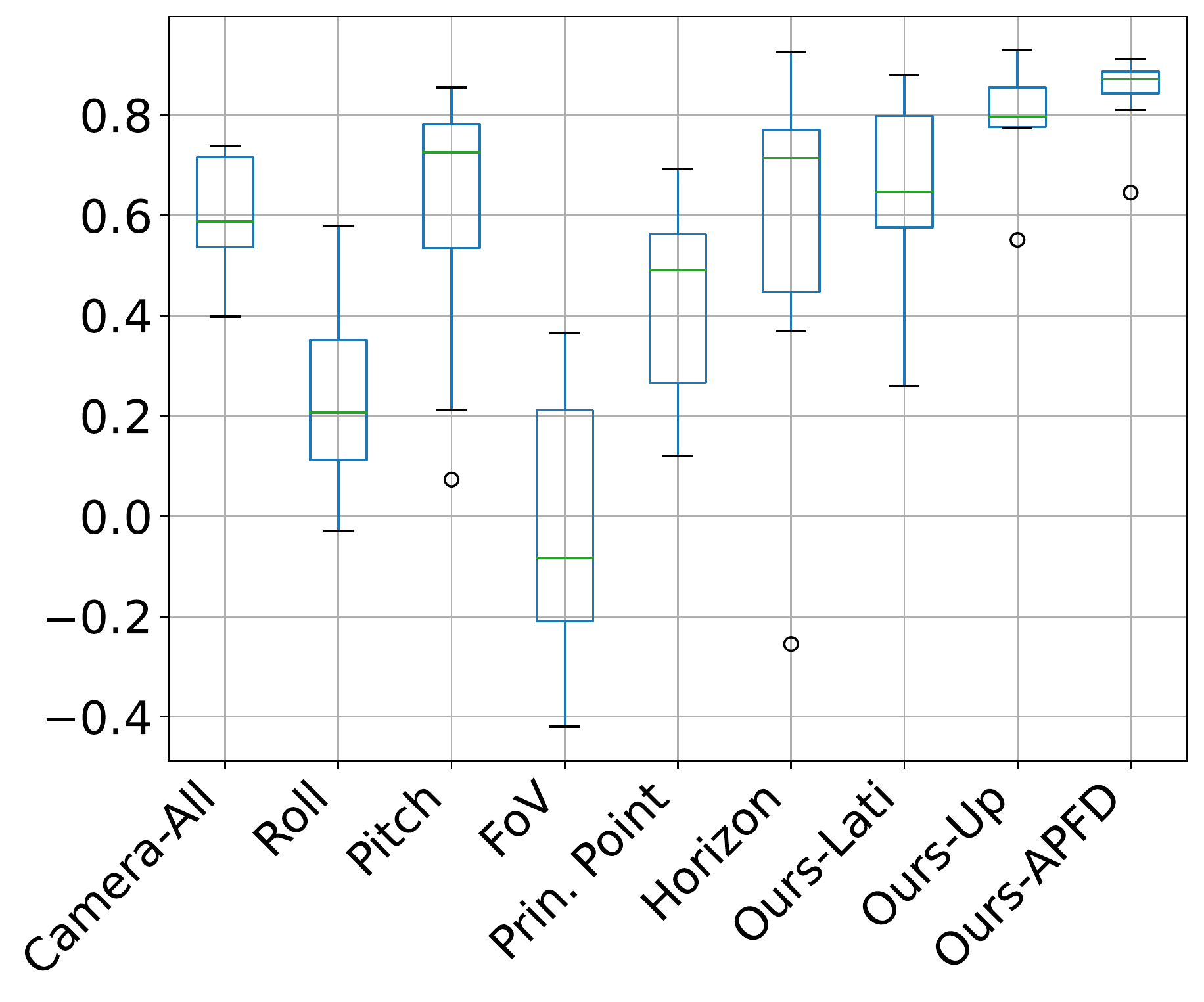}
    \end{tabular} 
    } %
    
    \caption{Pearson's correlation for different metrics \emph{\wrt} human perception. 
    Our APFD metric has the highest correlation with human perception.}
    \label{fig:supp:user_corr_boxplot}
\end{figure}

%% file: fig/user_study/user_rankings_supp.tex
\begin{figure*}[h!]
    \centering
    \scriptsize
    \resizebox{\textwidth}{!}{
    \begin{tabular}{c@{\hskip4pt}c}
    \includegraphics[width=0.5\linewidth]{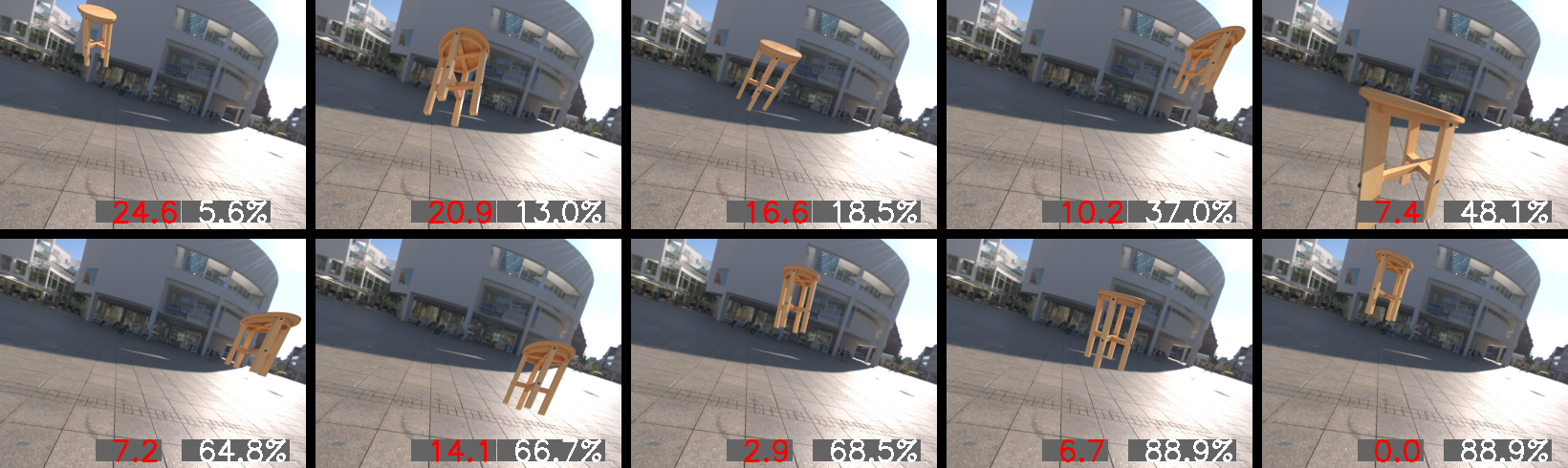} &
    \includegraphics[width=0.5\linewidth]{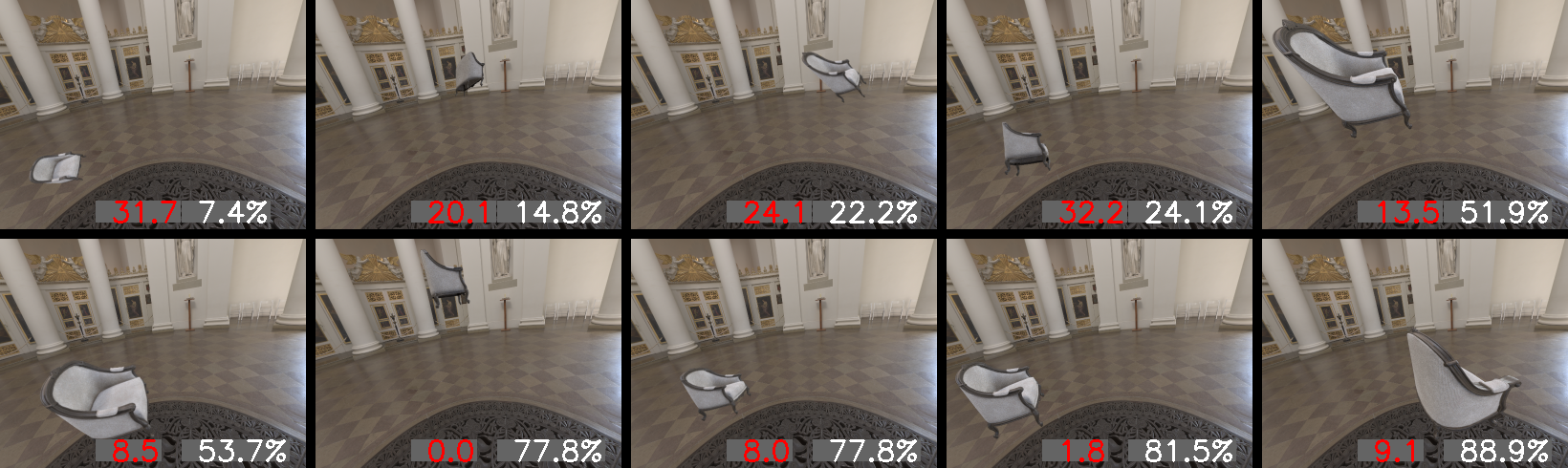}\\
    \includegraphics[width=0.5\linewidth]{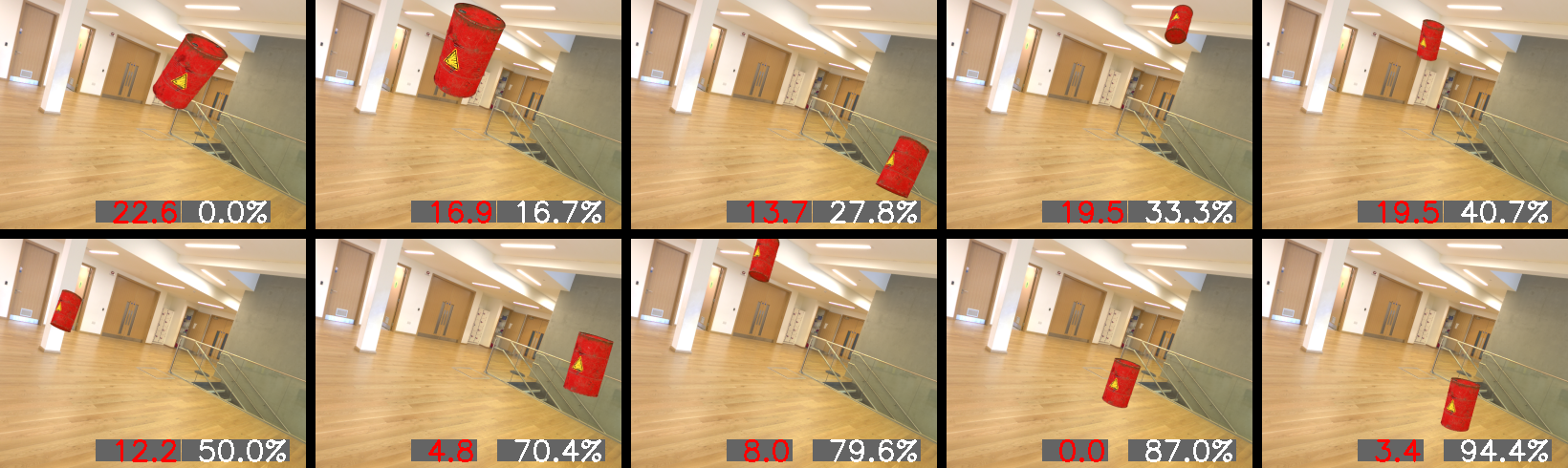} &
    \includegraphics[width=0.5\linewidth]{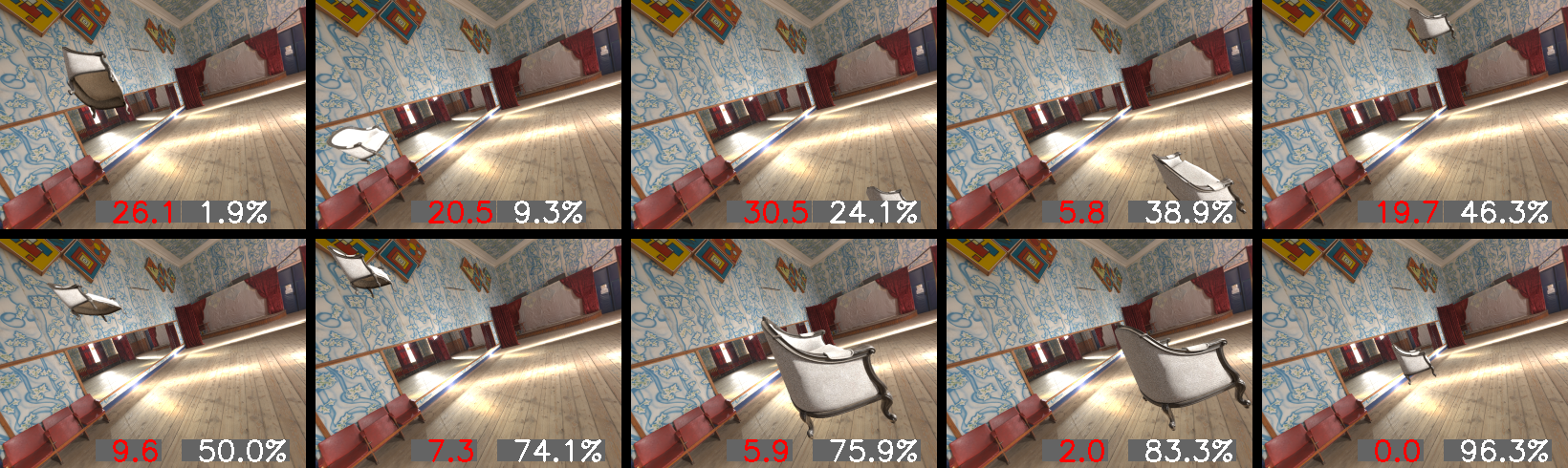}\\
    \includegraphics[width=0.5\linewidth]{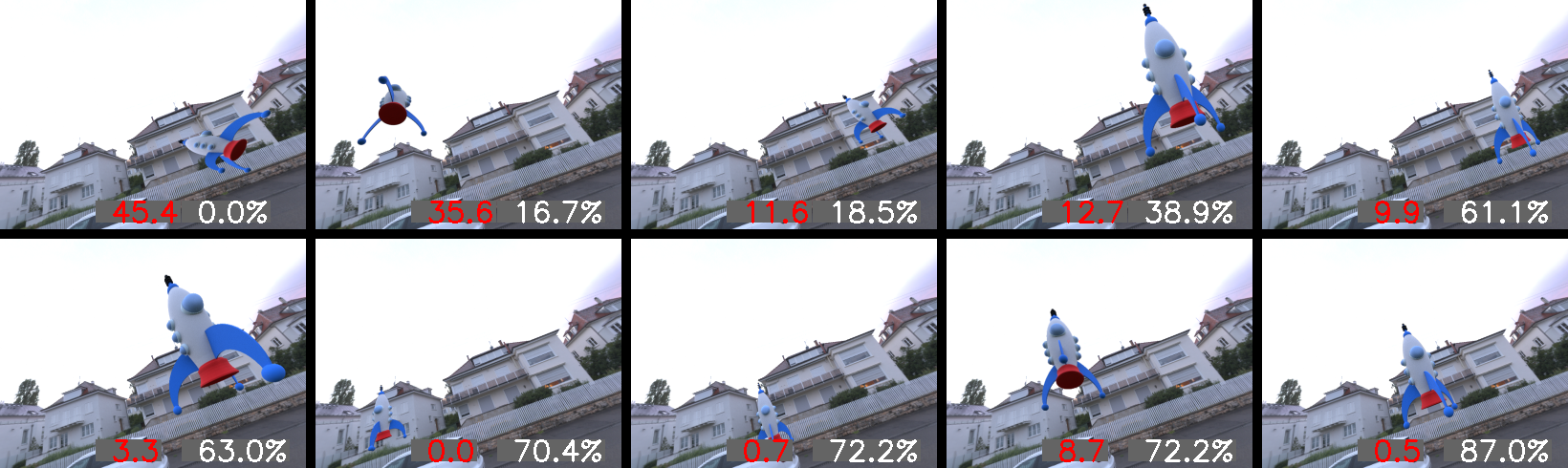} &
    \includegraphics[width=0.5\linewidth]{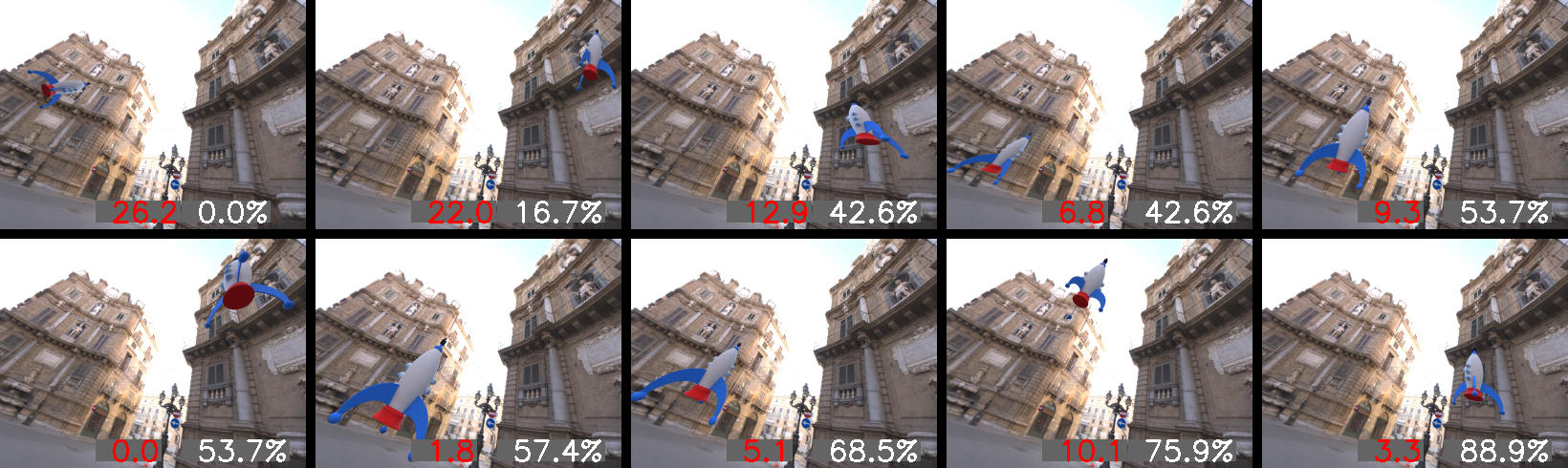}\\
    \includegraphics[width=0.5\linewidth]{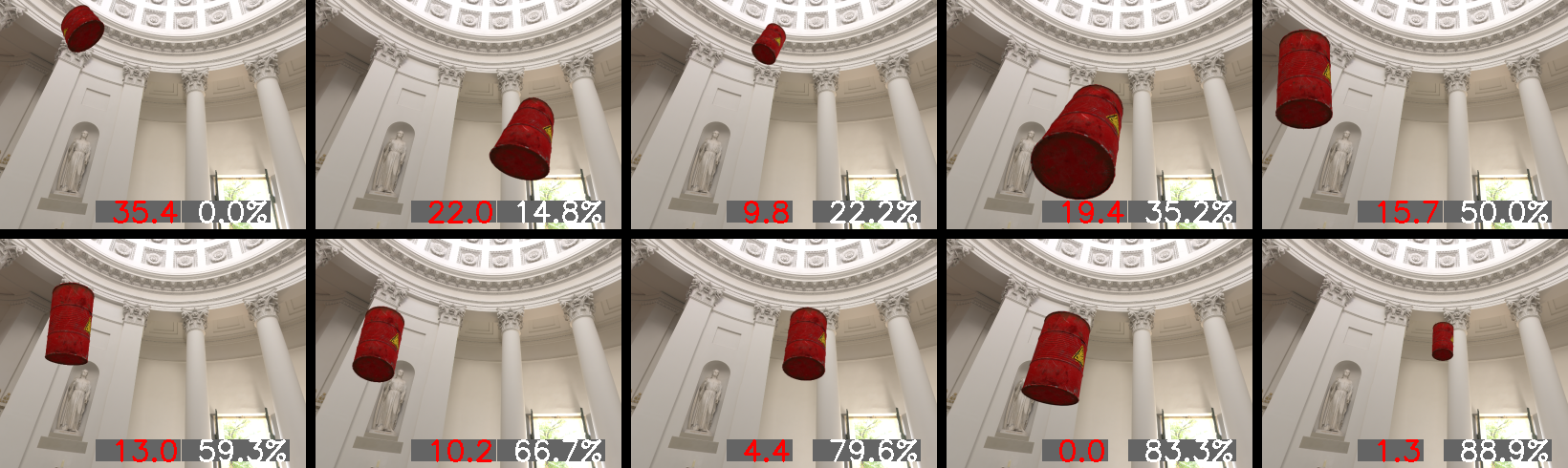} &
    \includegraphics[width=0.5\linewidth]{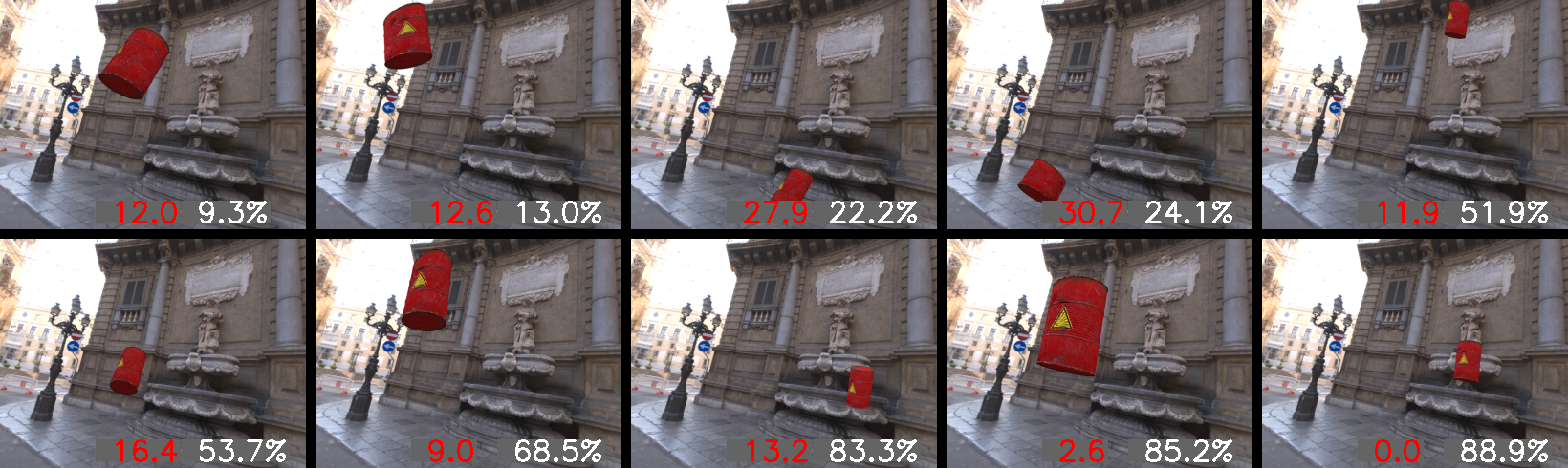}\\
    
    \end{tabular}
    }
    \caption{User study examples and results. Given a background image and an object, we randomly generate 10 compositing results with varied distortions. Pair-wise comparison is performed by a group of subjects. The white percentage number is the average winning rate based on human votes (the higher the better), and the red number is the APFD metric computed based on the Perspective Fields of the object and the background (the lower the better). There is a strong correlation between the perceptual quality and our metric.}
    \label{fig:user_ranking_supp}
\end{figure*}

%% file: fig/supp_dense_wall_img_stanford2d3d.tex
\begin{figure*}[h]
    \centering
    \scriptsize
    \resizebox{\textwidth}{!}{
    \begin{tabular}{cc@{\hskip4pt}c@{\hskip4pt}c@{\hskip4pt}c@{\hskip4pt}c@{\hskip4pt}c}
    Input & Upright~\cite{lee2014upright} & Percep.~\cite{hold2018perceptual} & CTRL-C~\cite{lee2021ctrl} & Ours & Ground Truth \\
    \midrule 
    
    \frame{\includegraphics[width=0.15\textwidth]{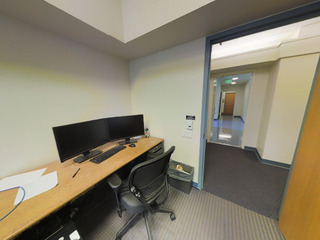}} &
\frame{\includegraphics[width=0.15\textwidth]{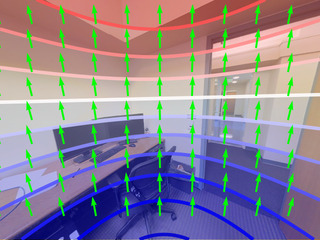}} &
\frame{\includegraphics[width=0.15\textwidth]{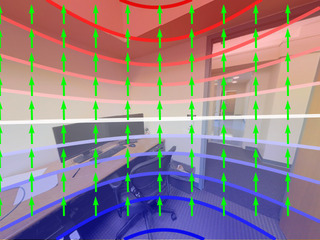}} &
\frame{\includegraphics[width=0.15\textwidth]{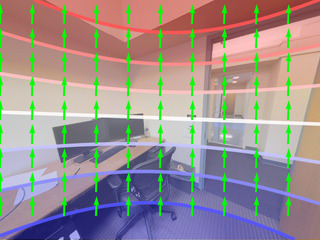}} &
\frame{\includegraphics[width=0.15\textwidth]{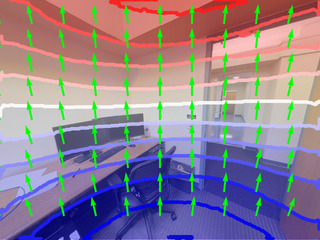}} &
\frame{\includegraphics[width=0.15\textwidth]{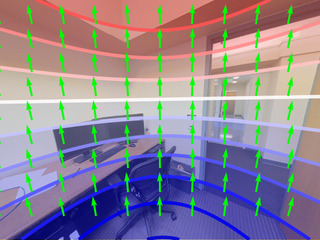}} \\
\frame{\includegraphics[width=0.15\textwidth]{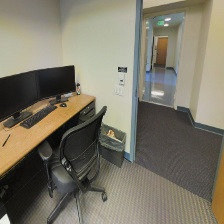}} &
\frame{\includegraphics[width=0.15\textwidth]{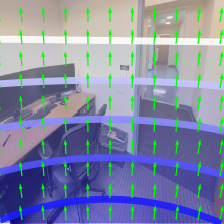}} &
\frame{\includegraphics[width=0.15\textwidth]{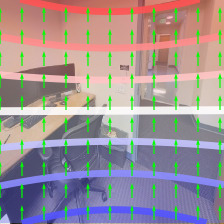}} &
\frame{\includegraphics[width=0.15\textwidth]{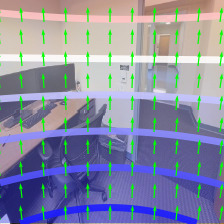}} &
\frame{\includegraphics[width=0.15\textwidth]{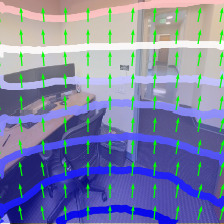}} &
\frame{\includegraphics[width=0.15\textwidth]{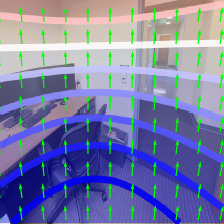}} \\

\frame{\includegraphics[width=0.15\textwidth]{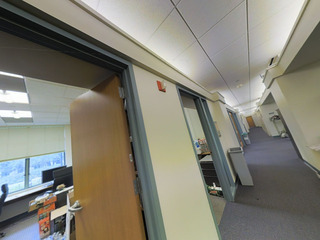}} &
\frame{\includegraphics[width=0.15\textwidth]{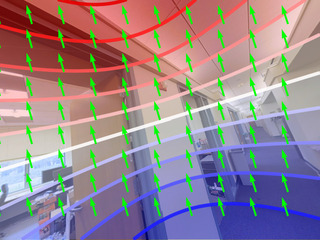}} &
\frame{\includegraphics[width=0.15\textwidth]{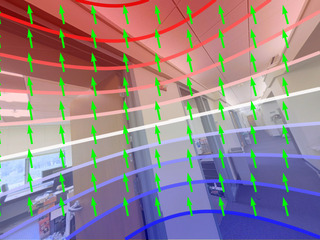}} &
\frame{\includegraphics[width=0.15\textwidth]{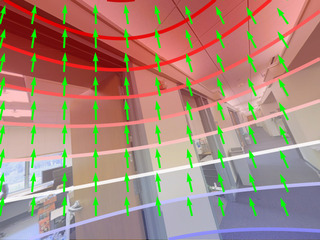}} &
\frame{\includegraphics[width=0.15\textwidth]{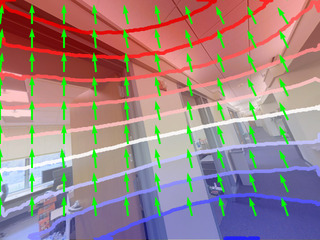}} &
\frame{\includegraphics[width=0.15\textwidth]{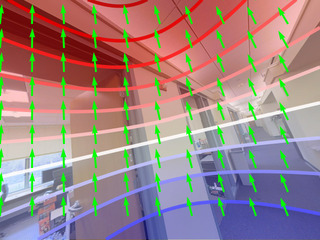}} \\
\frame{\includegraphics[width=0.15\textwidth]{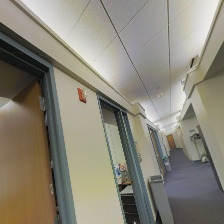}} &
\frame{\includegraphics[width=0.15\textwidth]{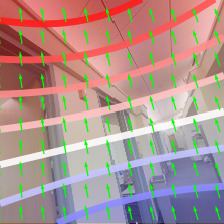}} &
\frame{\includegraphics[width=0.15\textwidth]{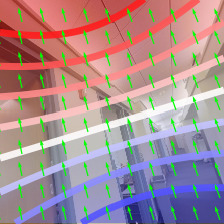}} &
\frame{\includegraphics[width=0.15\textwidth]{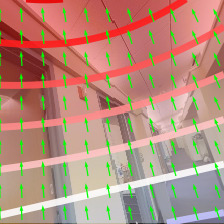}} &
\frame{\includegraphics[width=0.15\textwidth]{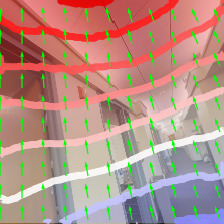}} &
\frame{\includegraphics[width=0.15\textwidth]{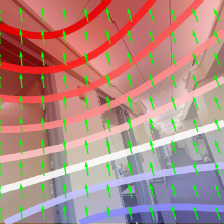}} \\
\frame{\includegraphics[width=0.15\textwidth]{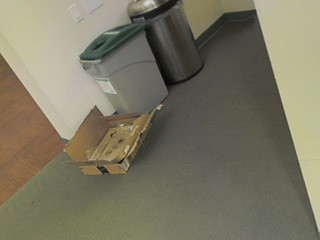}} &
\frame{\includegraphics[width=0.15\textwidth]{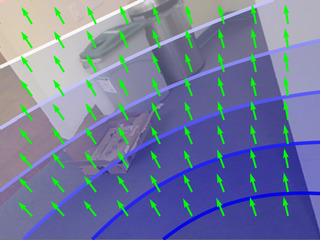}} &
\frame{\includegraphics[width=0.15\textwidth]{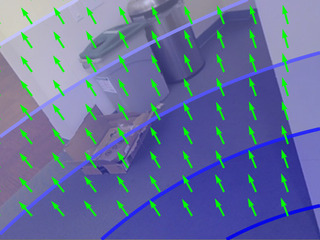}} &
\frame{\includegraphics[width=0.15\textwidth]{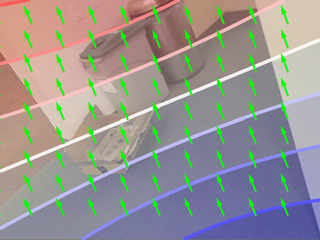}} &
\frame{\includegraphics[width=0.15\textwidth]{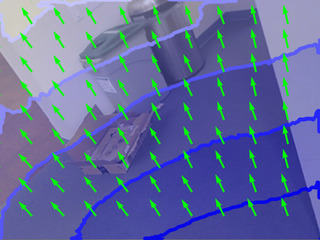}} &
\frame{\includegraphics[width=0.15\textwidth]{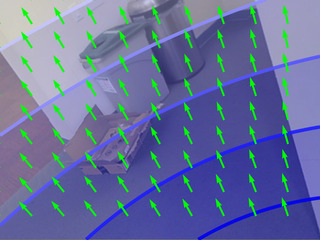}} \\
\frame{\includegraphics[width=0.15\textwidth]{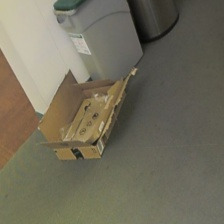}} &
\frame{\includegraphics[width=0.15\textwidth]{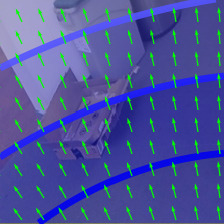}} &
\frame{\includegraphics[width=0.15\textwidth]{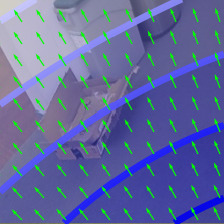}} &
\frame{\includegraphics[width=0.15\textwidth]{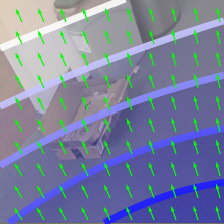}} &
\frame{\includegraphics[width=0.15\textwidth]{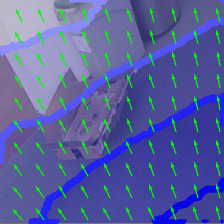}} &
\frame{\includegraphics[width=0.15\textwidth]{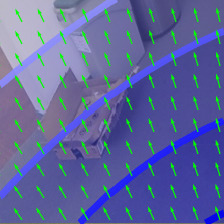}} \\
    \frame{\includegraphics[width=0.15\textwidth]{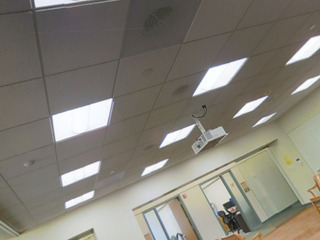}} &
\frame{\includegraphics[width=0.15\textwidth]{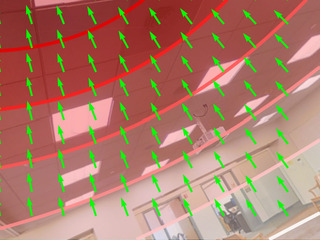}} &
\frame{\includegraphics[width=0.15\textwidth]{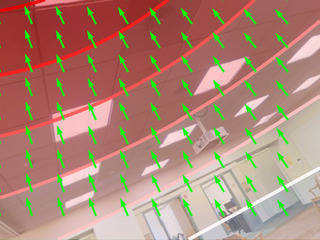}} &
\frame{\includegraphics[width=0.15\textwidth]{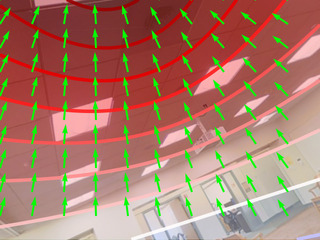}} &
\frame{\includegraphics[width=0.15\textwidth]{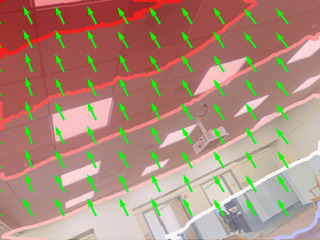}} &
\frame{\includegraphics[width=0.15\textwidth]{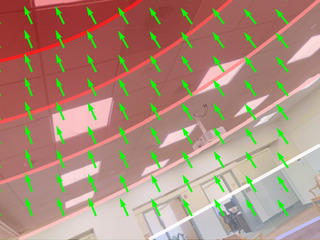}} \\
\frame{\includegraphics[width=0.15\textwidth]{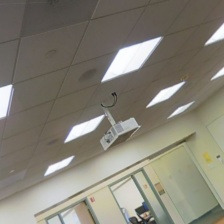}} &
\frame{\includegraphics[width=0.15\textwidth]{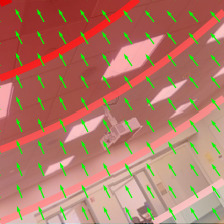}} &
\frame{\includegraphics[width=0.15\textwidth]{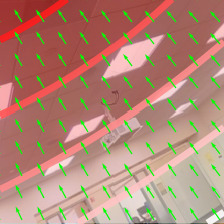}} &
\frame{\includegraphics[width=0.15\textwidth]{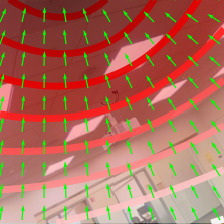}} &
\frame{\includegraphics[width=0.15\textwidth]{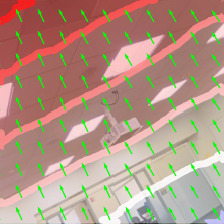}} &
\frame{\includegraphics[width=0.15\textwidth]{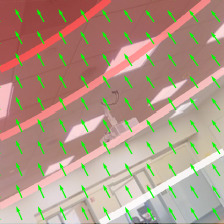}} \\
    \end{tabular}
    }
    
    \caption{Comparison between baselines on Stanford2D3D dataset. Each test scene has two rows: the first row is the original image with a standard pin-hole camera perspective; the second row is a randomly cropped image. Up-vectors in the green vectors. Latitude colormap: $-\pi/2$ \includegraphics[width=0.4in,height=8pt]{fig/seismic.png} $\pi/2$. }
    \label{fig:supp:densewallcomparison:stanford2d3d}
\end{figure*}

%% file: fig/supp_dense_wall_img_tartanair.tex
\begin{figure*}[h]
    \centering
    \scriptsize
    \resizebox{\textwidth}{!}{
    \begin{tabular}{cc@{\hskip4pt}c@{\hskip4pt}c@{\hskip4pt}c@{\hskip4pt}c@{\hskip4pt}c}
    Input & Upright~\cite{lee2014upright} & Percep.~\cite{hold2018perceptual} & CTRL-C~\cite{lee2021ctrl} & Ours & Ground Truth \\
    \midrule 
    \frame{\includegraphics[width=0.15\textwidth]{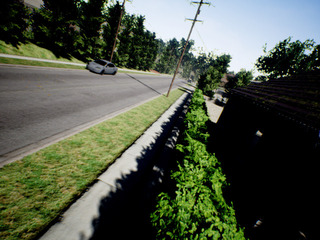}} &
    \frame{\includegraphics[width=0.15\textwidth]{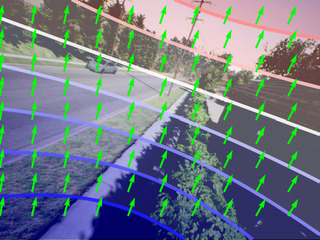}} &
    \frame{\includegraphics[width=0.15\textwidth]{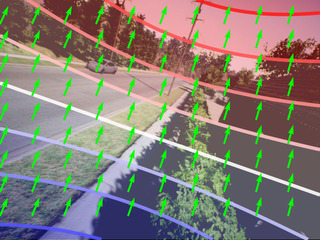}} &
    \frame{\includegraphics[width=0.15\textwidth]{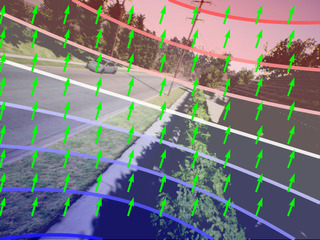}} &
    \frame{\includegraphics[width=0.15\textwidth]{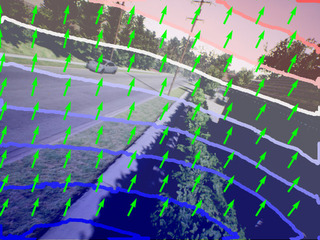}} &
    \frame{\includegraphics[width=0.15\textwidth]{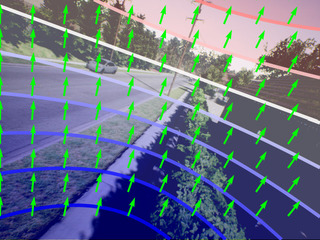}} \\
    \frame{\includegraphics[width=0.15\textwidth]{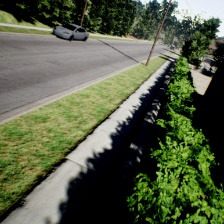}} &
    \frame{\includegraphics[width=0.15\textwidth]{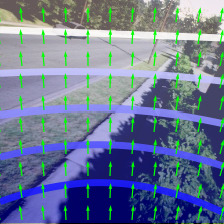}} &
    \frame{\includegraphics[width=0.15\textwidth]{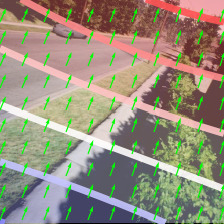}} &
    \frame{\includegraphics[width=0.15\textwidth]{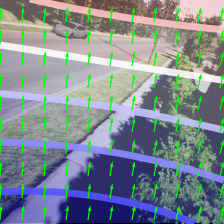}} &
    \frame{\includegraphics[width=0.15\textwidth]{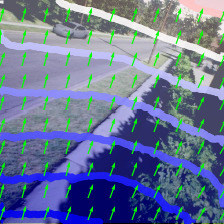}} &
    \frame{\includegraphics[width=0.15\textwidth]{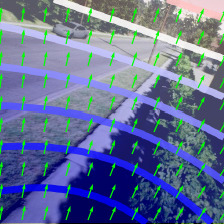}} \\
\frame{\includegraphics[width=0.15\textwidth]{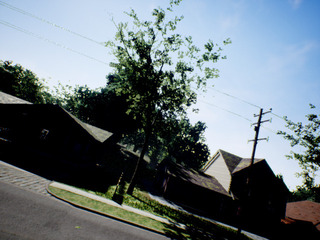}} &
\frame{\includegraphics[width=0.15\textwidth]{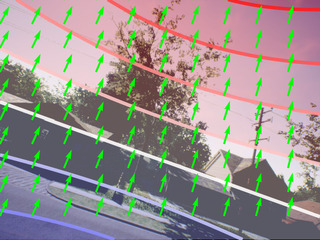}} &
\frame{\includegraphics[width=0.15\textwidth]{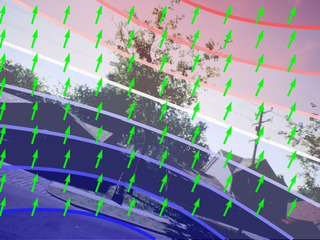}} &
\frame{\includegraphics[width=0.15\textwidth]{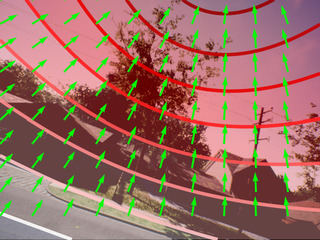}} &
\frame{\includegraphics[width=0.15\textwidth]{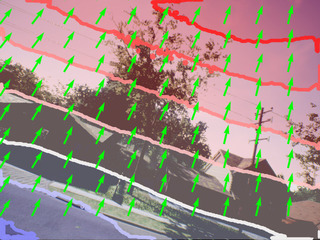}} &
\frame{\includegraphics[width=0.15\textwidth]{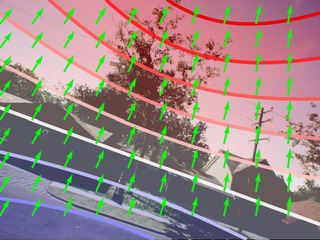}} \\
\frame{\includegraphics[width=0.15\textwidth]{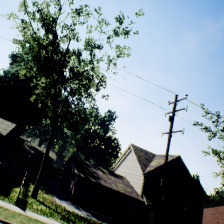}} &
\frame{\includegraphics[width=0.15\textwidth]{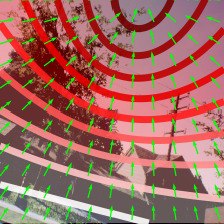}} &
\frame{\includegraphics[width=0.15\textwidth]{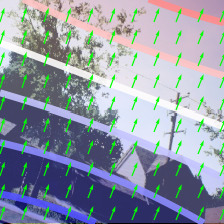}} &
\frame{\includegraphics[width=0.15\textwidth]{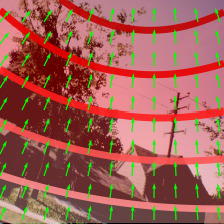}} &
\frame{\includegraphics[width=0.15\textwidth]{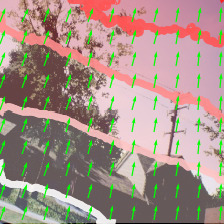}} &
\frame{\includegraphics[width=0.15\textwidth]{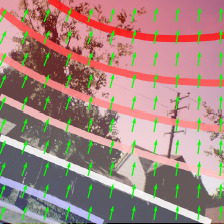}} \\
\frame{\includegraphics[width=0.15\textwidth]{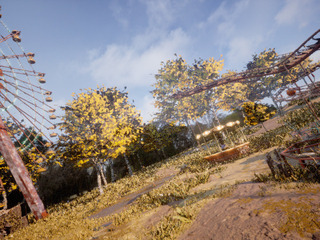}} &
\frame{\includegraphics[width=0.15\textwidth]{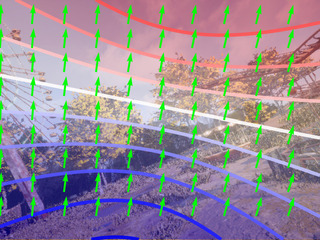}} &
\frame{\includegraphics[width=0.15\textwidth]{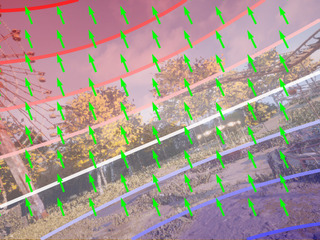}} &
\frame{\includegraphics[width=0.15\textwidth]{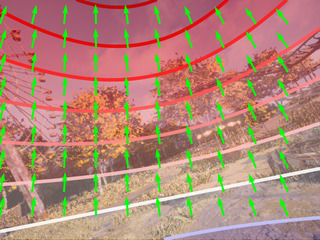}} &
\frame{\includegraphics[width=0.15\textwidth]{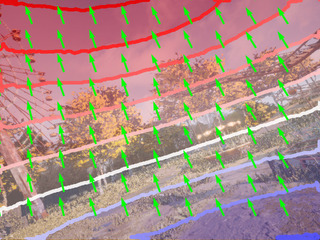}} &
\frame{\includegraphics[width=0.15\textwidth]{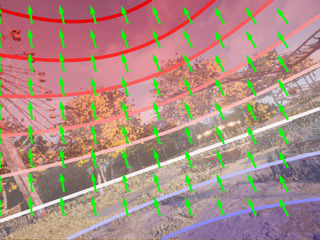}} \\
\frame{\includegraphics[width=0.15\textwidth]{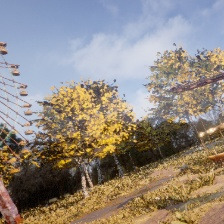}} &
\frame{\includegraphics[width=0.15\textwidth]{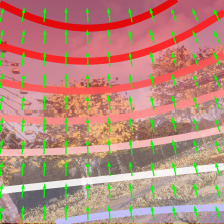}} &
\frame{\includegraphics[width=0.15\textwidth]{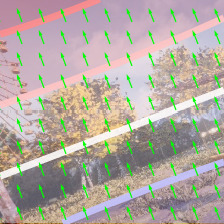}} &
\frame{\includegraphics[width=0.15\textwidth]{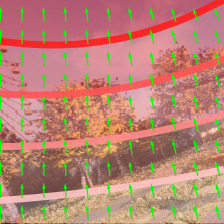}} &
\frame{\includegraphics[width=0.15\textwidth]{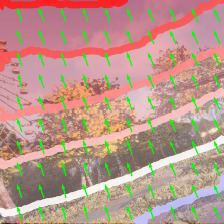}} &
\frame{\includegraphics[width=0.15\textwidth]{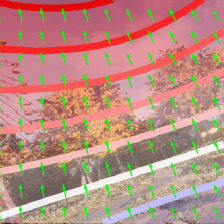}} \\
\frame{\includegraphics[width=0.15\textwidth]{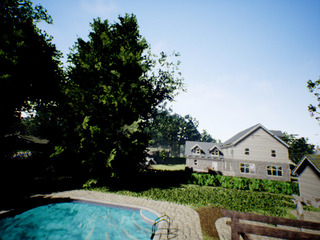}} &
\frame{\includegraphics[width=0.15\textwidth]{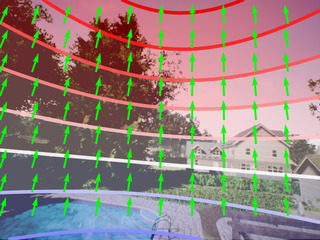}} &
\frame{\includegraphics[width=0.15\textwidth]{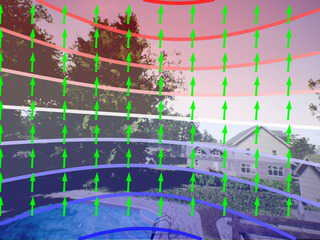}} &
\frame{\includegraphics[width=0.15\textwidth]{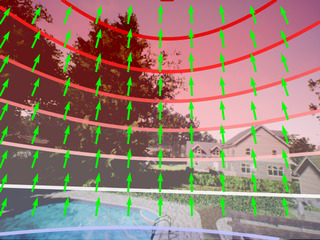}} &
\frame{\includegraphics[width=0.15\textwidth]{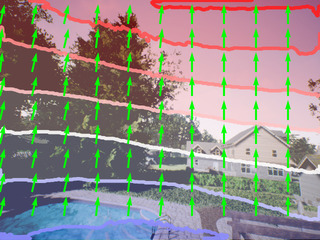}} &
\frame{\includegraphics[width=0.15\textwidth]{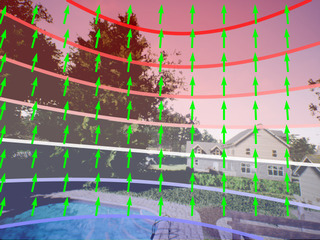}} \\
\frame{\includegraphics[width=0.15\textwidth]{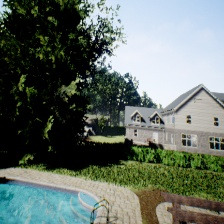}} &
\frame{\includegraphics[width=0.15\textwidth]{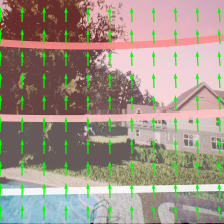}} &
\frame{\includegraphics[width=0.15\textwidth]{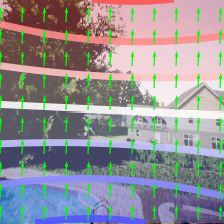}} &
\frame{\includegraphics[width=0.15\textwidth]{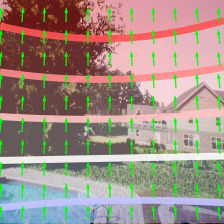}} &
\frame{\includegraphics[width=0.15\textwidth]{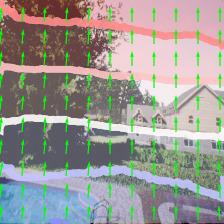}} &
\frame{\includegraphics[width=0.15\textwidth]{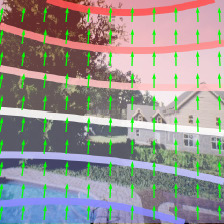}} \\
    \end{tabular}
    }
    \caption{Comparison between baselines on TartanAir dataset. Each test scene has two rows: the first row is the original image with a standard pin-hole camera perspective; the second row is a randomly cropped image. Up-vectors in the green vectors. Latitude colormap: $-\pi/2$ \includegraphics[width=0.4in,height=8pt]{fig/seismic.png} $\pi/2$. }
    \label{fig:supp:densewallcomparison:tartanair}
\end{figure*}

%% file: fig/wild/wild_supp.tex
\begin{figure*}[!t]
    \centering
    \scriptsize
    \vspace{-1em}
    \resizebox{\textwidth}{!}{
    \begin{tabular}{c@{\hskip4pt}c@{\hskip4pt}c@{\hskip4pt}c@{\hskip4pt}c}
    Input & Upright~\cite{lee2014upright} & Percep.~\cite{hold2018perceptual} & CTRL-C~\cite{lee2021ctrl} & Ours \\
    \midrule
    \frame{\includegraphics[width=0.15\textwidth]{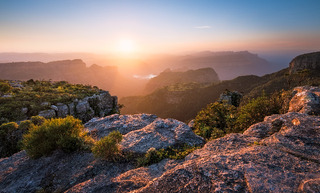}} &
\frame{\includegraphics[width=0.15\textwidth]{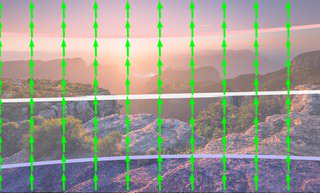}} &
\frame{\includegraphics[width=0.15\textwidth]{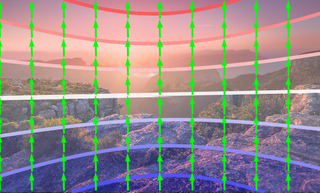}} &
\frame{\includegraphics[width=0.15\textwidth]{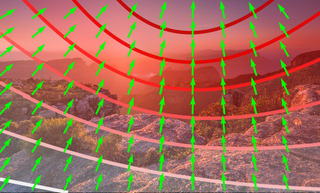}} &
\frame{\includegraphics[width=0.15\textwidth]{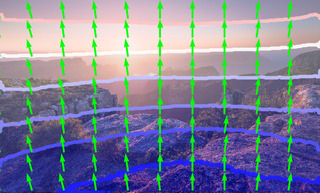}} \\
\frame{\includegraphics[width=0.15\textwidth]{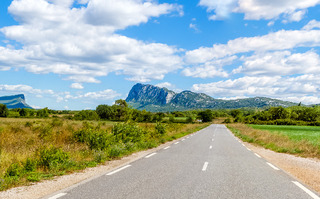}} &
\frame{\includegraphics[width=0.15\textwidth]{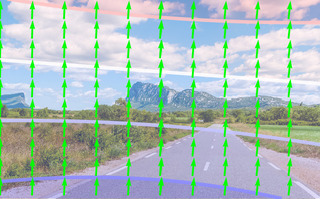}} &
\frame{\includegraphics[width=0.15\textwidth]{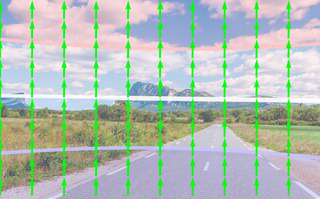}} &
\frame{\includegraphics[width=0.15\textwidth]{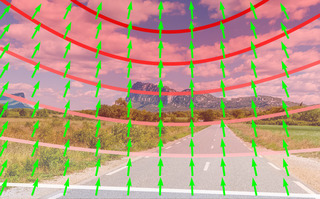}} &
\frame{\includegraphics[width=0.15\textwidth]{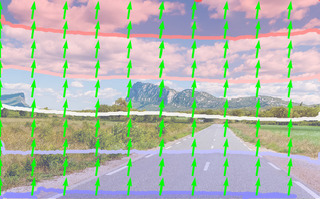}} \\
\frame{\includegraphics[width=0.15\textwidth]{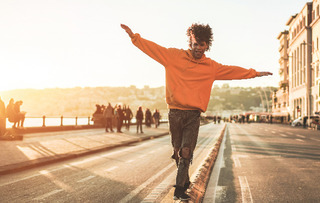}} &
\frame{\includegraphics[width=0.15\textwidth]{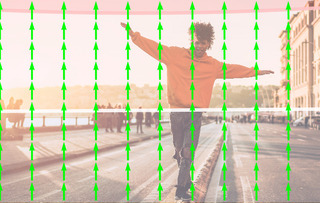}} &
\frame{\includegraphics[width=0.15\textwidth]{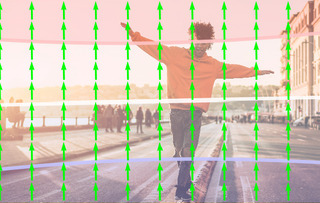}} &
\frame{\includegraphics[width=0.15\textwidth]{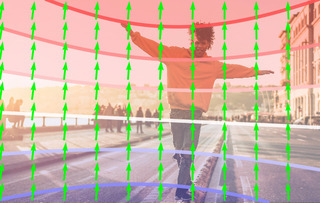}} &
\frame{\includegraphics[width=0.15\textwidth]{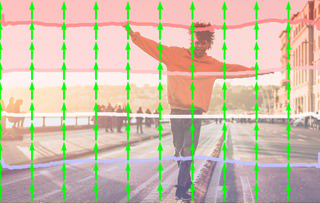}} \\
\frame{\includegraphics[width=0.15\textwidth]{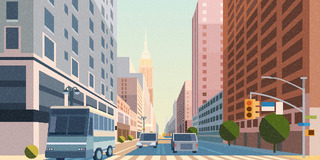}} &
\frame{\includegraphics[width=0.15\textwidth]{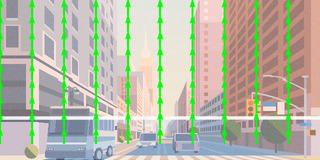}} &
\frame{\includegraphics[width=0.15\textwidth]{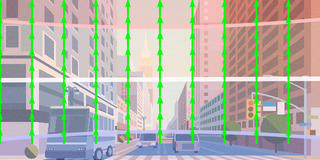}} &
\frame{\includegraphics[width=0.15\textwidth]{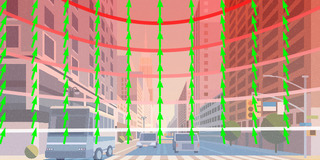}} &
\frame{\includegraphics[width=0.15\textwidth]{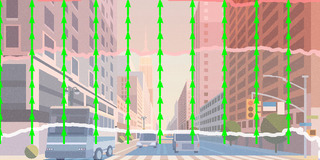}} \\
\frame{\includegraphics[width=0.15\textwidth]{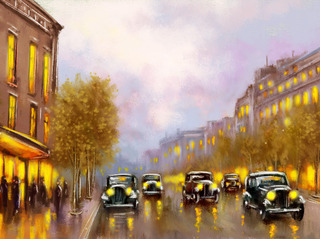}} &
\frame{\includegraphics[width=0.15\textwidth]{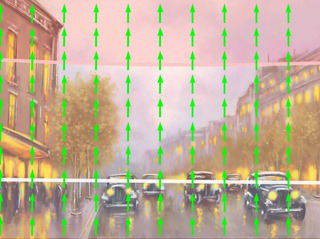}} &
\frame{\includegraphics[width=0.15\textwidth]{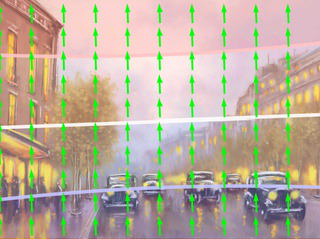}} &
\frame{\includegraphics[width=0.15\textwidth]{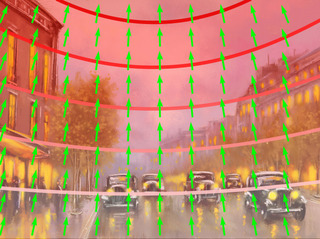}} &
\frame{\includegraphics[width=0.15\textwidth]{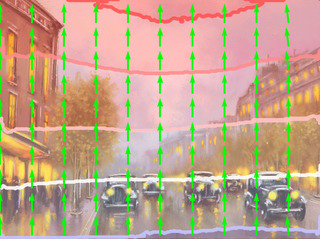}} \\
    \frame{\includegraphics[width=0.15\textwidth]{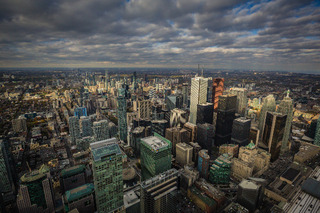}} &
    \frame{\includegraphics[width=0.15\textwidth]{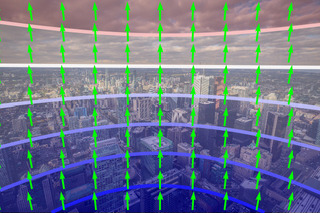}} &
    \frame{\includegraphics[width=0.15\textwidth]{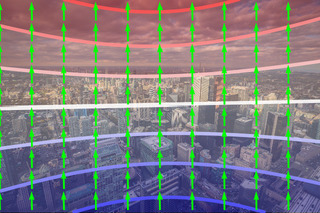}} &
    \frame{\includegraphics[width=0.15\textwidth]{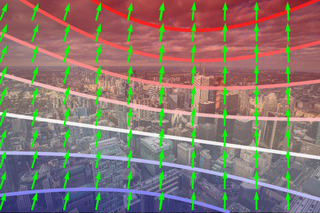}} &
    \frame{\includegraphics[width=0.15\textwidth]{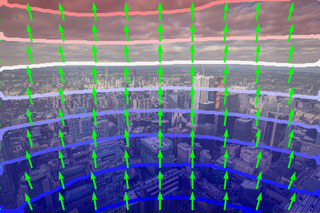}} \\
    \frame{\includegraphics[width=0.15\textwidth]{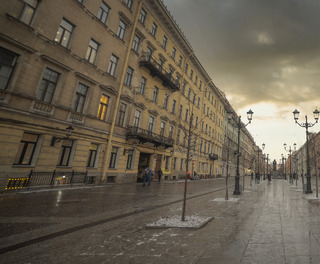}} &
    \frame{\includegraphics[width=0.15\textwidth]{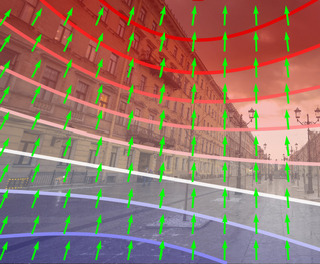}} &
    \frame{\includegraphics[width=0.15\textwidth]{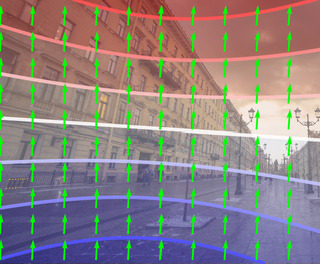}} &
    \frame{\includegraphics[width=0.15\textwidth]{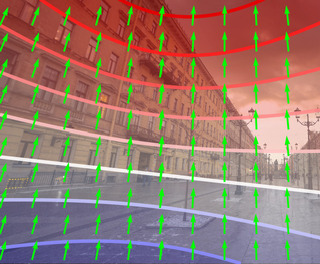}} &
    \frame{\includegraphics[width=0.15\textwidth]{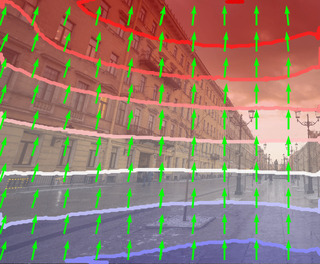}} \\
\frame{\includegraphics[width=0.15\textwidth]{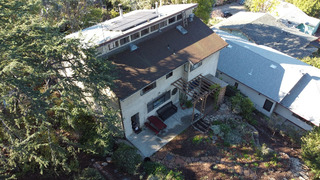}} &
\frame{\includegraphics[width=0.15\textwidth]{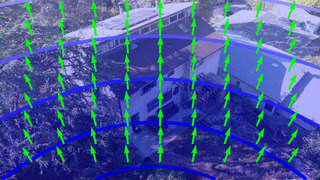}} &
\frame{\includegraphics[width=0.15\textwidth]{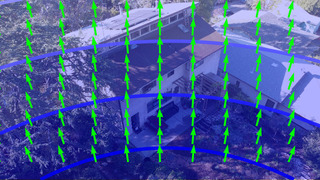}} &
\frame{\includegraphics[width=0.15\textwidth]{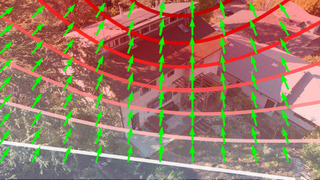}} &
\frame{\includegraphics[width=0.15\textwidth]{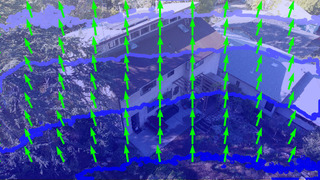}} \\
\frame{\includegraphics[width=0.15\textwidth]{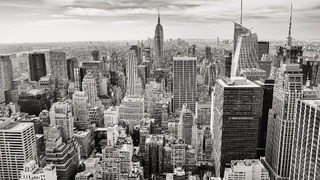}} &
\frame{\includegraphics[width=0.15\textwidth]{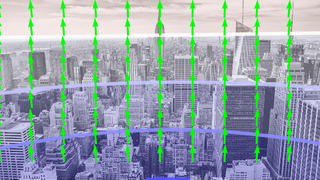}} &
\frame{\includegraphics[width=0.15\textwidth]{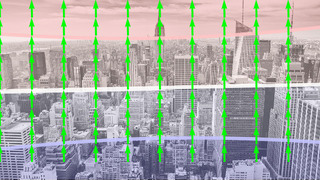}} &
\frame{\includegraphics[width=0.15\textwidth]{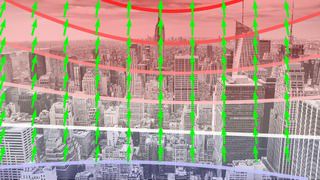}} &
\frame{\includegraphics[width=0.15\textwidth]{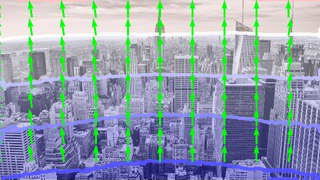}} \\
\frame{\includegraphics[width=0.15\textwidth]{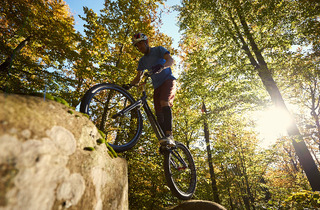}} &
\frame{\includegraphics[width=0.15\textwidth]{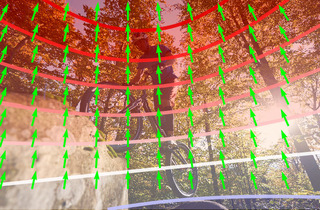}} &
\frame{\includegraphics[width=0.15\textwidth]{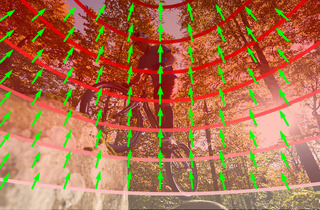}} &
\frame{\includegraphics[width=0.15\textwidth]{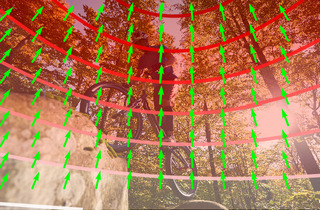}} &
\frame{\includegraphics[width=0.15\textwidth]{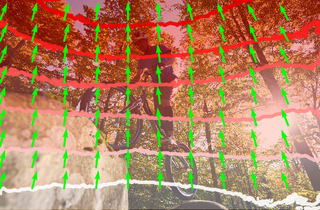}} \\

    \end{tabular}
    }
    \caption{Additional qualitative results on web images, extending Fig. 5. Our approach produces better results compared to~\cite{lee2014upright}, \cite{hold2018perceptual}, and~\cite{lee2021ctrl}.  There is no
ground truth available.
    }
    \label{fig:wild_supp}
\end{figure*}

%% file: fig/wild/wild_supp_obj.tex
\begin{figure*}[!t]
    \centering
    \scriptsize
    \vspace{-1em}
    \resizebox{\textwidth}{!}{
    \begin{tabular}{c@{\hskip4pt}c@{\hskip4pt}c@{\hskip4pt}c@{\hskip4pt}c}
    Input & Upright~\cite{lee2014upright} & Percep.~\cite{hold2018perceptual} & CTRL-C~\cite{lee2021ctrl} & Ours-Distill \\
    \midrule 
    \frame{\includegraphics[width=0.15\textwidth]{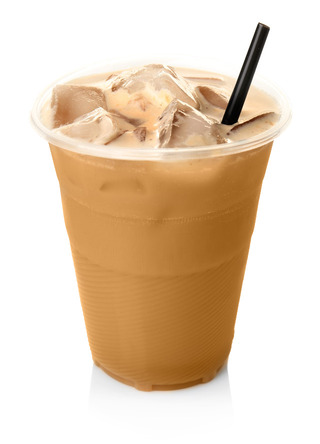}} &
    \frame{\includegraphics[width=0.15\textwidth]{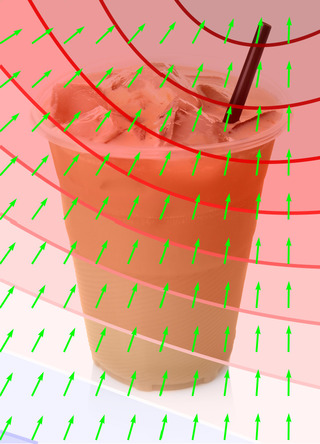}} &
    \frame{\includegraphics[width=0.15\textwidth]{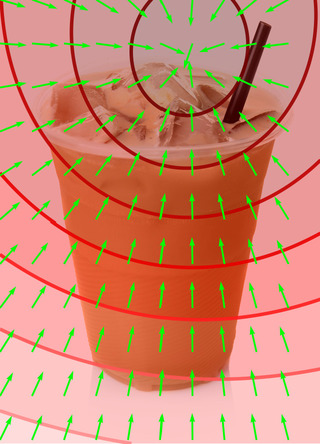}} &
    \frame{\includegraphics[width=0.15\textwidth]{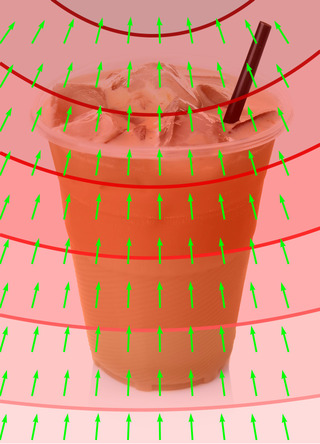}} &
    \frame{\includegraphics[width=0.15\textwidth]{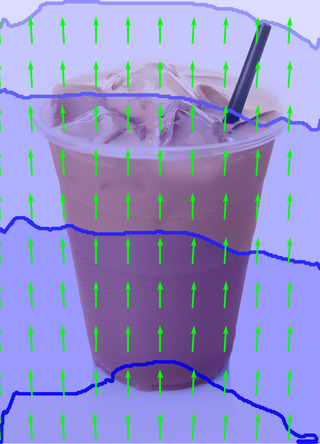}} \\
    \frame{\includegraphics[width=0.15\textwidth]{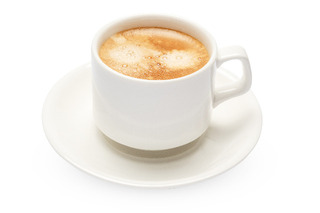}} &
    \frame{\includegraphics[width=0.15\textwidth]{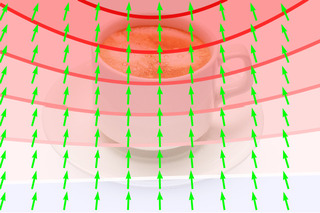}} &
    \frame{\includegraphics[width=0.15\textwidth]{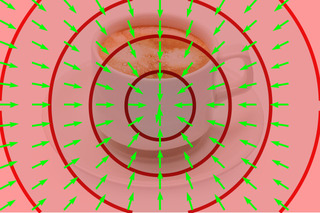}} &
    \frame{\includegraphics[width=0.15\textwidth]{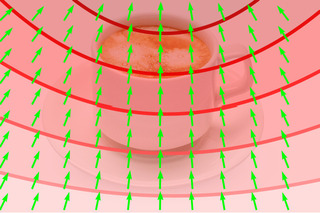}} &
    \frame{\includegraphics[width=0.15\textwidth]{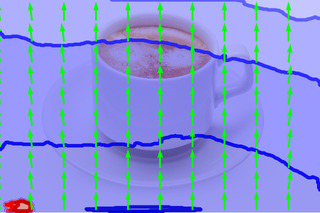}} \\
    \frame{\includegraphics[width=0.15\textwidth]{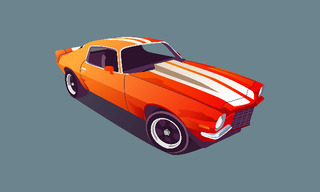}} &
    \frame{\includegraphics[width=0.15\textwidth]{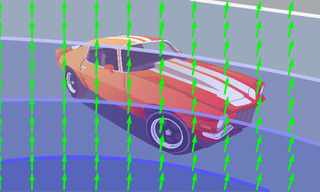}} &
    \frame{\includegraphics[width=0.15\textwidth]{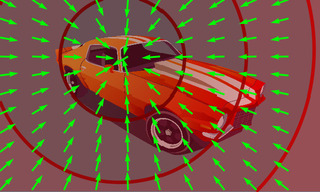}} &
    \frame{\includegraphics[width=0.15\textwidth]{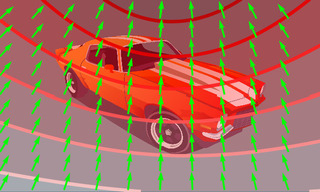}} &
    \frame{\includegraphics[width=0.15\textwidth]{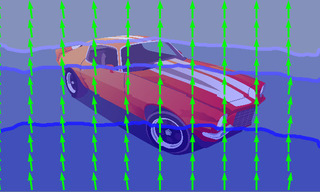}} \\
    \frame{\includegraphics[width=0.15\textwidth]{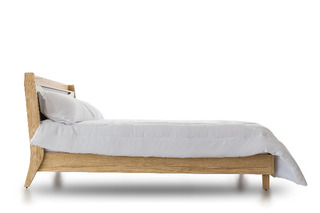}} &
    \frame{\includegraphics[width=0.15\textwidth]{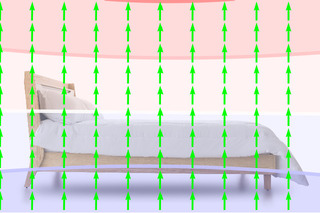}} &
    \frame{\includegraphics[width=0.15\textwidth]{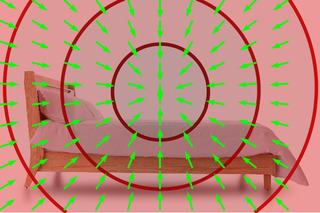}} &
    \frame{\includegraphics[width=0.15\textwidth]{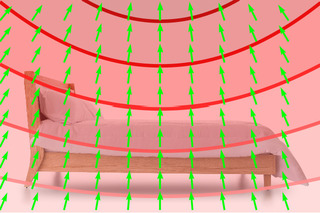}} &
    \frame{\includegraphics[width=0.15\textwidth]{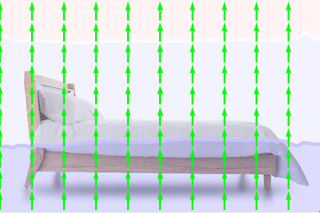}} \\
    \frame{\includegraphics[width=0.15\textwidth]{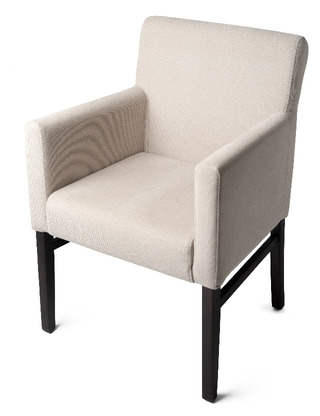}} &
    \frame{\includegraphics[width=0.15\textwidth]{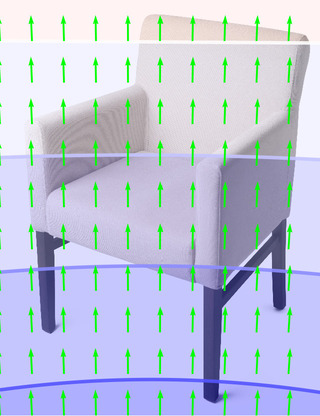}} &
    \frame{\includegraphics[width=0.15\textwidth]{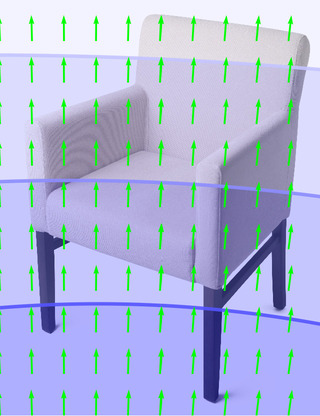}} &
    \frame{\includegraphics[width=0.15\textwidth]{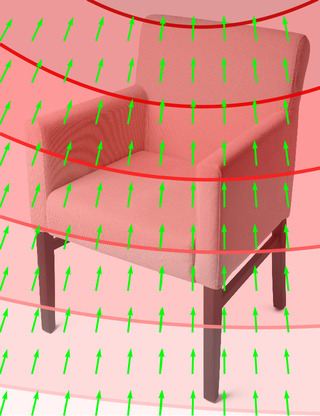}} &
    \frame{\includegraphics[width=0.15\textwidth]{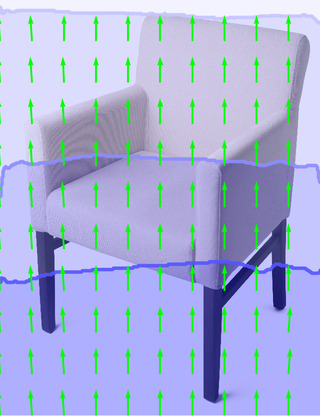}} \\
    \frame{\includegraphics[width=0.15\textwidth]{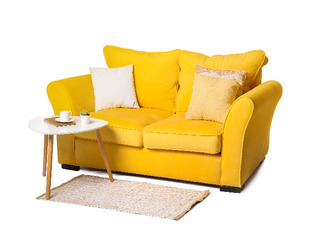}} &
    \frame{\includegraphics[width=0.15\textwidth]{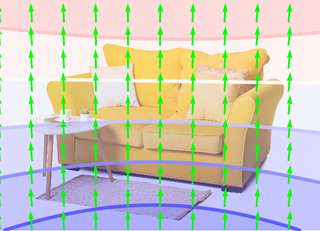}} &
    \frame{\includegraphics[width=0.15\textwidth]{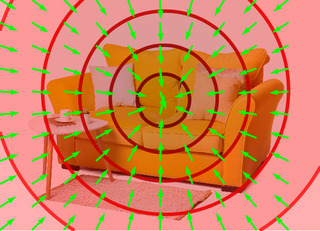}} &
    \frame{\includegraphics[width=0.15\textwidth]{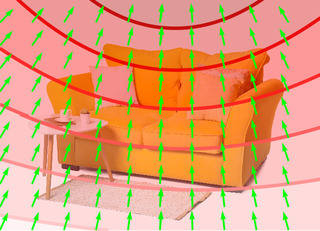}} &
    \frame{\includegraphics[width=0.15\textwidth]{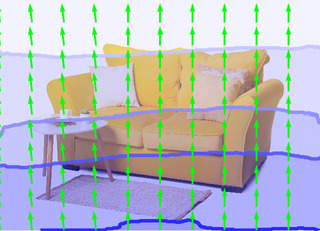}} \\

    \frame{\includegraphics[width=0.15\textwidth]{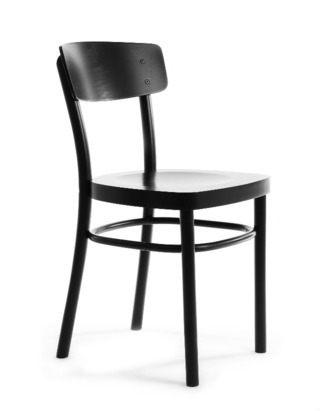}} &
    \frame{\includegraphics[width=0.15\textwidth]{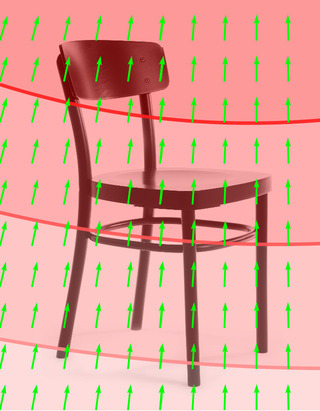}} &
    \frame{\includegraphics[width=0.15\textwidth]{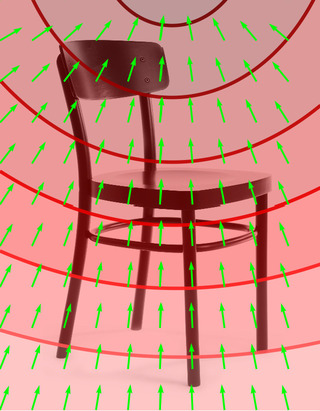}} &
    \frame{\includegraphics[width=0.15\textwidth]{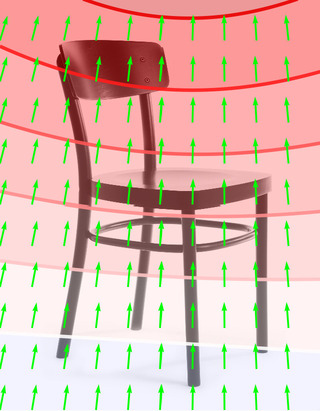}} &
    \frame{\includegraphics[width=0.15\textwidth]{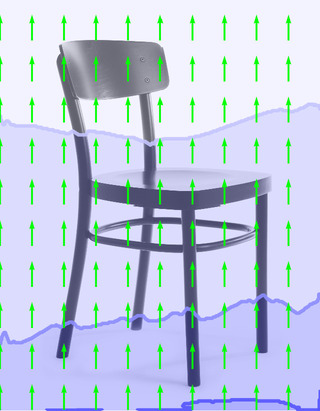}} \\
    \end{tabular}
    }
    \caption{Additional qualitative results on web images, extending Fig. 5. Our approach produces better results compared to~\cite{lee2014upright}, \cite{hold2018perceptual}, and~\cite{lee2021ctrl}.
    }
    \label{fig:wild_supp_obj}
\end{figure*}

%% file: fig/supp_fisheye_wall.tex
\begin{figure*}[h]
\centering
\scriptsize
\resizebox{\textwidth}{!}{
\begin{tabular}{c@{\hskip4pt}c@{\hskip4pt}cc@{\hskip4pt}c@{\hskip4pt}c}
    Input & Sliding Win. & Fine-tune & Input & Sliding Win. & Fine-tune \\
\midrule 
    
\frame{\includegraphics[width=0.2\textwidth]{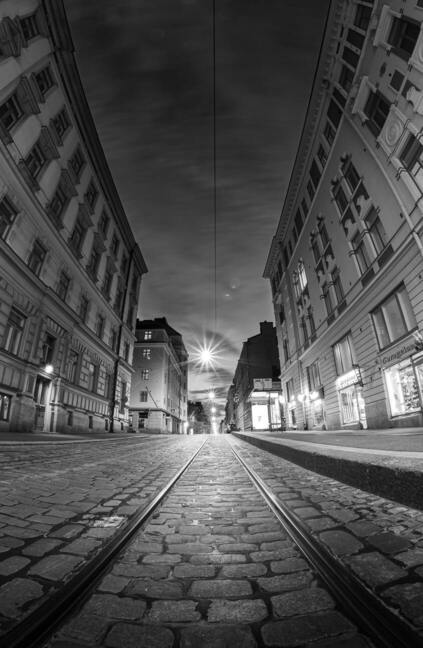}} &
\frame{\includegraphics[width=0.2\textwidth]{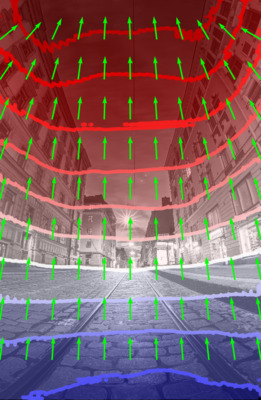}} &
\frame{\includegraphics[width=0.2\textwidth]{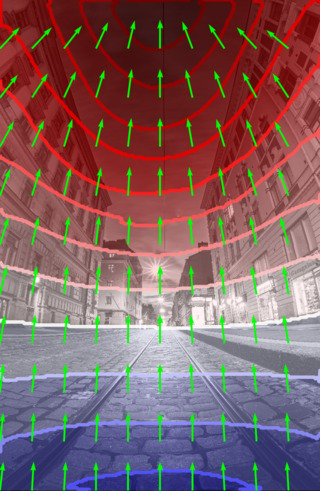}} &
\frame{\includegraphics[width=0.2\textwidth]{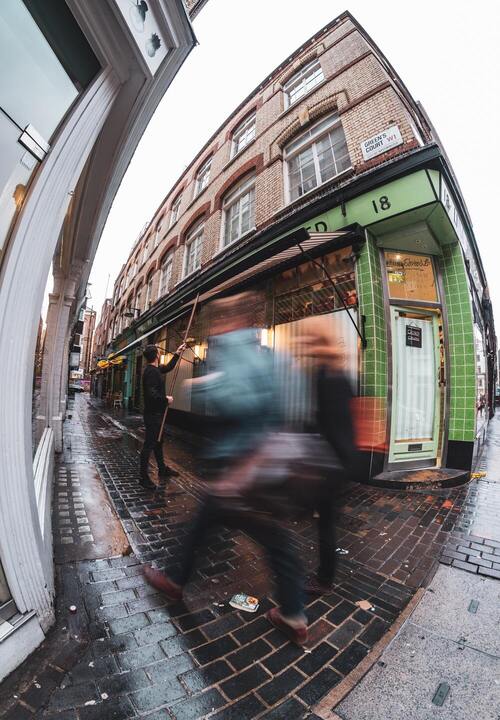}} &
\frame{\includegraphics[width=0.2\textwidth]{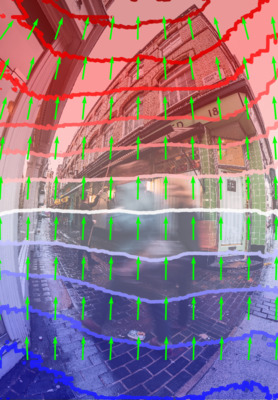}} &
\frame{\includegraphics[width=0.2\textwidth]{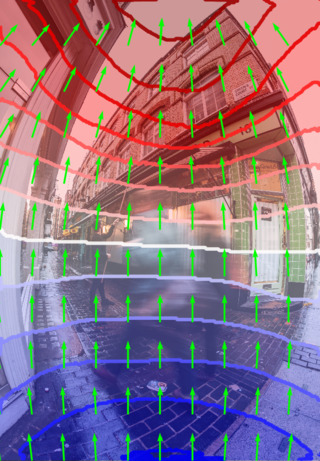}}\\
\frame{\includegraphics[width=0.2\textwidth]{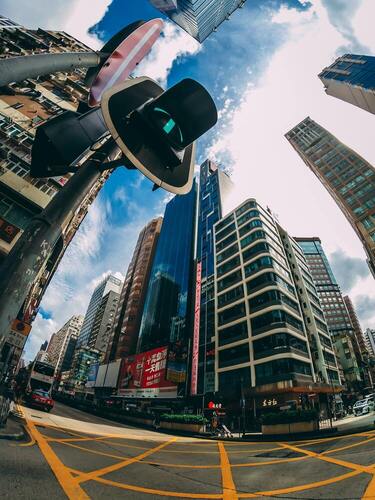}} &
\frame{\includegraphics[width=0.2\textwidth]{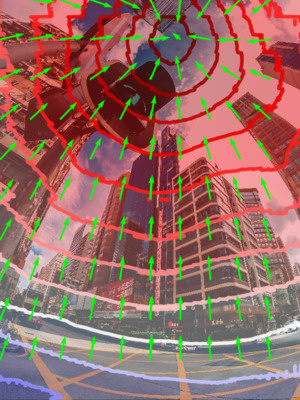}} &
\frame{\includegraphics[width=0.2\textwidth]{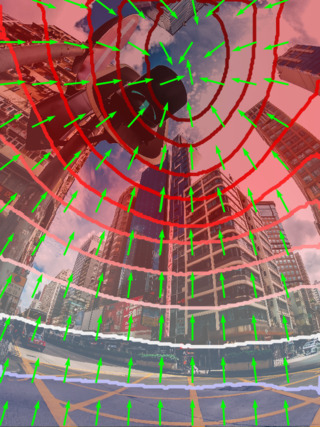}} &
\frame{\includegraphics[width=0.2\textwidth]{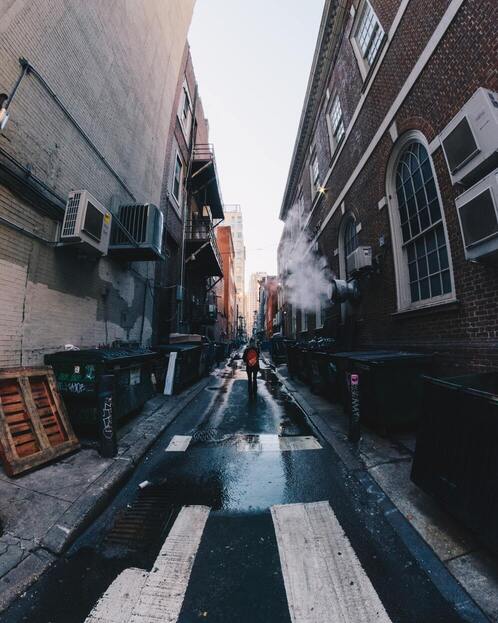}} &
\frame{\includegraphics[width=0.2\textwidth]{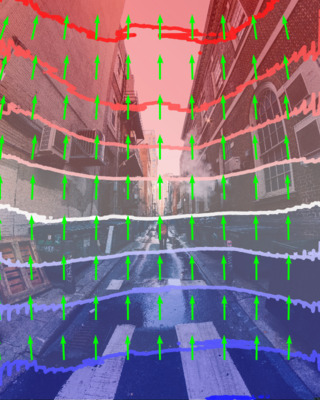}} &
\frame{\includegraphics[width=0.2\textwidth]{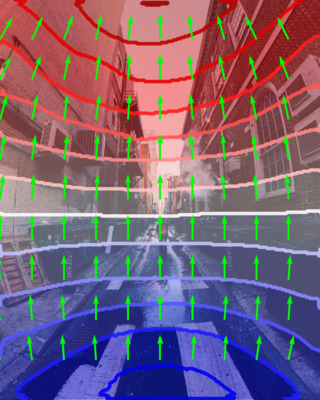}}  \\
\frame{\includegraphics[width=0.2\textwidth]{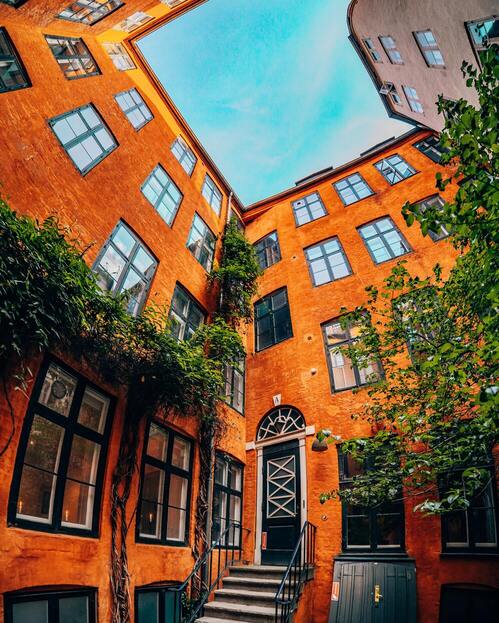}} &
\frame{\includegraphics[width=0.2\textwidth]{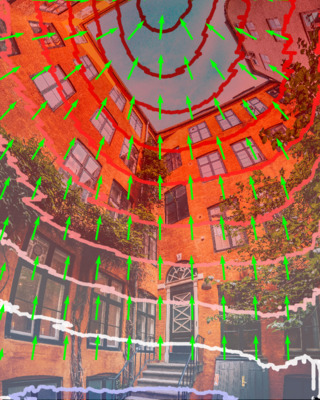}} &
\frame{\includegraphics[width=0.2\textwidth]{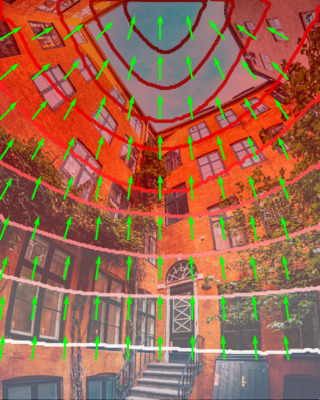}} &
\frame{\includegraphics[width=0.2\textwidth]{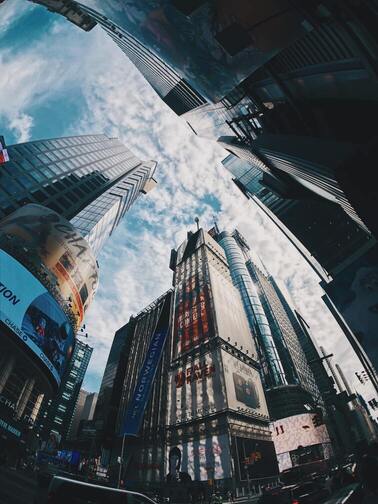}} &
\frame{\includegraphics[width=0.2\textwidth]{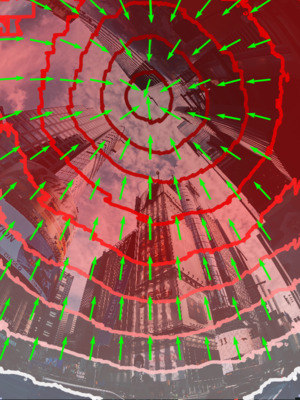}} &
\frame{\includegraphics[width=0.2\textwidth]{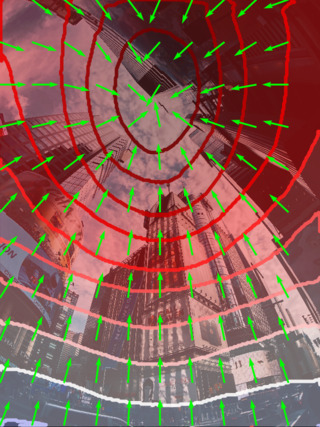}} \\
\frame{\includegraphics[width=0.2\textwidth]{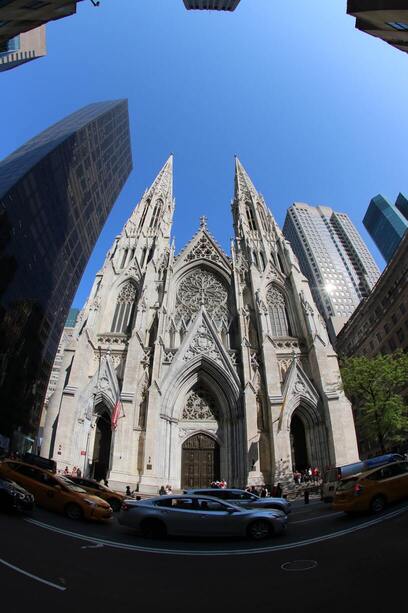}} &
\frame{\includegraphics[width=0.2\textwidth]{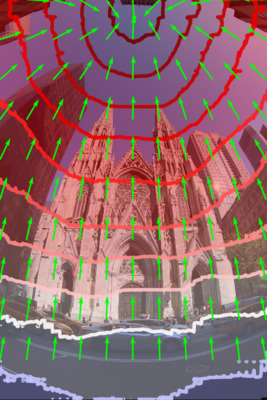}} &
\frame{\includegraphics[width=0.2\textwidth]{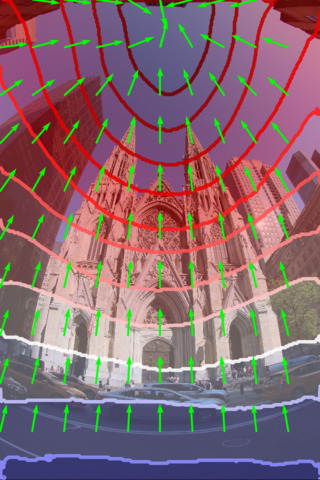}} &
\frame{\includegraphics[width=0.2\textwidth]{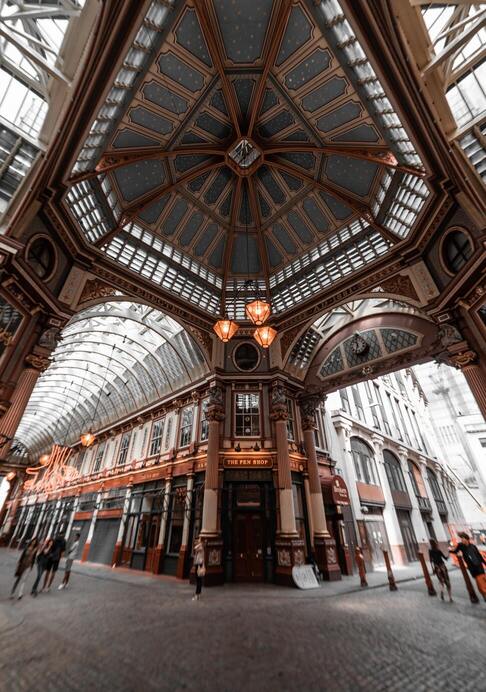}} &
\frame{\includegraphics[width=0.2\textwidth]{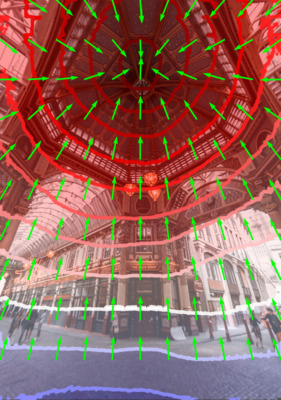}} &
\frame{\includegraphics[width=0.2\textwidth]{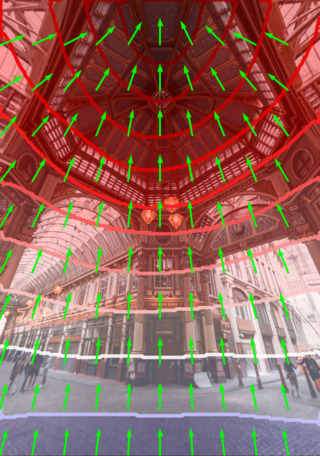}} \\
\frame{\includegraphics[width=0.2\textwidth]{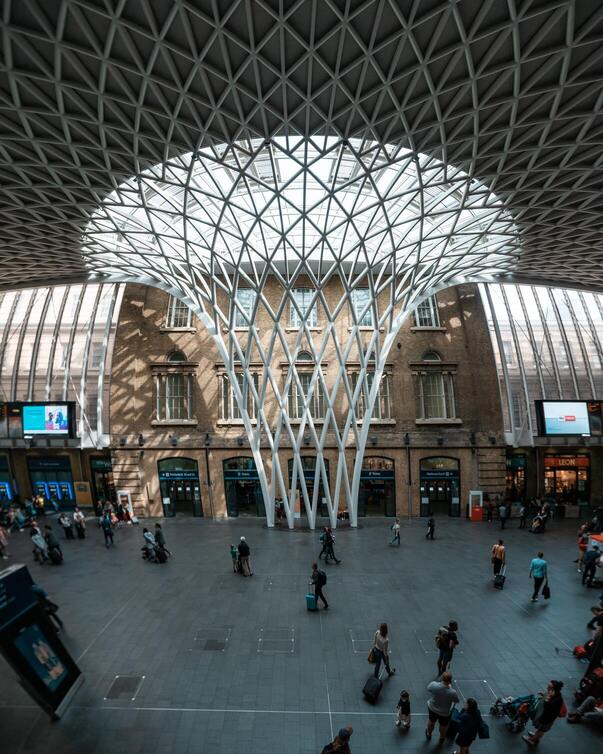}} &
\frame{\includegraphics[width=0.2\textwidth]{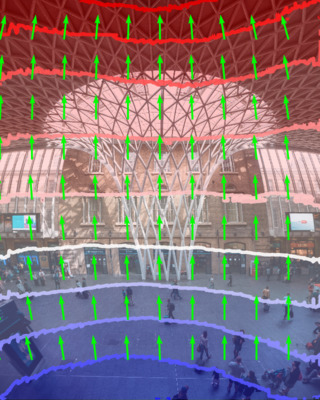}} &
\frame{\includegraphics[width=0.2\textwidth]{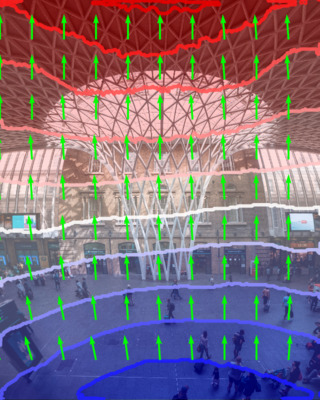}} &
\frame{\includegraphics[width=0.2\textwidth]{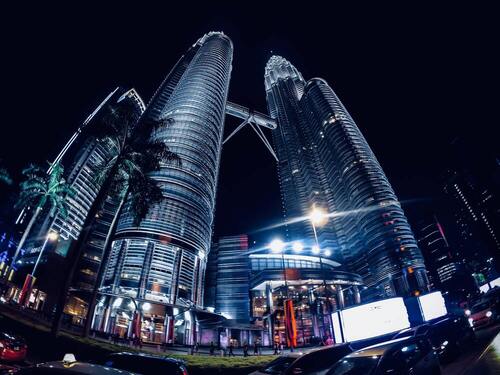}} &
\frame{\includegraphics[width=0.2\textwidth]{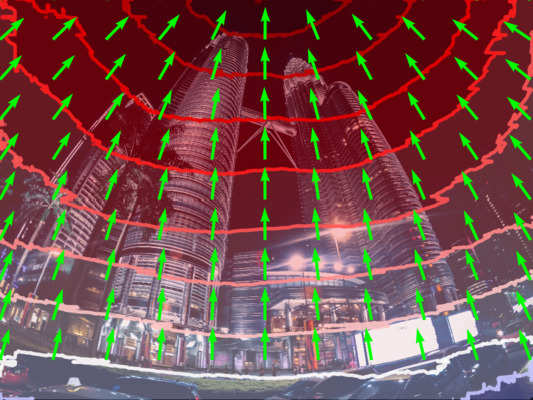}} &
\frame{\includegraphics[width=0.2\textwidth]{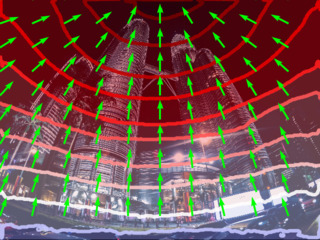}} \\

    \end{tabular}
    }
    
    \caption{Qualitative results on fisheye images from the wild using both the sliding window and fine-tune techniques. Up-vectors in the green vectors. Latitude colormap: $-\pi/2$ \includegraphics[width=0.4in,height=8pt]{fig/seismic.png} $\pi/2$. }
    \label{fig:supp:fisheye}
\end{figure*}

%% file: fig/supp_pers.tex
\begin{figure*}[h]
\centering
\resizebox{\textwidth}{!}{
\begin{tabular}{c@{\hskip4pt}c@{\hskip4pt}c@{\hskip4pt}cc@{\hskip4pt}c@{\hskip4pt}c@{\hskip4pt}c}
    Input & PerspectiveNet & + ParamNet & Ground Truth & Input & PerspectiveNet & + ParamNet & GroundTruth \\
\midrule 
    
\frame{\includegraphics[width=0.2\textwidth]{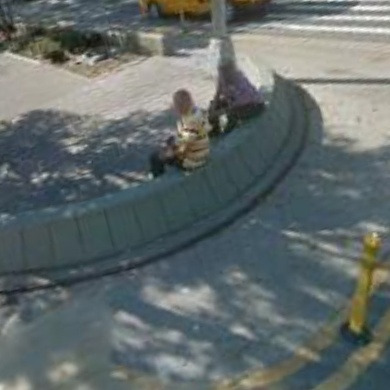}} &
\frame{\includegraphics[width=0.2\textwidth]{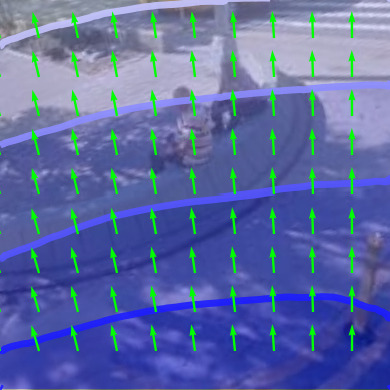}} &
\frame{\includegraphics[width=0.2\textwidth]{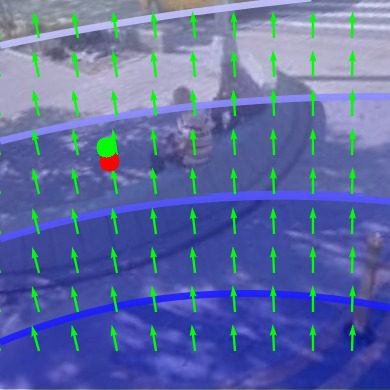}} &
\frame{\includegraphics[width=0.2\textwidth]{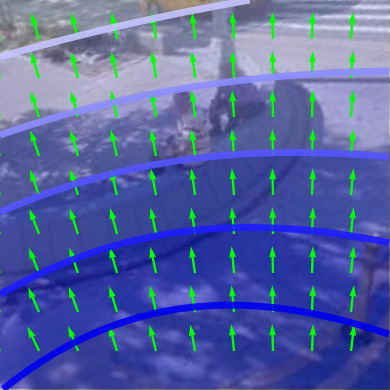}} &
\frame{\includegraphics[width=0.2\textwidth]{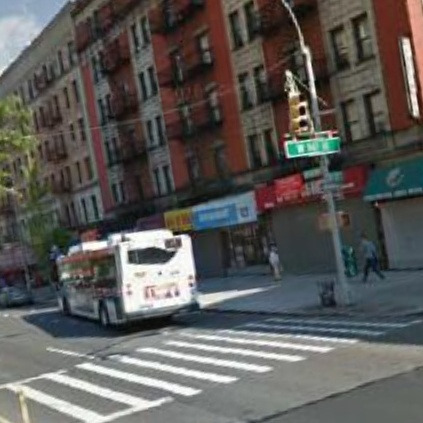}} &
\frame{\includegraphics[width=0.2\textwidth]{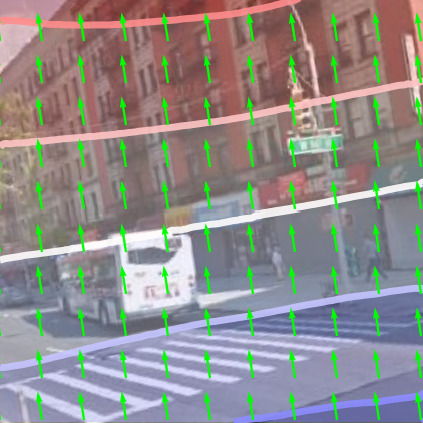}} &
\frame{\includegraphics[width=0.2\textwidth]{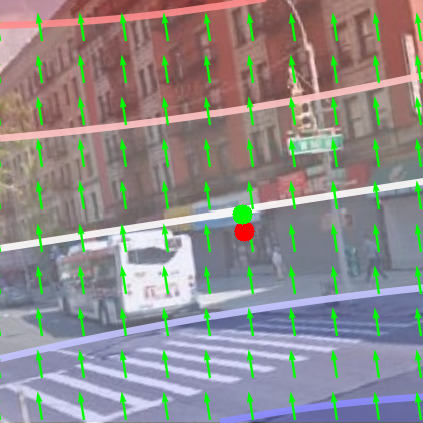}} &
\frame{\includegraphics[width=0.2\textwidth]{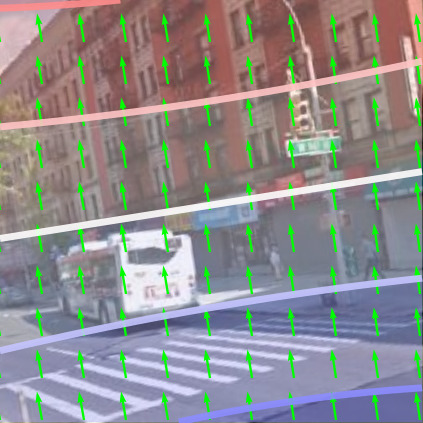}}\\

\frame{\includegraphics[width=0.2\textwidth]{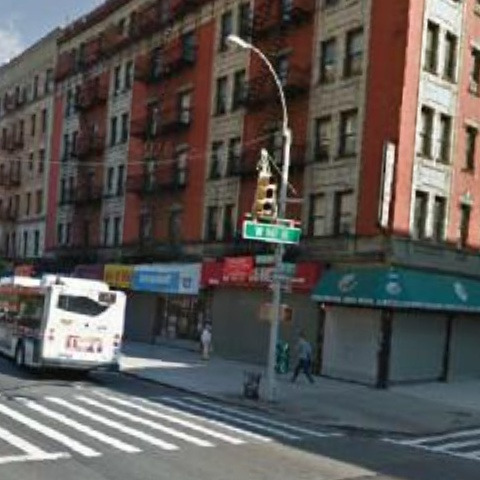}} &
\frame{\includegraphics[width=0.2\textwidth]{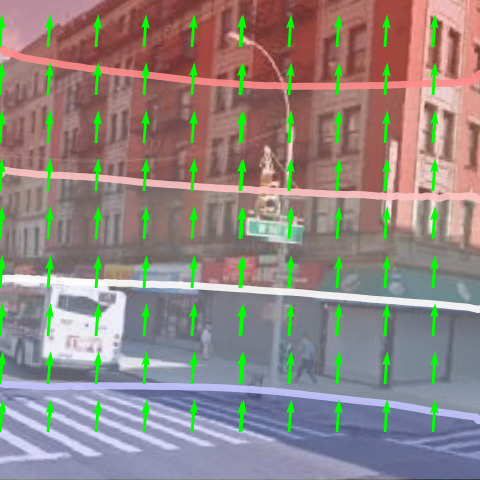}} &
\frame{\includegraphics[width=0.2\textwidth]{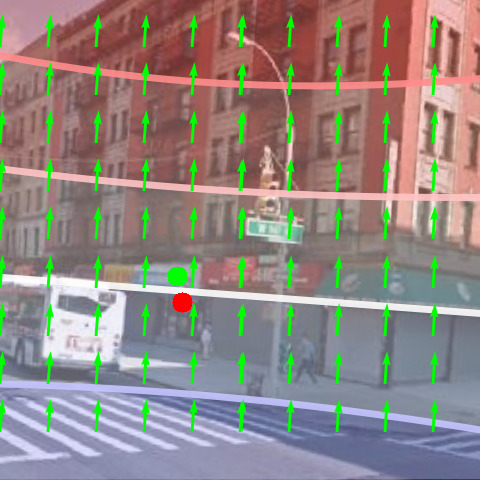}} &
\frame{\includegraphics[width=0.2\textwidth]{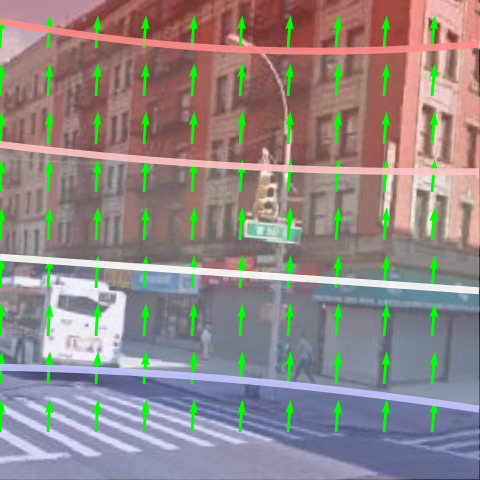}} &
\frame{\includegraphics[width=0.2\textwidth]{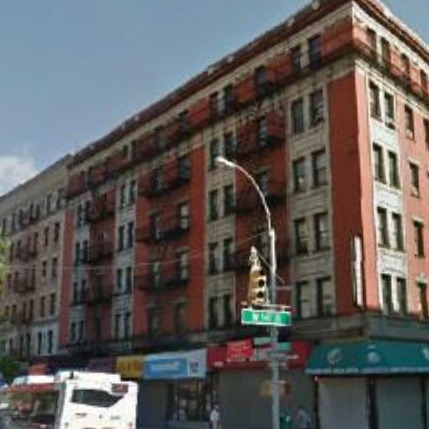}} &
\frame{\includegraphics[width=0.2\textwidth]{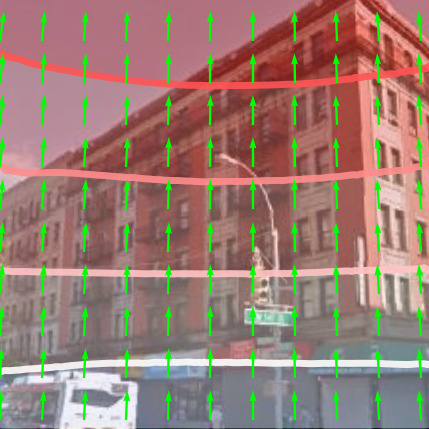}} &
\frame{\includegraphics[width=0.2\textwidth]{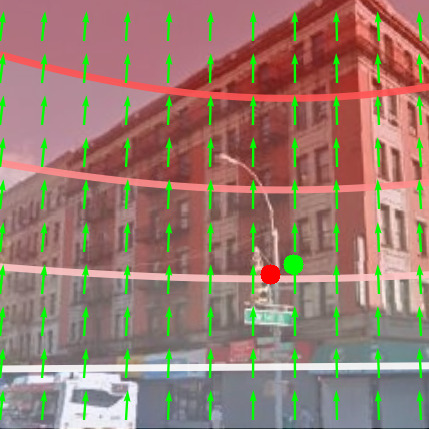}} &
\frame{\includegraphics[width=0.2\textwidth]{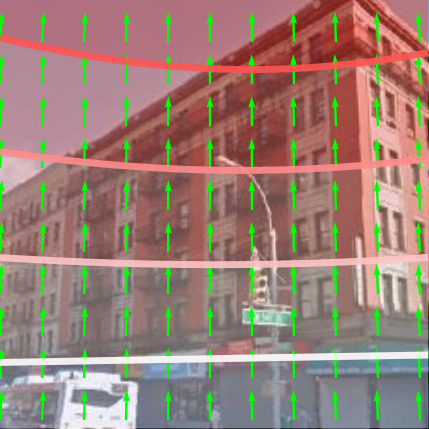}}\\

\frame{\includegraphics[width=0.2\textwidth]{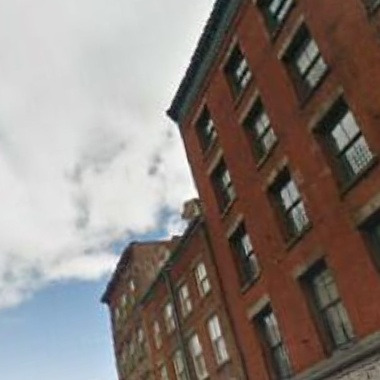}} &
\frame{\includegraphics[width=0.2\textwidth]{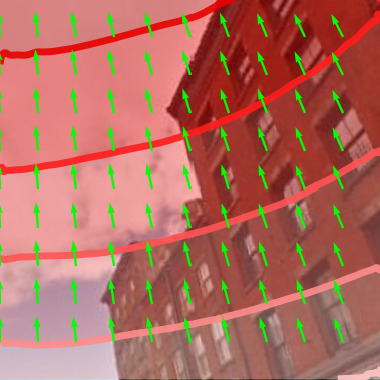}} &
\frame{\includegraphics[width=0.2\textwidth]{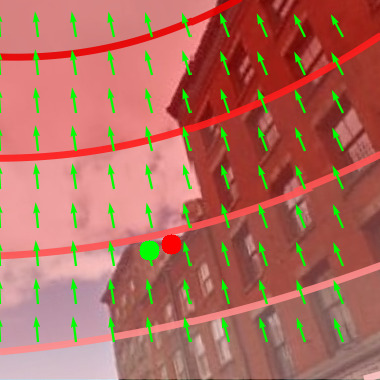}} &
\frame{\includegraphics[width=0.2\textwidth]{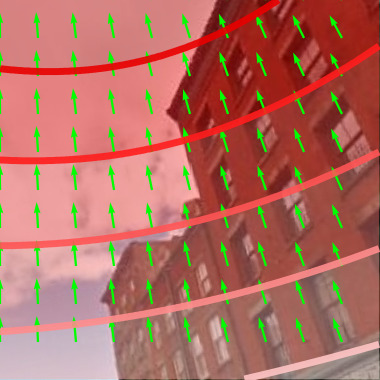}} &
\frame{\includegraphics[width=0.2\textwidth]{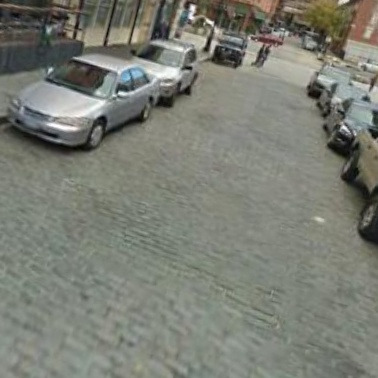}} &
\frame{\includegraphics[width=0.2\textwidth]{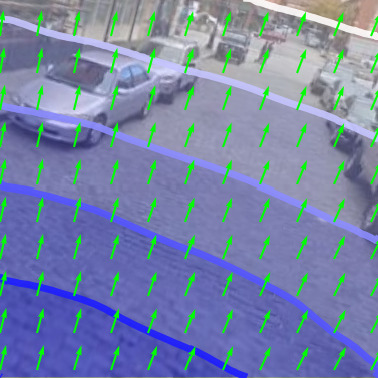}} &
\frame{\includegraphics[width=0.2\textwidth]{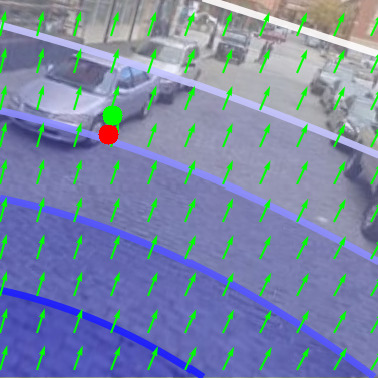}} &
\frame{\includegraphics[width=0.2\textwidth]{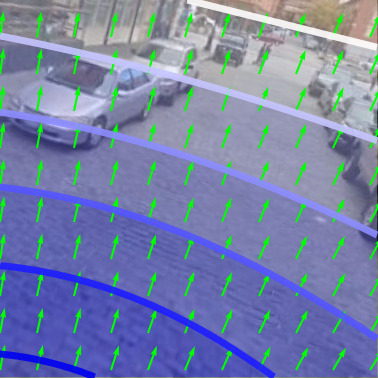}} \\

\frame{\includegraphics[width=0.2\textwidth]{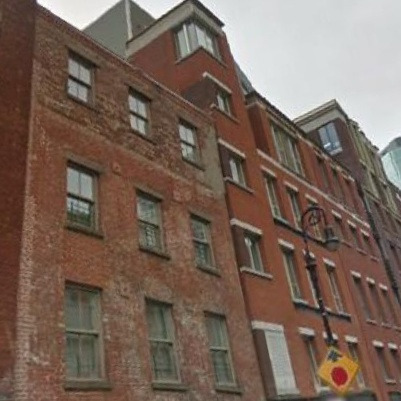}} &
\frame{\includegraphics[width=0.2\textwidth]{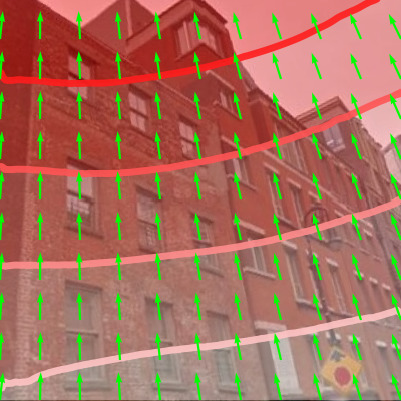}} &
\frame{\includegraphics[width=0.2\textwidth]{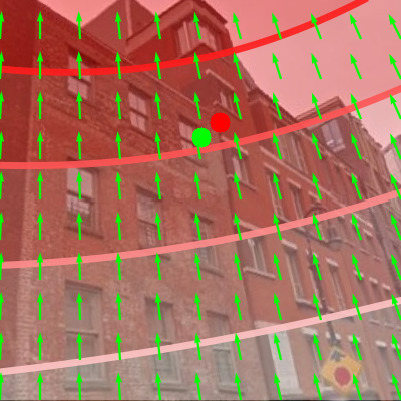}} &
\frame{\includegraphics[width=0.2\textwidth]{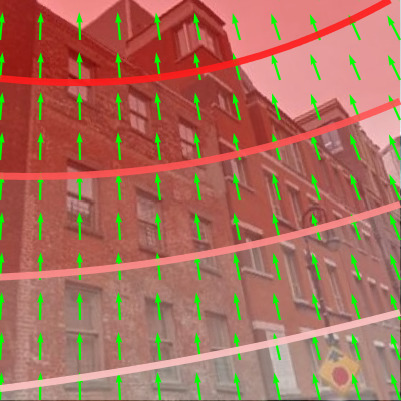}} &
\frame{\includegraphics[width=0.2\textwidth]{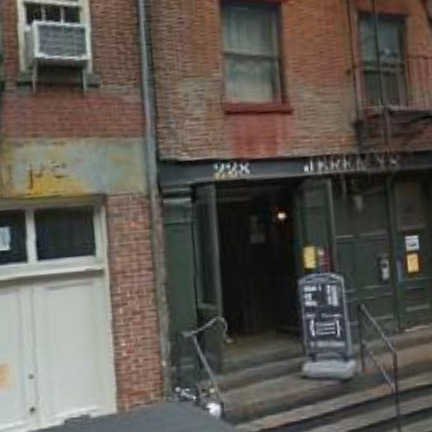}} &
\frame{\includegraphics[width=0.2\textwidth]{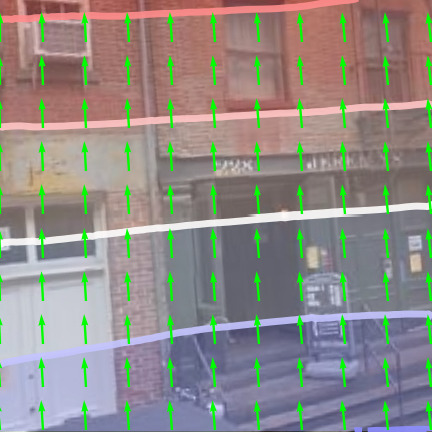}} &
\frame{\includegraphics[width=0.2\textwidth]{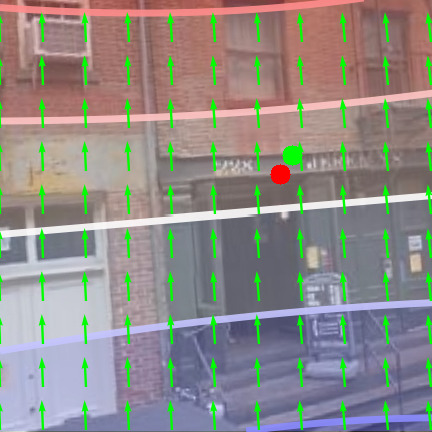}} &
\frame{\includegraphics[width=0.2\textwidth]{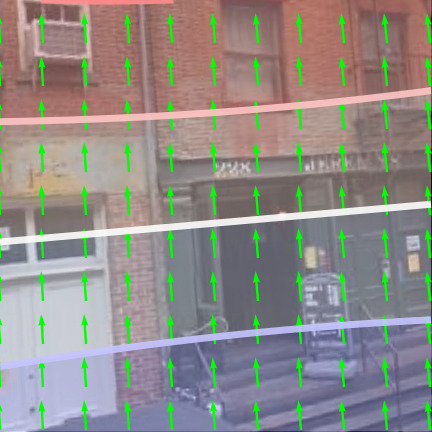}}\\

\frame{\includegraphics[width=0.2\textwidth]{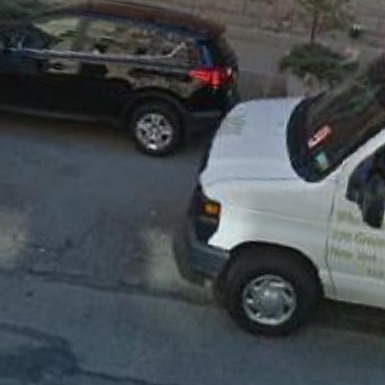}} &
\frame{\includegraphics[width=0.2\textwidth]{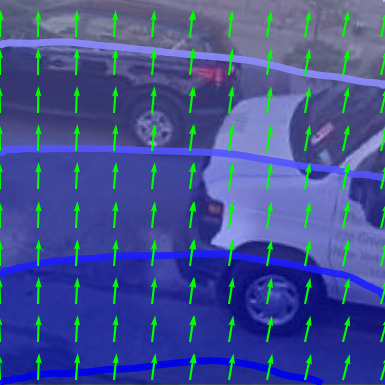}} &
\frame{\includegraphics[width=0.2\textwidth]{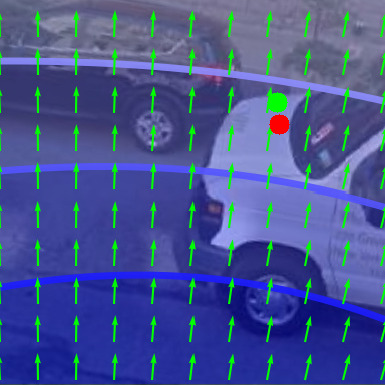}} &
\frame{\includegraphics[width=0.2\textwidth]{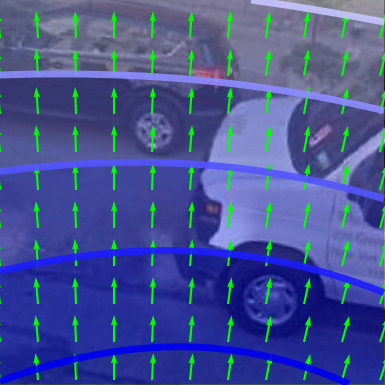}} &
\frame{\includegraphics[width=0.2\textwidth]{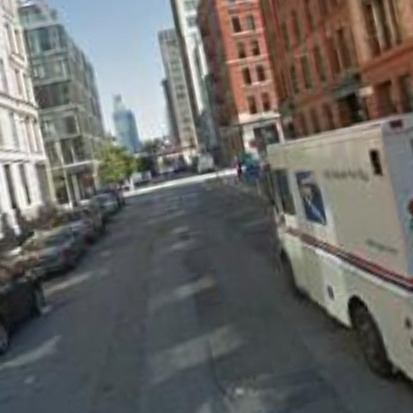}} &
\frame{\includegraphics[width=0.2\textwidth]{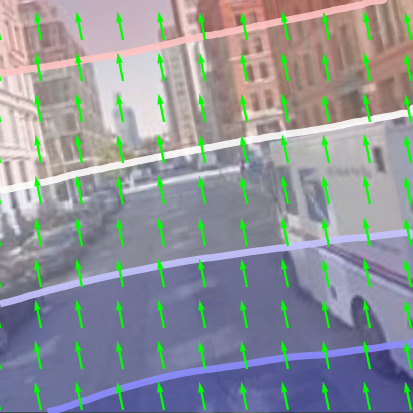}} &
\frame{\includegraphics[width=0.2\textwidth]{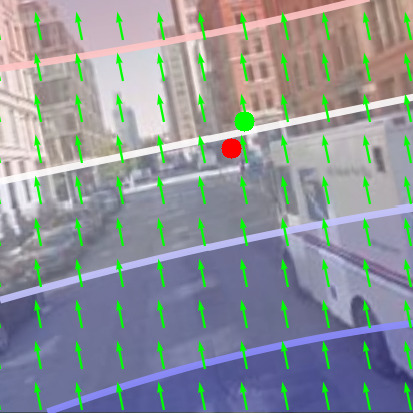}} &
\frame{\includegraphics[width=0.2\textwidth]{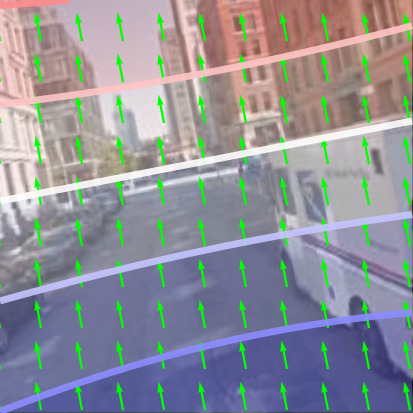}}\\

\frame{\includegraphics[width=0.2\textwidth]{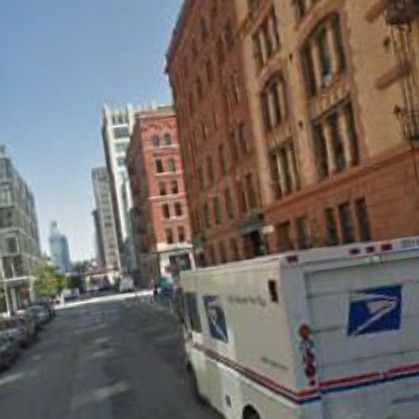}} &
\frame{\includegraphics[width=0.2\textwidth]{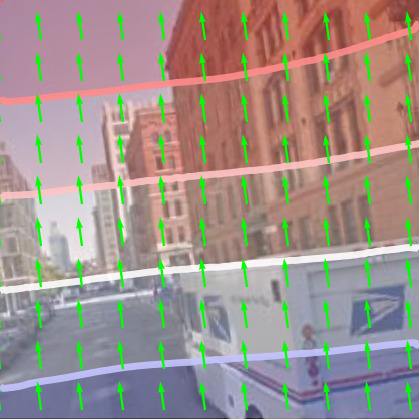}} &
\frame{\includegraphics[width=0.2\textwidth]{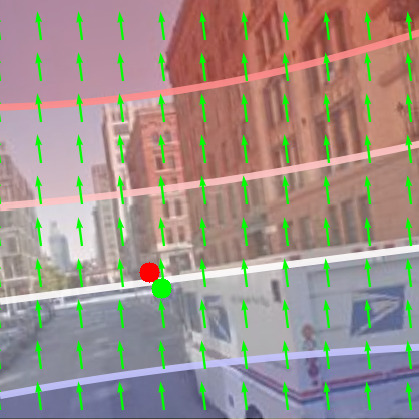}} &
\frame{\includegraphics[width=0.2\textwidth]{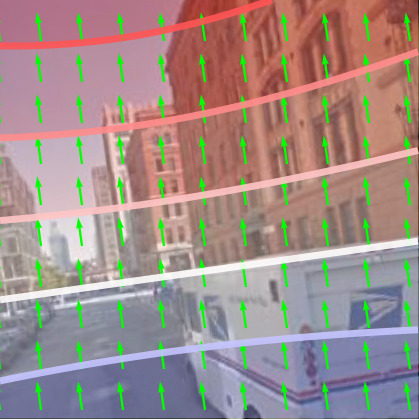}} &
\frame{\includegraphics[width=0.2\textwidth]{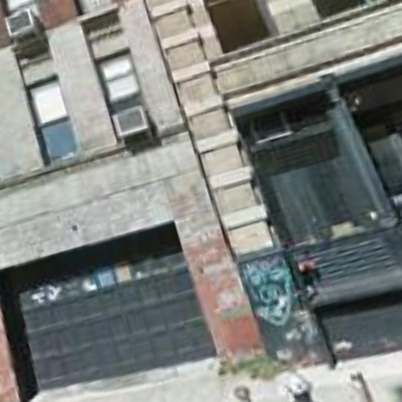}} &
\frame{\includegraphics[width=0.2\textwidth]{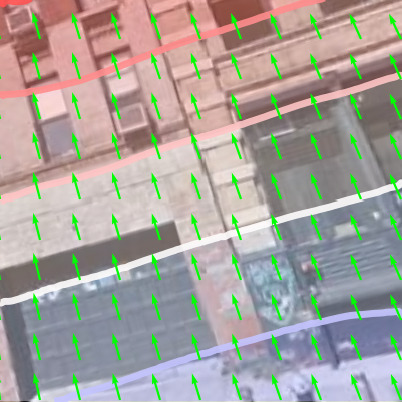}} &
\frame{\includegraphics[width=0.2\textwidth]{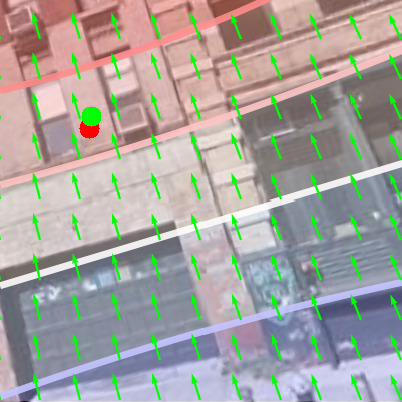}} &
\frame{\includegraphics[width=0.2\textwidth]{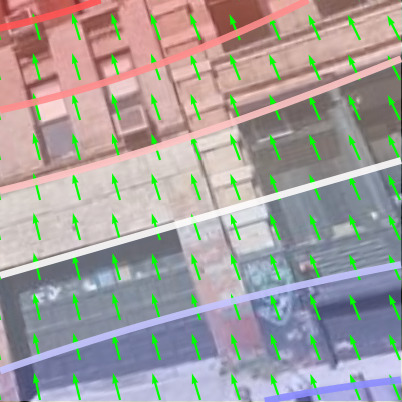}} \\

\frame{\includegraphics[width=0.2\textwidth]{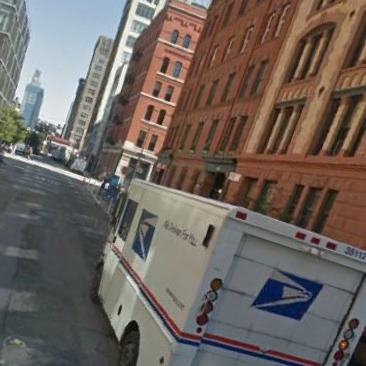}} &
\frame{\includegraphics[width=0.2\textwidth]{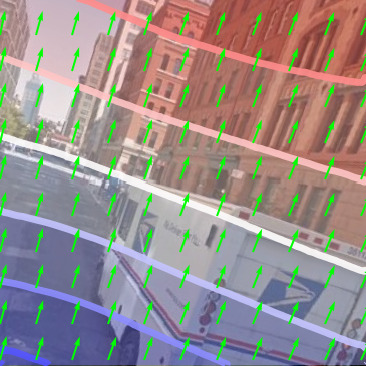}} &
\frame{\includegraphics[width=0.2\textwidth]{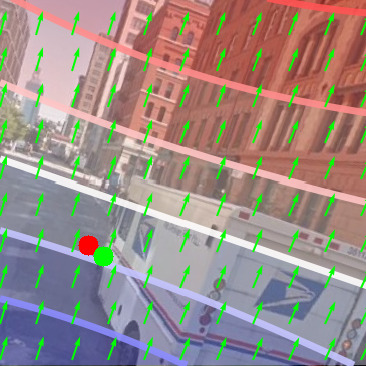}} &
\frame{\includegraphics[width=0.2\textwidth]{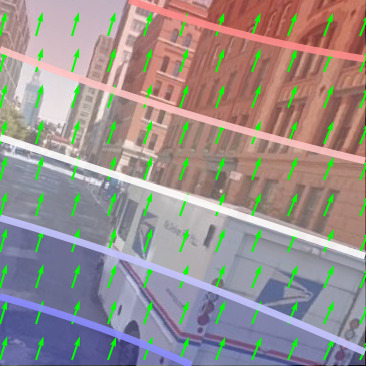}} &
\frame{\includegraphics[width=0.2\textwidth]{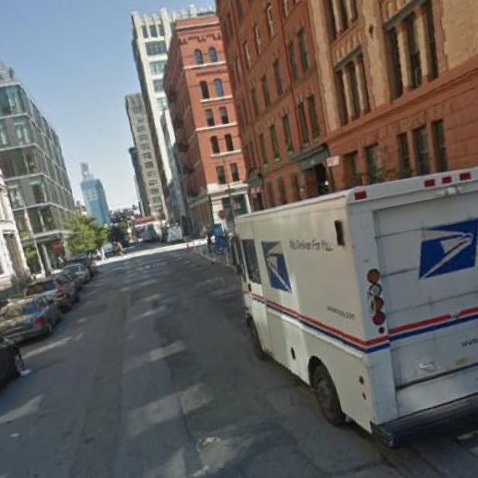}} &
\frame{\includegraphics[width=0.2\textwidth]{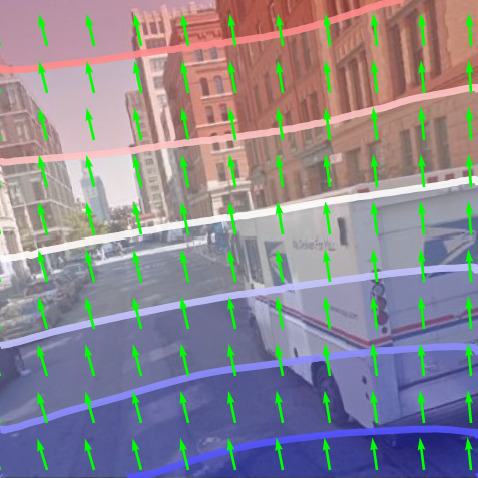}} &
\frame{\includegraphics[width=0.2\textwidth]{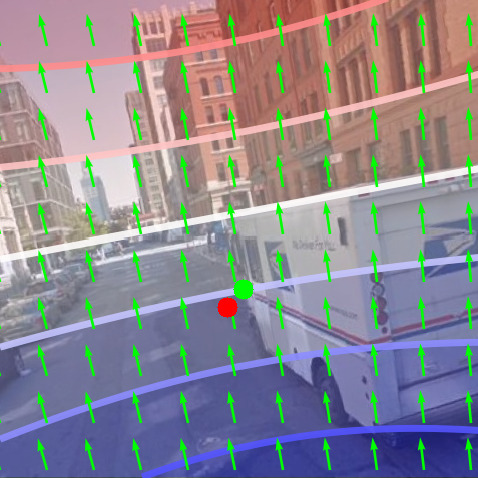}} &
\frame{\includegraphics[width=0.2\textwidth]{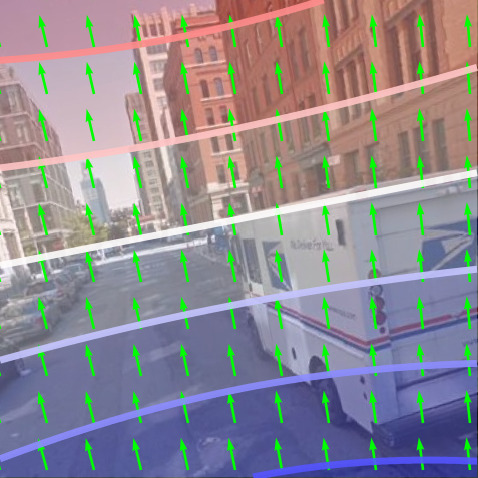}} \\

\frame{\includegraphics[width=0.2\textwidth]{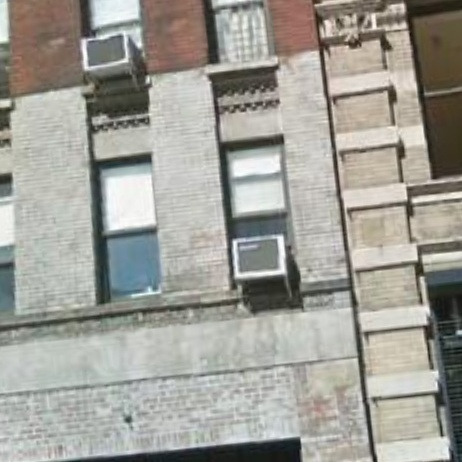}} &
\frame{\includegraphics[width=0.2\textwidth]{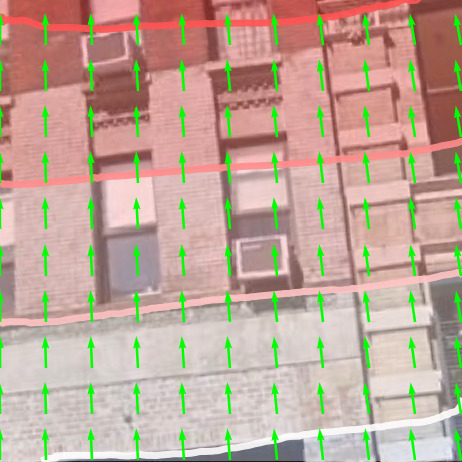}} &
\frame{\includegraphics[width=0.2\textwidth]{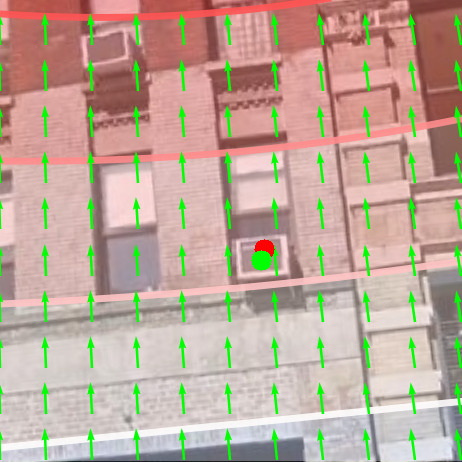}}  &
\frame{\includegraphics[width=0.2\textwidth]{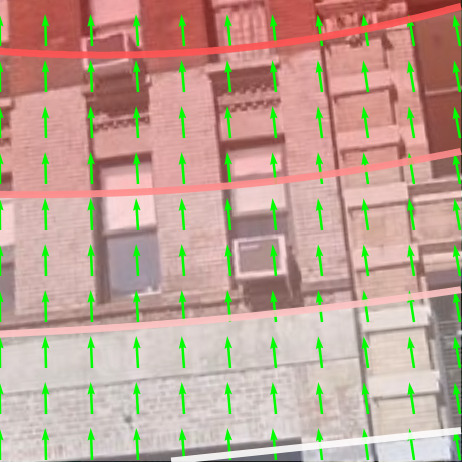}} &
\frame{\includegraphics[width=0.2\textwidth]{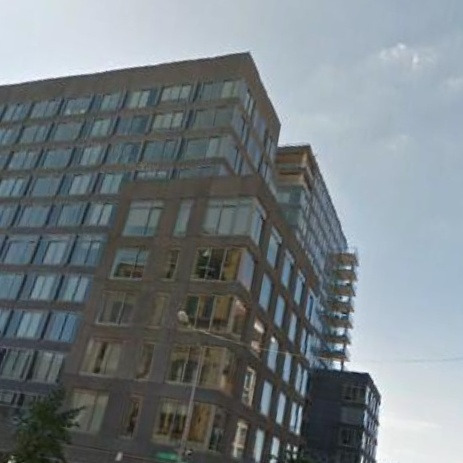}} &
\frame{\includegraphics[width=0.2\textwidth]{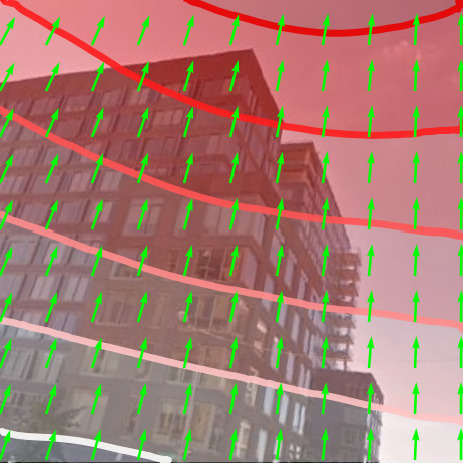}} &
\frame{\includegraphics[width=0.2\textwidth]{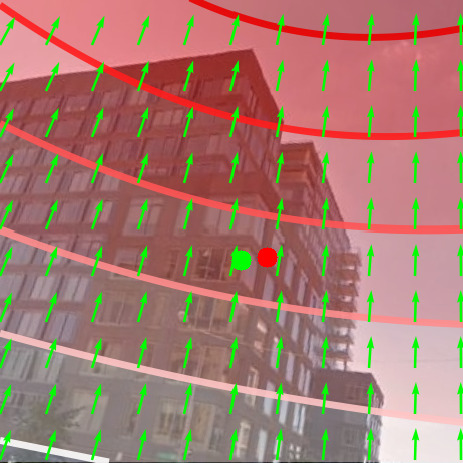}} &
\frame{\includegraphics[width=0.2\textwidth]{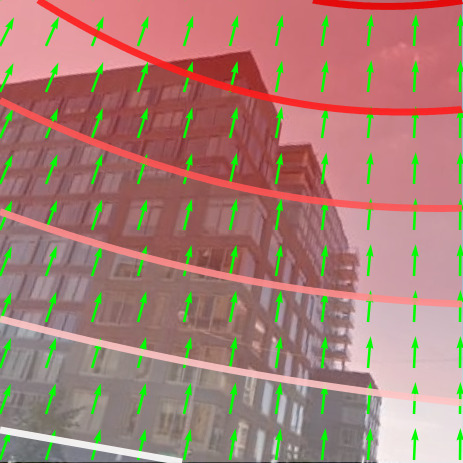}} \\

    \end{tabular}
    }
    
    \caption{Additional qualitative results of PerspectiveNet and ParamNet on GSV {\it uncentered principal-point} images. In the ParamNet column, the ground truth principal point is indicated with a {\color{red} red} dot and the predicted principle point is labeled with a {\color{Green} green} dot. Up-vectors in the green vectors. Latitude colormap: $-\pi/2$ \includegraphics[width=0.4in,height=8pt]{fig/seismic.png} $\pi/2$. }
    \label{fig:supp:gsv_pers}
\end{figure*}

%% file: fig/user_study/instruction.tex
\begin{figure*}[!h]
    \centering
    \scriptsize
    \includegraphics[width=0.8\linewidth]{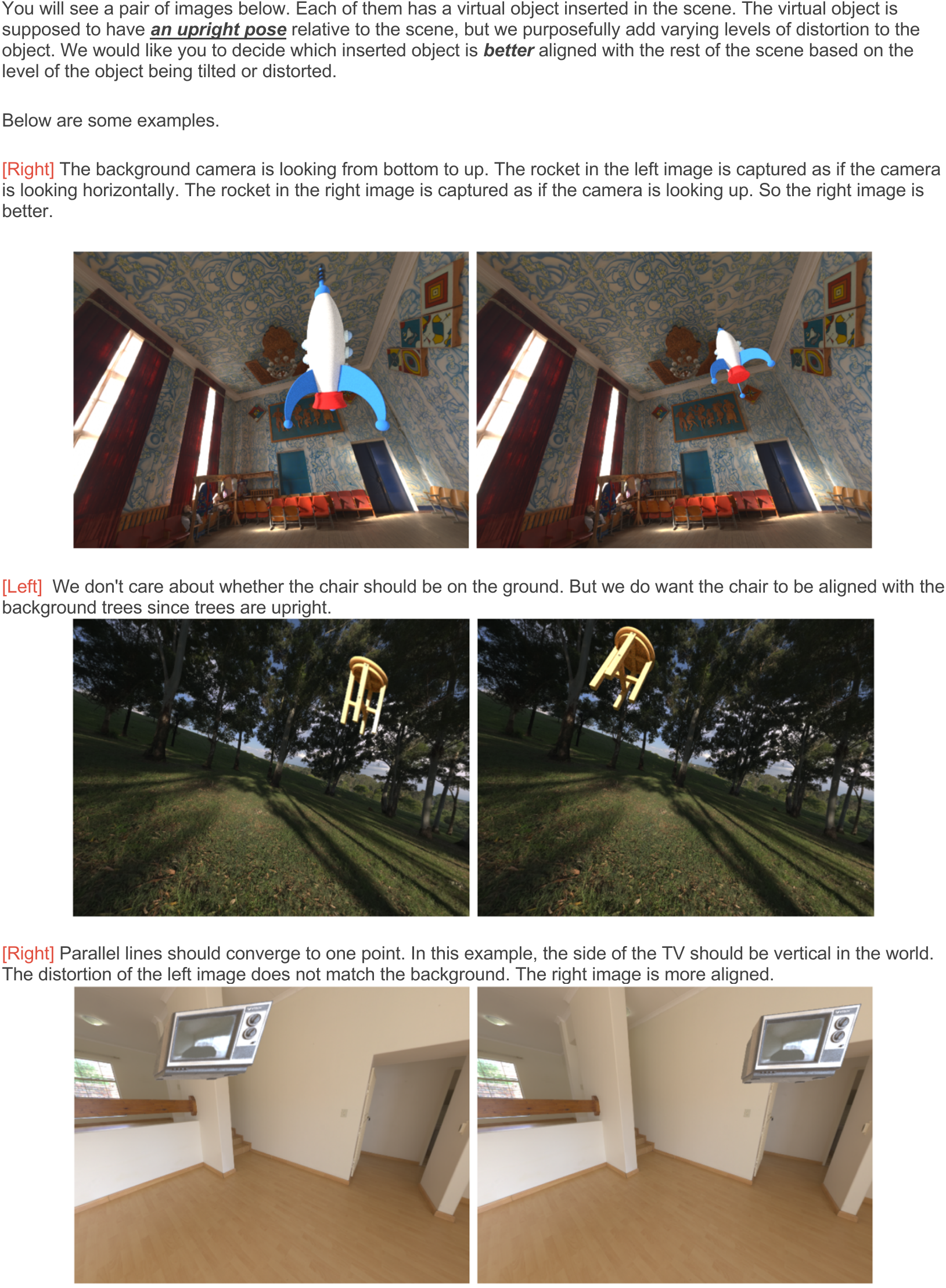}
    \caption{Instructions users see before creating annotation.}
    \label{fig:instruction}
\end{figure*}

%% file: fig/user_study/interface.tex
\begin{figure*}[t]
    \centering
    \scriptsize
    \includegraphics[width=1\linewidth]{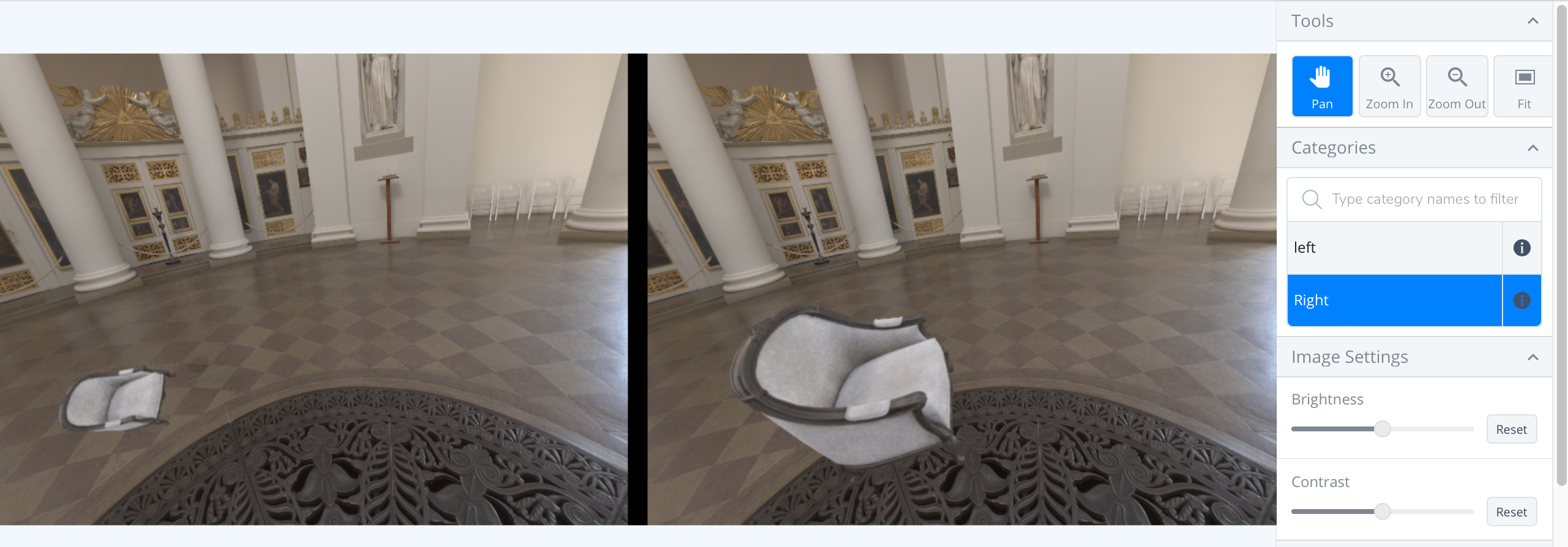}
    \caption{Example user study interface.}
    \label{fig:interface}
\end{figure*}